\documentclass{article}

 \usepackage[dandb, final]{neurips_2025}

\usepackage[utf8]{inputenc} % allow utf-8 input
\usepackage[T1]{fontenc}    % use 8-bit T1 fonts
\usepackage{hyperref}       % hyperlinks
\usepackage{url}            % simple URL typesetting
\usepackage{array,booktabs,tabularx}       % professional-quality tables
\usepackage{amsfonts}       % blackboard math symbols
\usepackage[table]{xcolor}  % for \rowcolor in tables
\usepackage{nicefrac}       % compact symbols for 1/2, etc.
\usepackage{microtype}      % microtypography
\usepackage{xcolor}         % colors
\usepackage{graphicx} % 导入图像的宏包、单图
\usepackage{subfigure} % 导入图像的宏包、子图
\usepackage{pifont}
\usepackage{amsmath}
\usepackage{makecell}
\usepackage{caption}    % 精准控制caption
\usepackage{multirow}   % 表格 占用多行
\usepackage{tcolorbox}  % 有眼色的box
\usepackage{float}      % 浮动
\usepackage{fontawesome}
\usepackage{inconsolata}
\usepackage[export]{adjustbox} % 让 \includegraphics 支持 valign

\usepackage{xspace}

\definecolor{softred}{HTML}{D46A6A}
\definecolor{softyellow}{HTML}{F1D36B}
\definecolor{softblue}{HTML}{7CB6E6}
\definecolor{softgreen}{HTML}{7AC49B}
\newcommand{\dataset}{%
  \textsc{%
    \textcolor{softred}{T}%
    \textcolor{softyellow}{i}%
    \textcolor{softblue}{m}%
    \textcolor{softgreen}{E}%
  }\xspace
}
\newcommand{\titlelogo}{\raisebox{-0.5ex}{\includegraphics[height=2em]{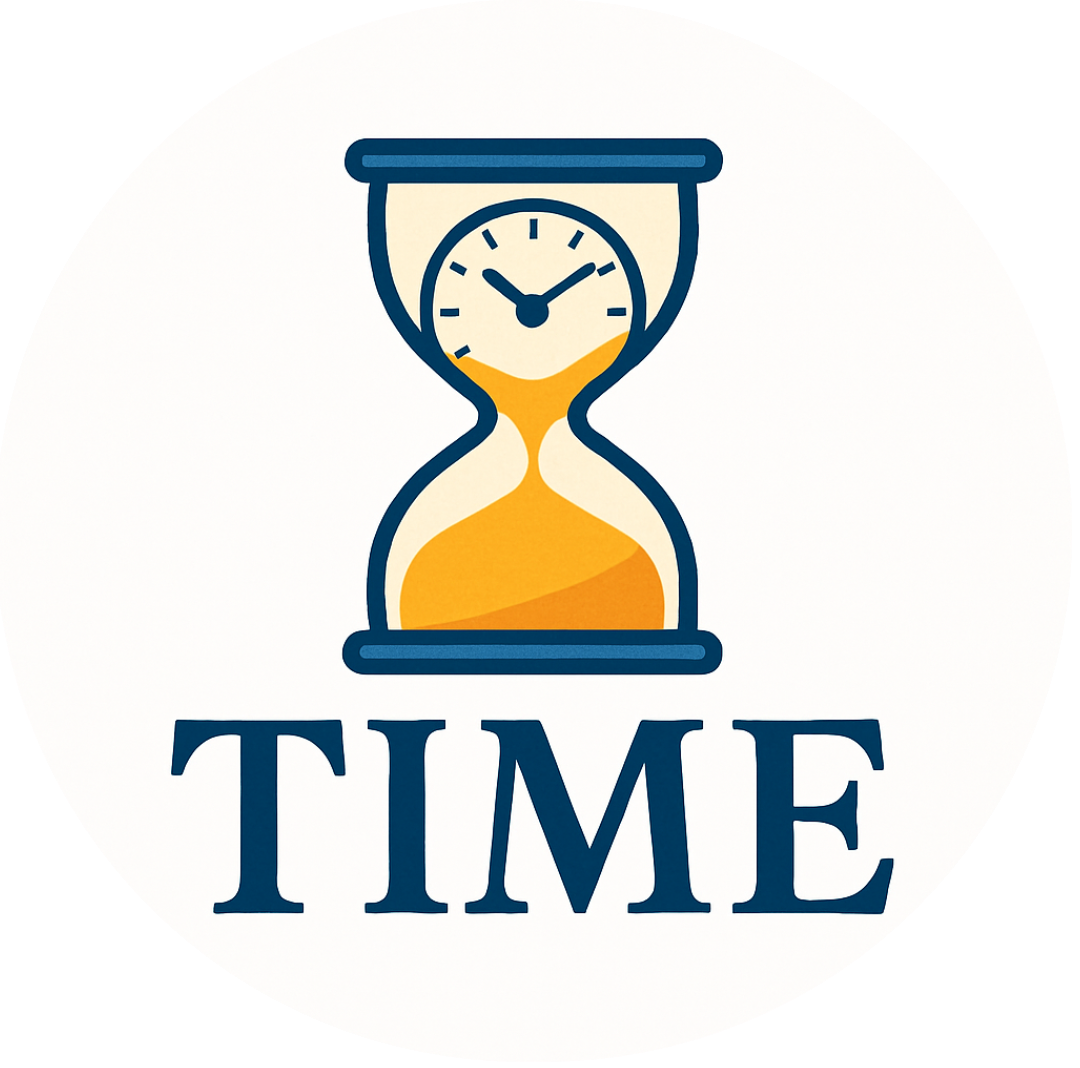}}}

\title{\titlelogo\ \dataset: A Multi-level Benchmark for Temporal Reasoning of LLMs in Real-World Scenarios}

\author{%
  Shaohang Wei\textsuperscript{1}, Wei Li\textsuperscript{1}, Feifan Song\textsuperscript{1}, Wen Luo\textsuperscript{1}\\ 
  \textbf{Tianyi Zhuang}\textsuperscript{2}, \textbf{Haochen Tan}\textsuperscript{2}, \textbf{Zhijiang Guo}\textsuperscript{2}, \textbf{Houfeng Wang}\textsuperscript{1}\\
  \textsuperscript{1}State Key Laboratory for Multimedia Information Processing,\\School of Computer Science, Peking University \\
  \textsuperscript{2}Huawei Noah’s Ark Lab\\
  \texttt{shaohang@stu.pku.edu.cn\space\space wanghf@pku.edu.cn} \\
  \texttt{\{zhuangtianyi, haochen.tan\}@huawei.com}\space\space\texttt{cartusguo@gmail.com}
}

\begin{document}

% \section{Introduction for Supplementary Materials}
% Here are some supplementary materials for the submitted paper \textit{\dataset: A Multi-level Benchmark for Temporal Reasoning of LLMs in Real-World Scenarios}.

% Note that these supplementary materials is not another version of appendix for this paper. For the convenience of reviewing, we provide reviewers with 3 parts of supplementary materials: (1) the details of temporal knowledge graph construction for \textsc{TimE-Wiki}. (2) the prompts for QA synthesis. (3) More details for dataset statistics. (4) QA examples. (5) some supplementary performance for some LLMs (and human annotators) on the \textsc{TimE-Lite}.

% For code, please refer to \url{https://github.com/sylvain-wei/TIME}. For dataset, please refer to \url{https://huggingface.co/datasets/SylvainWei/TIME}. And \dataset's GitHub Webpage URL is \url{https://omni-time.github.io}.
% 0_Abstract
\maketitle
\begin{abstract}
%   The abstract paragraph should be indented \nicefrac{1}{2}~inch (3~picas) on
%   both the left- and right-hand margins. Use 10~point type, with a vertical
%   spacing (leading) of 11~points.  The word \textbf{Abstract} must be centered,
%   bold, and in point size 12. Two line spaces precede the abstract. The abstract
%   must be limited to one paragraph.
  % The abstract paragraph should be indented \nicefrac{1}{2}~inch (3~picas) on
  % both the left- and right-hand margins. Use 10~point type, with a vertical
  % spacing (leading) of 11~points.  The word \textbf{Abstract} must be centered,
  % bold, and in point size 12. Two line spaces precede the abstract. The abstract
  % must be limited to one paragraph.

  % 时间是大语言模型理解真实世界环境的线索。尽管当前大语言模型具有较强的推理能力，但是大语言模型通过理解和推理时间信息处理复杂世界信息的能力仍然未被充分探究。为全面探究大语言模型面对真实环境的时间推理能力，我们设计了一套合成数据方法，并提出了基准TIME，一个包含3个递进level，一共11个子任务，覆盖世界知识（TIME-Wiki）、复杂时间动态信息处理（TIME-News）、very-long 对话（TIME-Dial）三个真实世界场景。我们对现有的开源和闭源的普通LLMs，以及蒸馏版本和最先进的Large Reasoning Models进行了全面评估。我们对不同子任务之间的相关性进行了充分的分析，并发现不同任务对大语言模型的检索能力和推理能力分别构成了不同程度的挑战。我们的TIME基准具备反映真实世界现实挑战、细粒度、情景-任务全面性的特点。我们随机抽取并雇佣人工标注TIME-Lite数据集，以方便未来工作的评测和时间推理研究。

  % 时间信息作为大语言模型理解现实世界的关键线索，其重要性不言而喻。尽管当前大语言模型展现出强大的推理能力，但在处理复杂时间信息方面的潜力仍有待深入挖掘。本研究旨在全面评估大语言模型在真实场景下的时间推理能力，为此我们创新性地设计了一套合成数据方法，并构建了TIME基准测试。该基准包含3个递进层级，共11个子任务，涵盖世界知识（TIME-Wiki）、复杂时间动态信息处理（TIME-News）以及超长对话（TIME-Dial）三大真实场景。我们对现有开源与闭源大语言模型、蒸馏版本以及最先进的大型推理模型进行了系统性评估。通过深入分析各子任务间的关联性，我们发现不同任务对大语言模型的检索与推理能力提出了不同程度的挑战。TIME基准具有以下显著特征：真实反映现实世界挑战、细粒度任务设计、全面覆盖多种情景。此外，我们通过随机抽样并聘请专业人员标注了TIME-Lite数据集，为后续时间推理研究提供了便利的评测框架和基准。
\label{sec: abstract}
  % Temporal reasoning is pivotal for large language models (LLMs) to comprehend the real world. Despite strong general reasoning capabilities, current LLMs' proficiency in processing complex temporal information and performing nuanced temporal reasoning remains largely underexplored in real-world scenarios. To bridge this gap, we introduce \dataset, a multi-level benchmark designed to comprehensively evaluate the temporal reasoning capabilities of LLMs in realistic scenarios. 
  % \dataset features three progressive levels with 11 sub-tasks across key domains: intensive world knowledge(\textsc{TimE-Wiki}), complex temporal dynamics, and very-long multi-session dialogs. We conduct extensive evaluations on a diverse spectrum of models, including large reasoning models and vanilla models. 
  % Our in-depth analysis of inter-task correlations reveals that these sub-tasks distinctively challenge LLMs' retrieval and reasoning capabilities. \dataset is characterized by its realistic challenges, fine-grained task design, and comprehensive scenario coverage. Additionally, we release \textsc{TimE-Lite}, a human-annotated subset to foster future research and standardized evaluation in temporal reasoning.
  % \footnote{Code is available at \url{https://github.com/sylvain-wei/TIME}}
  % \footnote{Dataset is available at \url{https://huggingface.co/datasets/SylvainWei/TIME}}

  Temporal reasoning is pivotal for Large Language Models (LLMs) to comprehend the real world. However, existing works neglect the real-world challenges for temporal reasoning: (1) intensive temporal information, (2) fast-changing event dynamics, and (3) complex temporal dependencies in social interactions. To bridge this gap, we propose a multi-level benchmark \dataset, designed for temporal reasoning in real-world scenarios. \dataset consists of 38,522 QA pairs, covering 3 levels with 11 fine-grained sub-tasks. This benchmark encompasses 3 sub-datasets reflecting different real-world challenges: \textsc{TimE-Wiki}, \textsc{TimE-News}, and \textsc{TimE-Dial}. We conduct extensive experiments on reasoning models and non-reasoning models. And we conducted an in-depth analysis of temporal reasoning performance across diverse real-world scenarios and tasks, and summarized the impact of test-time scaling on temporal reasoning capabilities. Additionally, we release \textsc{TimE-Lite}, a human-annotated subset to foster future research and standardized evaluation in temporal reasoning.
  % \footnote{Code is available at \url{https://github.com/sylvain-wei/TIME}}
  % \footnote{Dataset is available at \url{https://huggingface.co/datasets/SylvainWei/TIME}}
  % \footnote{Project Website: \url{https://sylvain-wei.github.io/TIME/}}

% \begin{center}
% \begin{tabular}{@{}ll@{}}

% \raisebox{-0.2em}{\includegraphics[height=1.4em]{icons/github.png}} &
% \textbf{\texttt{Github Repo}} \hspace{0.3em} 
% \texttt{\href{https://github.com/...}{\textcolor[HTML]{1565C0}{[GitHub Page]}}} \\[0.6em]

% \raisebox{-0.2em}{\includegraphics[height=1.4em]{icons/hf.png}} &
% \textbf{\texttt{TIME}} \hspace{0.3em} 
% \texttt{\href{https://github.com/...}{\textcolor[HTML]{1565C0}{[Huggingface Dataset]}}} \\[0.6em]

% \raisebox{-0.2em}{\includegraphics[height=1.4em]{icons/hf.png}} &
% \textbf{\texttt{TIME-Lite}} \hspace{0.3em} 
% \texttt{\href{https://projectpage.com}{\textcolor[HTML]{1565C0}{[Huggingface Dataset]}}} \\[0.6em]

% \raisebox{-0.2em}{\includegraphics[height=1.4em]{icons/page.png}} &
% \textbf{\texttt{Project Page}} \hspace{0.3em} 
% \texttt{\href{https://huggingface.co/datasets/...}{\textcolor[HTML]{1565C0}{[Project Page \& Leaderboard]}}}
% \end{tabular}
% \end{center}

\begin{center}
\renewcommand{\arraystretch}{1.15}
\setlength{\tabcolsep}{0.5em}
\begin{tabular}{@{}>{\centering\arraybackslash}m{1.6em} >{\ttfamily\bfseries}l >{\ttfamily}l@{}}
\includegraphics[height=1.4em]{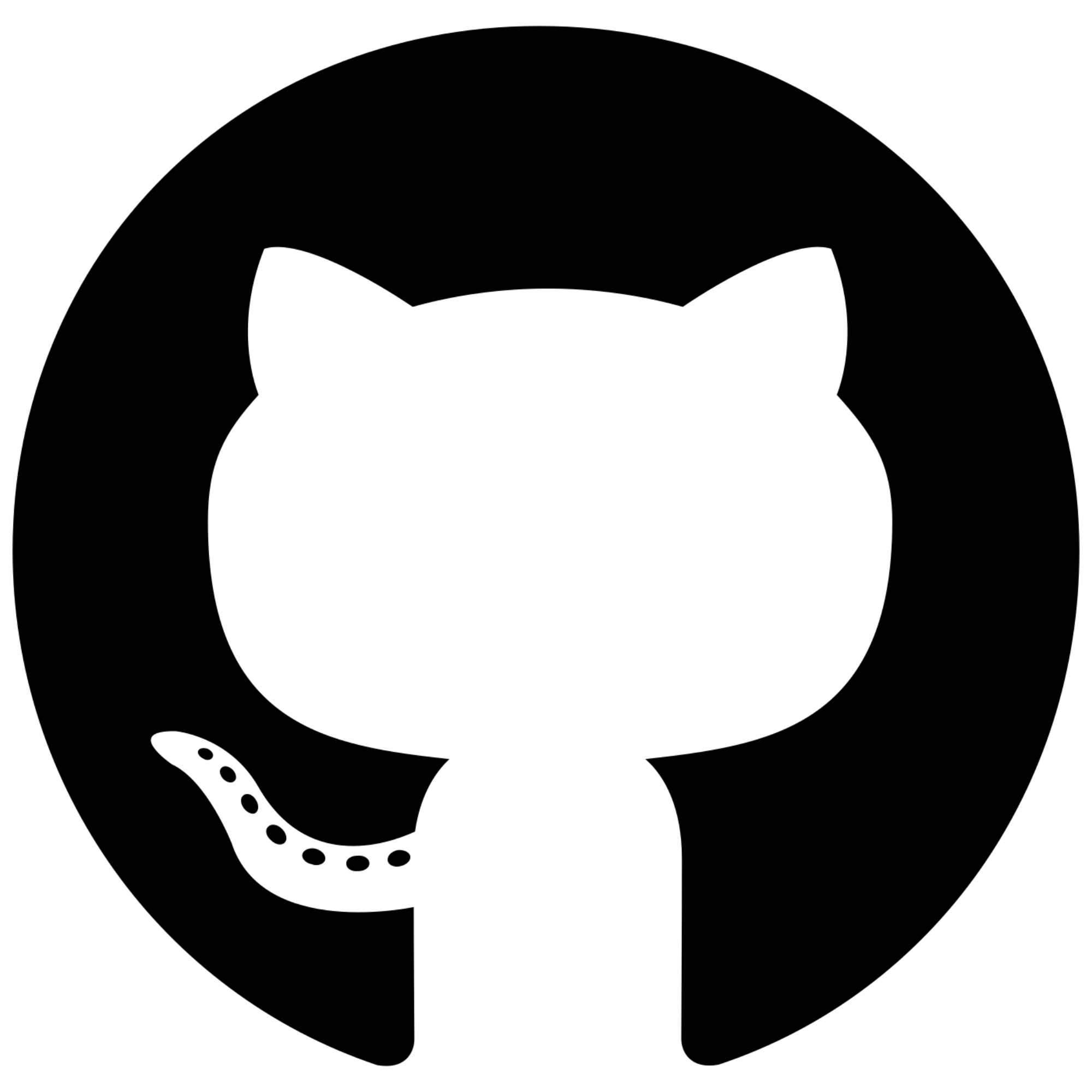} & Github Repo  &
\href{https://github.com/sylvain-wei/TIME}{\textcolor[HTML]{1565C0}{[GitHub Page]}} \\[0.6em]

\includegraphics[height=1.4em]{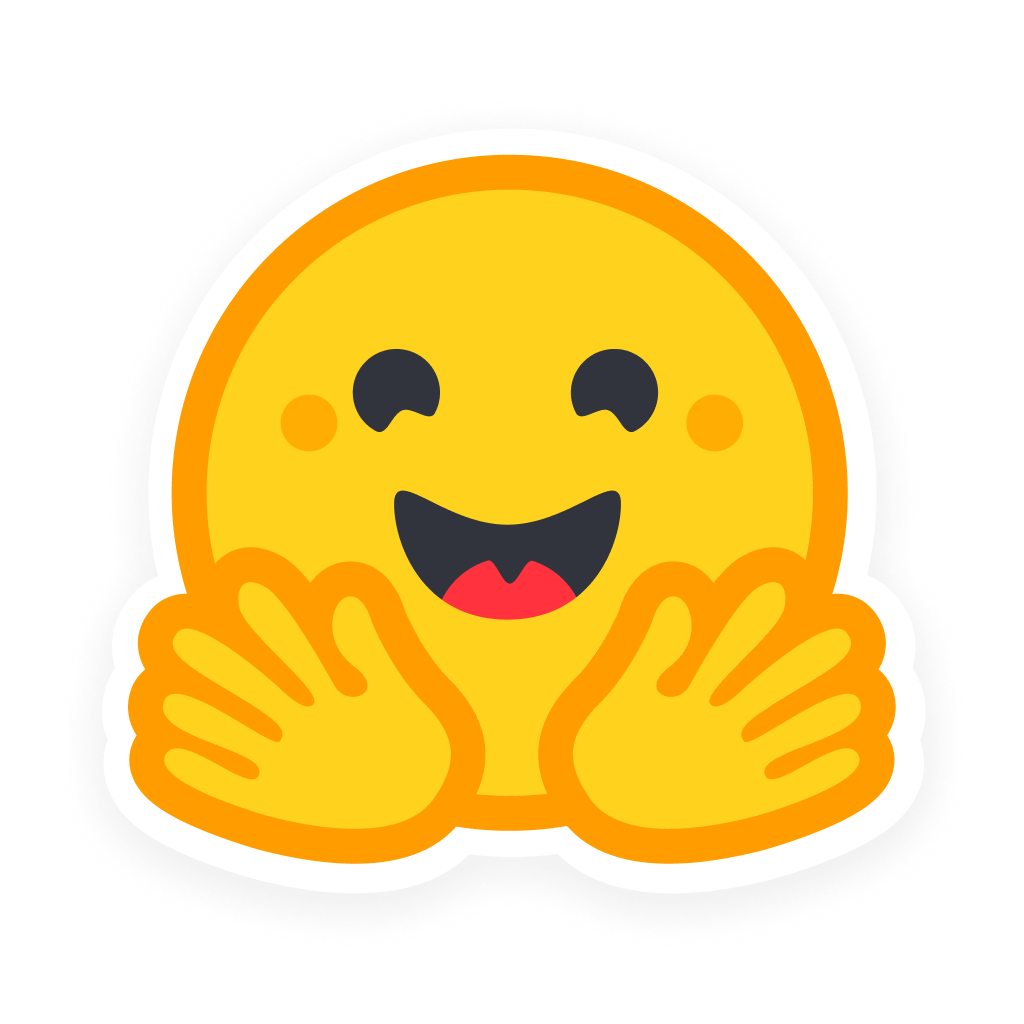} & TIME &
\href{https://huggingface.co/datasets/SylvainWei/TIME}{\textcolor[HTML]{1565C0}{[Huggingface Dataset]}} \\[0.6em]

\includegraphics[height=1.4em]{icons/hf.png} & TIME-Lite &
\href{https://huggingface.co/datasets/SylvainWei/TIME-Lite}{\textcolor[HTML]{1565C0}{[Huggingface Dataset]}} \\[0.6em]

\includegraphics[height=1.4em]{icons/page.png} & Project Page &
\href{https://sylvain-wei.github.io/TIME/}{\textcolor[HTML]{1565C0}{[Project Page \& Leaderboard]}}
\end{tabular}
\end{center}

\end{abstract}

% % 1_Introduction
\section{Introduction\label{sec: introduction}}

Time serves as the thread that weaves together complex events in the real world. Effective temporal reasoning is crucial for Large Language Models (LLMs) to process and comprehend complex events with human-like understanding, particularly in applications requiring integration of historical data and real-time progress tracking. Despite good capabilities of current LLMs across a wide range of reasoning tasks~\citep{SurveyReason2025}, including mathematical problem-solving~\cite{DBLP:journals/corr/abs-2410-07985omnimath,DBLP:conf/nips/HendrycksBKABTS21-math,AutoPSV2024} and code generation~\cite{DBLP:conf/iclr/JimenezYWYPPN24swebench,DBLP:journals/corr/abs-2403-07974livecodebench,EffiLeaner2024,EffiCode2024, DBLP:conf/acl/LiZGM0PHWL25}, they still face challenges in managing temporal understanding in reality.

Temporal reasoning in real-world contexts presents complex challenges: (1) the density of temporal information embedded within world knowledge, (2) the rapid evolution of event details over time, and (3) the complexity of temporal dependencies in social interactions, but existed benchmarks, like TimeBench\cite{chu2024timebench} and TRAM\cite{DBLP:conf/acl/Wang024tram} primarily focus onsimplified scenarios, such as basic temporal commonsense and relationships within short texts and simple QA tasks.
% , without incorporating real-world contextual complexities. 
Consequently, a significant gap remains in exploring temporal reasoning in depth.
% in real-world scenarios.

% 第二段：细致分析LLM面对复杂时间应该分成哪些部分来分析
% A comprehensive evaluation of LLM's temporal reasoning is crucial for deeper understanding about real-world temporal complex events. Different from other reasoning tasks of singular nature, temporal reasoning is a complex and hierarchical set of fine-grained abilities.
% 然而，现有的基准仍然缺少面向现实场景的时间推理全面评估。
% TReMu\cite{DBLP:journals/corr/abs-2502-01630tremu}虽然关注到了多轮对话复杂场景下的event cross-session dependency现象，但它仅关注neuro-symbolic temporal reasoning的方面，缺少对更细粒度的时间推理能力，如对LLM时间计算等能力的分析；而TCELongBench\cite{DBLP:conf/acl/Zhang00MLC24TCELongbench}强调对complex temporal facts的细节检索和未来事件预测，缺少对时间概念本身的考察和理解。
% Due to the intrinsic process of understanding temporal relationships between events, temporal reasoning should cover aspects from basic temporal abilities to complex event-event temporal reasoning. This calls for a new comprehensive evaluation framework for temporal reasoning in real-world scenarios.

% A comprehensive evaluation of temporal reasoning in LLMs is essential for advancing our understanding of complex real-world temporal events. 
On the other hand, temporal reasoning constitutes a hierarchical framework of fine-grained abilities, which is also different from other reasoning tasks that focus on singular capabilities, but is still ignored by current works. For example, 
% However, existing benchmarks fail to provide a holistic assessment of temporal reasoning in realistic scenarios. While 
TReMu\cite{DBLP:journals/corr/abs-2502-01630tremu} involves only neuro-symbolic temporal reasoning while neglecting temporal computation, and TCELongBench\cite{DBLP:conf/acl/Zhang00MLC24TCELongbench} overlooks fundamental temporal concept understanding. In contrast, a robust evaluation framework should encompass both basic temporal abilities and complex event-event temporal reasoning, necessitating the development of a new comprehensive benchmark.

% 第三段：讲自己为了解决这些gap，所提出的思想
% To bridge these gaps, we present \textbf{\dataset}, a multi-level and comprehensive benchmark for temporal reasoning of LLMs in real-world scenarios, 涵盖了3个层次，3种不同real-world挑战性场景，总共有38522个QA。
% Our benchmark 包含了3个数据集:\textsc{TimE-Wiki}, 用于评估模型在复杂世界知识场景中面对时间信息密集挑战时的时间推理能力; \textsc{TimE-News}，用于评估模型在复杂新闻场景中面对事件动态快速变化挑战时的时间推理能力; \textsc{TimE-Dial}则用于评估模型在复杂交互场景中面对大量的时序记忆依赖的时间推理能力.
% 此外，我们分别从3个数据集中分别抽取了子集，并通过人工标注和验证，组成高质量light-weight benchmark\textsc{TimE-Lite}，which 总共有 938 QAs，使未来的时间推理评估更高效、可信.
% \dataset and \textsc{TimE-Lite}中，我们设计了3个层次的任务：(1) basic temporal understanding and retrieval; (2) temporal expression reasoning; (3) complex temporal relationship reasoning. 每一个层次的任务分别从多个细粒度维度共同评估模型的时间推理能力。

To address these limitations, we introduce \textbf{\dataset}, a multi-level comprehensive benchmark for evaluating temporal reasoning in LLMs across diverse real-world scenarios, comprising 38,522 instances. \dataset consists of three datasets: \textsc{TimE-Wiki} assesses temporal reasoning in knowledge-intensive scenarios, \textsc{TimE-News} evaluates temporal understanding in rapidly evolving news contexts, and \textsc{TimE-Dial} examines temporal reasoning in complex interactive settings with extensive temporal dependencies in very-long dialogs. Additionally, we construct \textsc{TimE-Lite}, a high-quality lightweight benchmark containing 938 carefully curated instances through manual annotation and verification, enabling efficient and reliable temporal reasoning evaluation. Both \dataset and \textsc{TimE-Lite} feature a multi-level task structure: (1) basic temporal understanding and retrieval, (2) temporal expression reasoning, and (3) complex temporal relationship reasoning, with each level incorporating multiple fine-grained dimensions to comprehensively assess temporal reasoning capabilities.

% 第四段：总结自己的核心贡献。（参考tianyi姐的思路）
% Our contributions are as follows: xxx 让审稿人一目了然 1）数据集方法；2）数据集本身；3）in-depth analysis

% Our contributions are as follows:
% (1) We propose a new and comprehensive multi-level benchmarking method for probing temporal reasoning abilities of LLMs.
% (2) We present \dataset, the most comprehensive benchmark reflecting key and challenging real-world scenarios.
% (3) We conduct comprehensive evaluation and in-depth analysis on temporal reasoning for a wide range of vanilla and test-time scaled models. 

Our main contributions can be summarized as follows:
\begin{itemize}
    \item We introduce an innovative multi-level evaluation framework that systematically assesses temporal reasoning capabilities across different granularities in LLMs.
    \item We construct \dataset, a comprehensive benchmark that captures the complexity of temporal reasoning in diverse real-world scenarios, including knowledge-intensive, dynamic events, and multi-session interactive contexts.
    \item We conduct a comprehensive evaluation and in-depth analysis on temporal reasoning for a wide range of LLMs.
\end{itemize}

\section{Related Work}

\noindent \textbf{Temporal Understanding in Natural Language}
Temporal understanding in Natural Language Processing (NLP) has a rich history, initially focusing on extracting time expressions and temporal relationships \cite{pustejovsky2003timeml, vashishtha2019fine, verhagen2007semeval, verhagen2010semeval, uzzaman2013semeval, llorens2015semeval, mathur2021timers, strotgen2010heideltime}. The advent of pre-trained language models facilitated the exploration of more complex phenomena, including implicit time reasoning \cite{zhenjia2018tempquestions, DBLP:conf/acl/TanNB23TempReason, DBLP:conf/acl/TanNB24complextr} and cross-event temporal relationships \cite{DBLP:conf/naacl/ZhouRNKSR21TRACIE, DBLP:conf/emnlp/WeiSMYLZZL23MenatQA, DBLP:conf/acl/XiongPKF24TGQA, DBLP:journals/corr/abs-2501-03040, DBLP:conf/acl/TanNB23TempReason, DBLP:conf/acl/TanNB24complextr, DBLP:conf/acl/SuLZZQ0BCZ24cotempqa}. Research has also addressed commonsense temporal knowledge \cite{DBLP:conf/lrec/VirgoCK22DurationQA, DBLP:conf/emnlp/ZhouKNR19MCTACO} and enhanced event-based temporal question-answering through knowledge graphs \cite{DBLP:conf/acl/XiongPKF24} or specialized language model training \cite{DBLP:conf/emnlp/SonO23, DBLP:conf/www/ChenWZ023, DBLP:conf/emnlp/YangLBL23, DBLP:conf/emnlp/Han0P21}. Unlike these prior efforts that often target specific temporal aspects, \dataset proposes a unified framework for a comprehensive evaluation of temporal understanding spanning time and events.
\noindent \textbf{Temporal Reasoning in Real-World Scenarios\xspace}
Advancements in language models have facilitated deeper exploration of temporal reasoning, particularly concerning event ordering and causality. Benchmarks like TimeQA \cite{DBLP:conf/nips/ChenWWW21timeqa} assess time-sensitive question answering, while RealTimeQA \cite{DBLP:conf/nips/KasaiST0A0RS0I23realtimeqa}, FreshLLM \cite{DBLP:conf/acl/VuI0CWWTSZLL24freshllms}, and StreamingQA \cite{DBLP:conf/icml/LiskaKGTSAdSZYG22streamingqa} address adaptation to dynamic knowledge. SituatedQA \cite{DBLP:conf/emnlp/ZhangC21situatedqa} and SituatedGen \cite{DBLP:conf/nips/0002023situatedgen} evaluate the integration of temporal and geographical commonsense into text understanding. Domain-specific studies further explore temporal reasoning. In news, TCELongBench \cite{DBLP:conf/acl/Zhang00MLC24TCELongbench} probes temporal understanding of complex events, focusing on details, ordering, and prediction. For conversations, TimeDial \cite{DBLP:conf/acl/QinGUHCF20timedial} examines everyday temporal commonsense, and TReMu \cite{DBLP:journals/corr/abs-2502-01630tremu} tackles temporal localization and long-range dependencies in long-form dialogues. Unlike TCELongBench and TReMu, which concentrate on specific temporal aspects, \dataset offers a comprehensive evaluation. It employs three progressive complexity levels and diverse subtasks for fine-grained analysis of temporal reasoning in extended contexts.

\noindent \textbf{Benchmarks for Temporal Reasoning}
Existing studies evaluate LLM temporal reasoning using various benchmarks like TRAM \cite{DBLP:conf/acl/Wang024tram}, which assesses event sequence understanding but lacks challenge for current models. Other efforts include a benchmark using six existing datasets \cite{jain2023language} and TimeBench \cite{chu2024timebench}, which aggregates 10 datasets across symbolic, common-sense, and event-temporal levels. However, these benchmarks have limitations: TimeBench's simpler tasks are less challenging for improving LLMs, and the diverse evaluation contexts and difficulty levels across datasets can introduce biases, hindering fair and consistent assessment. In contrast, \dataset provides a unified evaluation context, enabling fair, fine-grained, and challenging assessment of LLM temporal reasoning in real-world scenarios.

% Existing studies have comprehensively analyzed the temporal reasoning capabilities of LLMs. TRAM~\cite{DBLP:conf/acl/Wang024tram} evaluates models across 10 dimensions of event sequence understanding, but its benchmark lacks challenge for most state-of-the-art models. One benchmark~\cite{jain2023language} assesses LLMs using six existing datasets, while TimeBench~\cite{chu2024timebench}, a temporal reasoning benchmark, comprises 10 existing datasets covering symbolic, common-sense, and event-temporal levels. However, as LLM reasoning improves, TimeBench’s simple contextual temporal reasoning QA tasks become less challenging. Moreover, the varying evaluation contexts and difficulty levels across these datasets may introduce model-specific biases, hindering fair and consistent assessment of temporal understanding and reasoning capabilities. In contrast, our dataset evaluates LLM temporal reasoning within a unified context, enabling fair, fine-grained, and challenging assessments of their capabilities in real-world scenarios.

% 思路：广泛补全和关键对比
% 1. Temporal Reasoning （概念相关文章、关键benchmark按照分类形式来写）
% 2. Long context Reasoning

% % 3_Benchmark
\section{\dataset: Benchmark Construction}
% 预计有3页左右的占比（包括这部分的图片）
% 关于数据集的图片：
% 1. Overall illustraition（参考Omni-MATH figure 2）
% 2. Data collection and annotation process

% In this section, we explore the construction process of TIME. 这将involve任务的定义和设计原则、数据源选择、数据集构建和质量控制，最后我们提供了数据集的统计指标。
\begin{figure}
  \centering
  \includegraphics[width=\textwidth]{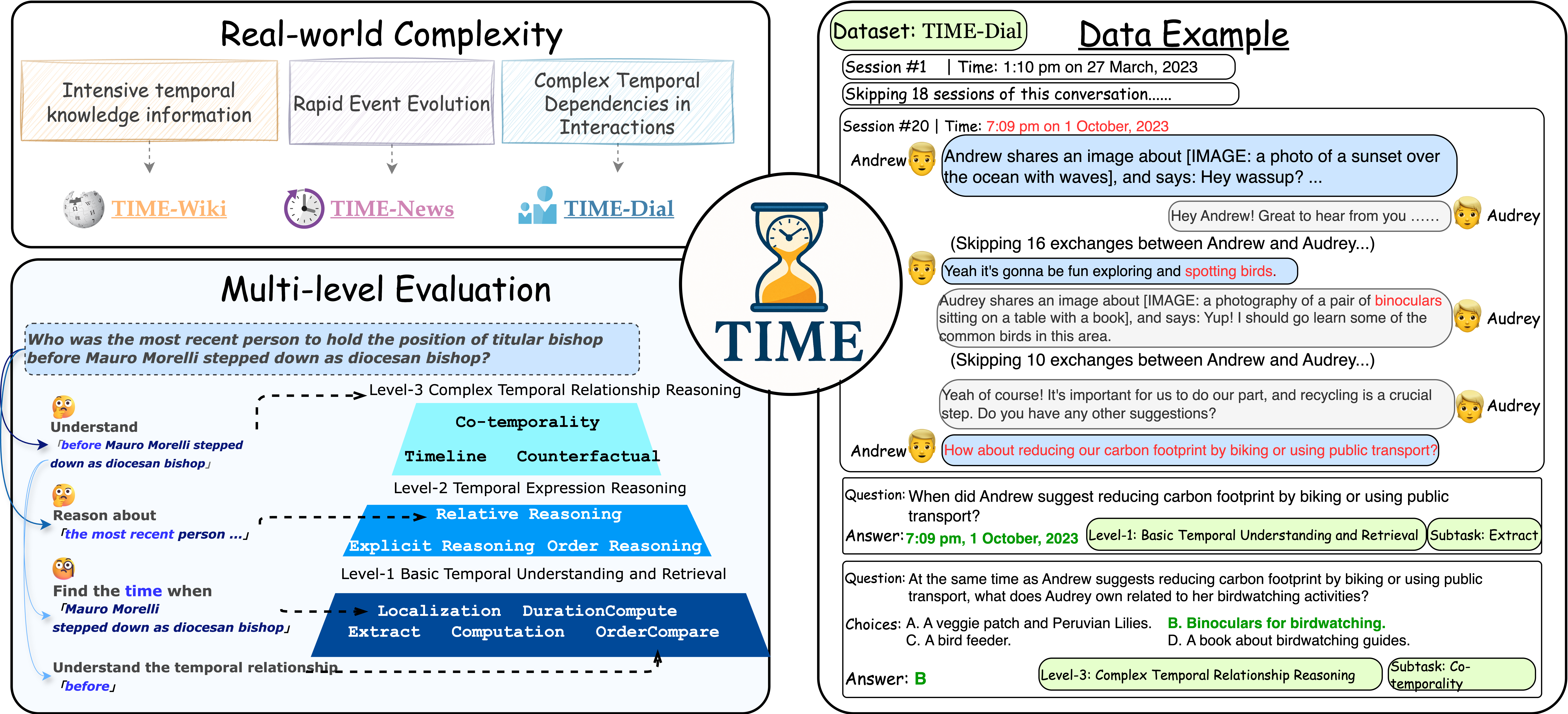}
  \caption{An overview of \dataset. The top-left block illustrates three key challenges of real-world complexity and their corresponding dataset construction. The bottom-left quadrant depicts a three-level tasks. One data example from \textsc{TimE-Dial} is shown on the right.
  \label{fig: An overview of the dataset.}}
\end{figure}

This section details the construction of \dataset, covering task definition and design principles (\S\ref{sec: Task Definition}), data source selection (\S\ref{sec: Data Source}), and dataset construction with quality control (\S\ref{sec: Dataset Construction}). Finally, we introduce \textsc{TimE-Lite}, a high-quality, manually verified sub-dataset (\S\ref{sec: TIME-Lite}).

% Dataset statistics can be seen in Table \ref{tab:dataset_statistics_short} and Table \ref{tab:complete_dataset_stats}.

% 在 LaTeX 中，要让一个单词（或短语）以“高级的衬线体”显示，通常不是指切换到另一种衬线字体（因为文档默认就是衬线体，如 Times New Roman），而是指应用特定的样式来突出显示，同时保持衬线字体特性。
% 几种常用的方法（保持衬线体并应用样式）：
% 1. 小写字母大写（Small Caps）：使用 \textsc{Word}。这会将单词中的小写字母转换为较小的大写字母，常用于缩写或专有名词，例如本节标题中的 \textsc{TimE}。
%    示例：\textsc{TimE} Benchmark
% 2. 加粗（Bold）：使用 \textbf{Word}。这是最常见的强调方式。
%    示例：\textbf{重要} 概念
% 3. 斜体（Italic）：使用 \textit{Word} 或 \emph{Word}（推荐使用 \emph{}，它会根据上下文自动选择斜体或其他强调方式）。常用于书名、术语首次出现等。
%    示例：这是一个 \textit{新术语}。
% 4. 罗马体（Roman）：使用 \textrm{Word}。这通常用于确保文本在切换到其他字体（如无衬线体 \textsf{} 或打字机字体 \texttt{}）后，能回到默认的衬线体。在默认就是衬线体的情况下，效果与普通文本相同。
%
% 对于像“TIME”这样的基准名称，\textsc{TimE} 是一个非常常见且符合“高级”感的样式。

\subsection{Task Definition\label{sec: Task Definition}}
% TIME旨在细粒度、全面深入地探索真实世界中时间推理的真实挑战。我们首先总结了时间推理中的挑战，并将其划分为3个层次递进的三个层次，并提出各类子任务以更好地对齐真实世界时间推理的内在本质过程。

\dataset is designed for a fine-grained and comprehensive exploration of real-world temporal reasoning challenges. We structure these challenges into three progressive levels and propose various task formats that better capture the intrinsic nature of temporal reasoning in real-world scenarios.

% 为什么设计3个不同的层次？（从high-level角度来解释这个选择的原因）
% 每一个层次里面有若干个子任务，分别是什么？（不介绍之间的联系。从动机到具体的事项这样推进）
% 一个类似于RULER这篇benchmark的热力图，表示每一个任务之间的关系（取一个标准的模型的结果来进行对比）

\subsubsection{Design Principal}
% 真实世界文本信息包含复杂的内容，而其中时间信息是关键的线索。TIME基准旨在模拟人类在处理复杂、动态的世界信息时利用时间概念更好地理解世界的过程，衡量LLM利用时间解决真实世界问题的能力。人类倾向于首先对时间概念进行准确捕捉和理解（即Level-1: Basic Temporal Understanding and Retrieval）；其次，需要对包含implicit和ambiguous的时间表达式整合上下文信息进行推理，以定位事件细节（即Level-2: Temporal Expression Reasoning）；最后，事件蕴含的复杂时间关系对于厘清时间线、潜在的事件因果至关重要(Level-3: Complex Temporal Relationship Reasoning)。这三类挑战层次递进，相互关联。下面，我们将具体介绍三个层次的时间推理任务。
Real-world text information contains complex content, among which temporal information is a crucial clue. \dataset aims to simulate the process by which humans utilize temporal concepts to better understand the complex and dynamic world information, and to measure the ability of LLMs to solve real-world problems using time. Humans tend to first accurately capture and understand temporal concepts (Level-1: Basic Temporal Understanding and Retrieval); secondly, it requires integrating contextual information to reason about implicit and ambiguous temporal expressions to locate event details (Level-2: Temporal Expression Reasoning); finally, the complex temporal relationships embedded in events are crucial for clarifying timelines and potential event causality (Level-3: Complex Temporal Relationship Reasoning). These three types of challenges are progressive and interconnected. The temporal reasoning tasks at three levels are detailed below:
\noindent \textbf{Level-1: Basic Temporal Understanding and Retrieval.}
Level-1 requires models to establish fundamental temporal information processing capabilities. We design five subtasks organized through three complementary aspects: (1) \texttt{Extract} assesses direct retrieval of temporal expressions (time points, periods, relative time) from text, paired with (2) \texttt{Localization} that evaluates event-time mapping accuracy through temporal positioning of occurrences - together forming the basis of temporal information retrieval. For temporal quantification, we develop (3) \texttt{Computation} testing duration calculation between explicit time markers, combined with (4) \texttt{DurationCompare} measuring interval comparison capability between events. Finally, temporal sequencing is addressed through (5) \texttt{OrderCompare} that examines chronological ordering understanding. This tripartite structure evaluates core competencies through: basic temporal retrieval (\texttt{Extract}+\texttt{Localization}), quantitative temporal analysis (\texttt{Computation}+\texttt{DurationCompare}), and sequential relationship understanding (\texttt{OrderCompare}).

% L2-1 (Explicit Reasoning)：基于明确的时间表达（时间点、时间范围）提问。两个明确时间节点不可以出现在原文中。LLM需要对时间进行理解和推理才可能定位正确事件细节。例 Q:"Mauro Morelli在1967-1973年之间的工作是？" A: "天主教牧师"
% L2-2 (Order Reasoning)：基于次序形式的时间表达提问。例 Q: "Mauro Morelli第二份工作是？" A: "天主教主教"
% L2-3 (Relative Reasoning)：基于相对时间表达提问。例 Q: "离A事件（context内某事件）发生前，Mauro Morelli最近的工作地点是？（或者 几天前/几个月前）" A: "天主教堂"

% Level-2任务要求模型通过时间表达式推理定位事件细节。给定的时间表达式均无法直接对应到原文，而是需要结合原文的时间点进行比较、计算才能推理得到问题所指向的原文事件。我们设计三个子任务：(1)显式时间推理：基于未在原文出现的时间点/范围进行推理；(2)次序推理：基于事件顺序定位；(3)相对时间推理：结合上下文推断相对时间关系。这些任务共同考察在需要对时间本身进行复杂的多跳推理前提下，模型对复杂事件细节的理解。
\noindent \textbf{Level-2: Temporal Expression Reasoning.}
Level-2 requires models to locate event details through temporal expression reasoning. We design three subtasks: (1) \texttt{Explicit Reasoning} that demands inference based on unmentioned time points/ranges (e.g., Q: ``What was Mauro Morelli's occupation between 1967-1973?''), (2) \texttt{Order Reasoning} that requires temporal positioning through ordinal expressions (e.g., ``Mauro Morelli's second job''), and (3) \texttt{Relative Reasoning} that involves contextual interpretation of relative temporal references (e.g., ``Where did Mauro Morelli work closest to Event A?''). These tasks collectively evaluate models' ability to understand complex event details under the premise of performing \textbf{multi-hop temporal reasoning}.

% L3-1 (Co-temp)：事件的共现关系判断。对两个在时间线上有重叠的对象进行事件细节提问。例如Q: "当Elon Musk在OpenAI工作的同时，还在哪里工作？" A: "SpaceX和Tesla"
% L3-2 (Timeline)：事件时序关系推断（多事件重排序）。要求模型输出结果为数字序列。例如Q: "请将下面8个事件按照事件顺序排序: (1)Obama administration said it is keeping some specifics from Israel; (2)Benjamin Netanyahu claims that he knows the details of the deal being forged with Iran over its nuclear program; (3) Netanyahu …" A: "(1)(2)(5)(4)(7)(3)(8)(6)"
% L3-3 (Counterfactual): 反事实/违背context假设的时间事实情况下，对上下文中提到的时间事实的提问。在这类任务中，每次query包含反事实的假设和question，该反事实的假设是与原文相违背的时间信息，而模型需要遵循给定的反事实假设，基于context进行推理。
% Level-3任务要求模型理解并推理多个事件之间的复杂时间关系。我们设计三个子任务：(1)共现关系判断（Co-temp）：识别时间线上重叠事件的共现关系；(2)时间线重构（Timeline）：推断多事件的正确时序关系；(3)反事实推理（Counterfactual）：在违背原文时间假设的前提下进行推理。这些任务共同考察模型对复杂时间交互关系的理解能力。
\noindent \textbf{Level-3: Complex Temporal Relationship Reasoning.}
Level-3 requires models to comprehend and reason about complex temporal relationships among multiple events. We develop three subtasks: (1) \texttt{Co-temporality} that identifies overlapping temporal relationships between concurrent events (e.g., ``Where did Elon Musk work concurrently with his OpenAI position?''), (2) \texttt{Timeline} that infers correct chronological ordering of multiple events (e.g., sorting 8 political events into temporal sequence), and (3) \texttt{Counterfactual Reasoning} that demands temporal inference under altered temporal premises contradicting the original context (e.g., ``If Event X occurred 3 years later, how would it affect Event Y?''). These tasks collectively evaluate models' capacity to handle \textbf{multi-event temporal interactions} through co-occurrence analysis, timeline construction, and hypothetical temporal reasoning.

\subsection{Data Source\label{sec: Data Source}}
% 这里需要画一个data construction pipeline（突出核心的部分）
\begin{figure}
  \centering
  \includegraphics[width=\textwidth]{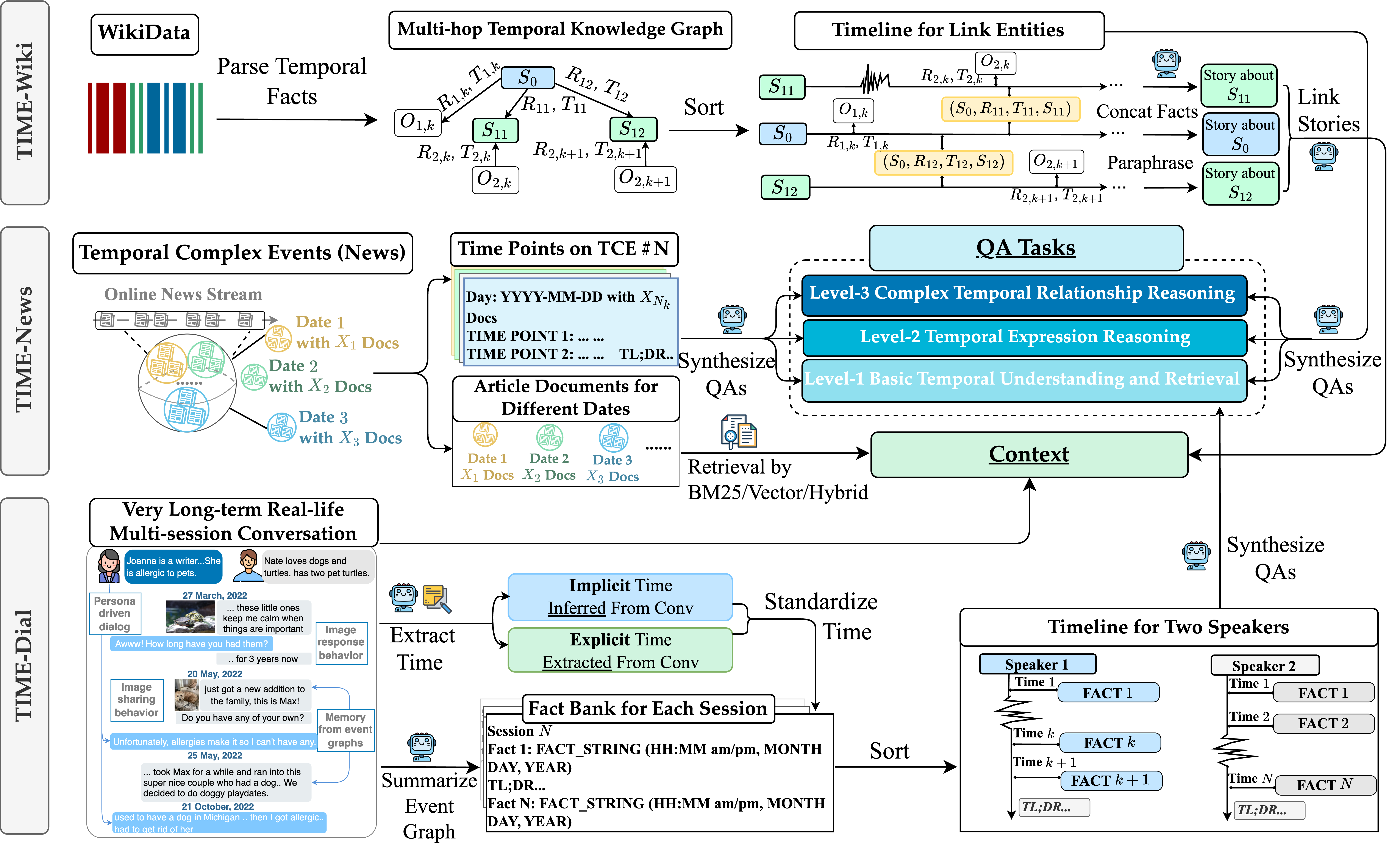}
  \caption{Dataset construction pipeline for \dataset. In the process of QA synthesis for each sub-dataset, we first collect temporal facts (temporal knowledge graphs for \textsc{TimE-Wiki}, time points for \textsc{TimE-News}, fact bank for \textsc{TimE-Dial}). Then timelines are generated for QA data synthesis.}
  \label{fig: dataset construction pipeline}
\end{figure}

% 介绍为什么选择这3个数据源？wikidata应该代表广泛的世界知识；news应该代表复杂世界知识在长上下文中的应用；long-dialog应该代表日常生活、常识相关的时间推理。
% 按照数据构建的pipeline，分点来写
% 介绍为什么选择LoCoMo和REALTALK的时候，需要说明它的特点：（1）相比于其他数据集，这两个数据集是multi-session的，并且最关键的是每一个session的具体时间是非常清楚的。

% 审稿人还可能会问，既然叫TIME-Wiki, TIME-News, TIME-Dial，为什么后面两种只用了1-2种数据集进行增广得到QA，而不是多样化的数据源？这里需要强调，我们选择TCELongBench来代表News是因为它的数据库数量足够庞大，并且容易获取，并且具有较为丰富的时间信息；我们选择LoCoMo和REALTALK代表Dial是因为它们的时间信息足够充分和复杂，而且上下文长度足够长，聊天内容较为多样（比如share photo的多模态信息，相比于一般的文本更复杂、丰富）和接近现实（比如很多small talk等语气、情景词句）。
% ---------------------------------------正文-------------------------------------------------------

% 我们分别从Wikidata、来自新闻的temporal complex events和very long-term real-life multi-session conversation来构造TIME数据集：TIME-Wiki, TIME-News以及TIME-Dial。

\noindent \textbf{\textsc{TimE-Wiki}}
% 我们选择WikiData作为TIME-Wiki的data source。WikiData具有大量持续更新的real-world temporal facts，其能够反映在广泛世界中的time-evolving facts。此外，WikiData提供了时间关系的标签，方便提取时间知识图谱。为了全面反映现实世界知识，我们精心选择了6个类、34种典型关系的WikiData Properties以检索事实关系，如表格\ref{tab:wikidata_relations}所示。
We choose Wikidata as the data source, leveraging its extensive collection of continuously updated real-world temporal facts that capture time-evolving knowledge across diverse domains. Wikidata's structured temporal relations facilitate efficient extraction of temporal knowledge graphs. To ensure comprehensive coverage of real-world knowledge, we systematically select 6 categories encompassing 34 representative Wikidata properties for fact retrieval, as detailed in Table \ref{tab:wikidata_relations}.

\noindent \textbf{\textsc{TimE-News}} 
% TODO:
% 1.在这里，我们并不强调News的动机，只是描述逻辑、细节。请在intro部分检查是否清楚陈述了TIME-Wiki、TIME-News等的动机。
% 2.这里需要画一个图，放在附录，来表达Temporal Complex Events是什么、提取的时间线是怎样的形式，用例子形式比较好（可以参考pipeline的图）。
% 3.这里需要引用一个附录中的表格，来说明这个数据源的statistics，比如说平均每一个Temporal Complex Events对应的article数量、长度等数据。
% 我们使用online news articles作为TIME-News的data source。工作\cite{DBLP:conf/acl/Zhang00MLC24TCELongbench}提供了大规模、高质量的online news articles和对应的Timelines，用于反映多个Temporal Complex Events。「TODO:这里要插入Temporal Complex Events的图片、statistics」如图xx所示，每一个complex event包含多个日期，复杂事件的多个细节随着日期不断发生变化。TIME-News的数据源statistics如表xxx所示。
We utilize online news articles as the data source. Previous work\cite{DBLP:conf/acl/Zhang00MLC24TCELongbench} provides large-scale, high-quality online news articles accompanied by corresponding timelines, effectively capturing multiple temporal complex events. A temporal complex event (TCE) refers to a sequence of interrelated events centered around a specific topic within a defined time period, such as the Israeli-Palestinian conflict that occurred between February and March 2015. As illustrated in Figure \ref{fig: example_TCE}, each temporal complex event encompasses multiple dates, with event details evolving. The statistical characteristics of the \textsc{TimE-News} data source are presented in Table \ref{tab:tce_statistics}.

\noindent \textbf{\textsc{TimE-Dial} }
% TODO:
% 1.这里需要引用/画一个表+图，在附录，表达45条very long-term multi-session conversations的数据分布。
% 1.1 表：上下文长度（总共的平均上下文长度、LoCoMo的平均上下文长度、REALTALK的平均上下文长度、整个数据源context中的最长、最短上下文tokens数量）
% 1.2 图：数据源LoCoMo占比35条，REALTALK占比10条。
% 2.需要放一个图，在附录，表示每一条数据的case。
% 我们使用very long-term multi-session conversation作为TIME-Dial的data source。我们选择LoCoMo\cite{DBLP:conf/acl/MaharanaLTBBF24locomo}和REALTALK\cite{DBLP:journals/corr/abs-2502-13270realtalk}两个数据集的公开数据构造。每一个conversation是由两个鲜明persona的speaker进行的多轮对话，conversation中包含了image sharing behavior，反映了speaker的persona信息，并且具有events之间的记忆和依赖特性。
We employ very long-term multi-session conversations as the data source, utilizing publicly available data from \textsc{LoCoMo}\cite{DBLP:conf/acl/MaharanaLTBBF24locomo} and \textsc{RealTalk}\cite{DBLP:journals/corr/abs-2502-13270realtalk}. Each conversation consists of multi-session dialogs between two distinct persona-driven speakers, incorporating image sharing behaviors, reflecting speaker personas and demonstrating complex temporal dependencies and contextual linkages across events.
An example is illustrated in Figure \ref{fig: dial data source demo}, with detailed statistical analysis presented in Table \ref{tab:statistics of data source of TIME-Dial, LoCoMo-35 and RealTalk}.

\subsection{Dataset Construction\label{sec: Dataset Construction}}
% TODO:这一节需要精确引用附录文字
% NOTE:注意 这里应该尽可能贴着pipeline的图来讲。细节全部对应到附录里面写。
% TODO: 修改思路1————如果要减少文字，同时提高逻辑连贯性，可以考虑修改下面的讲述逻辑，变成：（1）开头先介绍整体的四个大的步骤；（2）接着分不同的数据集如何按照四个大的步骤进行简要描述。
% TODO：修改思路2————将Temporal Facts Collection和Timeline Generation进行合并，统一称之为"Timeline Generation"，
% ----------------------------------
% 我们对每一个data source(\S\ref{sec: Data Source})，分别收集temporal facts，然后提取这些数据集的timeline。基于timeline，我们通过数据合成方法合成QA数据。最后，我们随机选取小批量数据进行人工标注，并得到高质量的\textsc{TimE-Lite}数据集。
As shown in Figure \ref{fig: dataset construction pipeline}, for each data source(\S\ref{sec: Data Source}), we systematically collect temporal facts(\S\ref{sec: Temporal Facts Collection}) and then extract corresponding timelines(\S\ref{sec: Timeline Generation}). Leveraging these timelines, we employ data synthesis methods to generate question-answer pairs(\S\ref{sec: Question-answer Synthesis}). To ensure data quality, we conduct human annotation on a randomly sampled subset, resulting in the high-quality \textsc{TimE-Lite} benchmark(\S\ref{sec: Quality Control}). Details can be seen in Appendix \ref{apdx: Benchmark Construction}.

% \subsubsection{Temporal Facts Collection\label{sec: Temporal Facts Collection}}

% 对于TIME-Wiki，我们使用SLING解析Wikidata数据源中的时间事实四元组，并得到multi-hop temporal knowledge graph(TKG)，细节见\ref{sec:TIME-Wiki TKG construction}. TKG中的每一个时间事实析可以被形式化为($S{i,j}$$,$$R_{i+1,k}, $T_{i+1, k}$, $O_{i+1,k}$$。特别地，我们选择其中的部分entity作为link entity，如图\ref{fig: dataset construction pipeline}中的$S_{0}$, $S_{11}$, $S_{12}$均为link entity，我们以这些link entities作为中心，收集与它们相关的所有时间事实，构成TKG。由此得到的TKG将是一个互相关联的完整故事。
% 对于TIME-News，在\cite{DBLP:conf/acl/Zhang00MLC24TCELongbench}所提供的数据中已经包含了高质量的time points，我们将沿用这部分temporal facts用于后续步骤。
% 对于TIME-Dial，我们使用LLM summarize event graph，得到每一个session的若干个时间事实。每个些时间事由括了speaker的原子事实以及他们的时间戳共同组成。由于原始conversation中存在大量的隐式时间表达式，我们使用LLM提取隐式时间（如"two days ago"）和显式时间（如"8:35 am, Feb 23, 2022"），并进行标准化。

% 针对TIME-Wiki，我们采用SLING框架解析Wikidata数据源，提取时间事实四元组并构建multi-hop temporal knowledge graph (TKG)，具体构建过程详见\ref{sec:TIME-Wiki TKG construction}。TKG中的每个时间事实可形式化为($S{i,j}$$,$$R_{i+1,k}, $T_{i+1, k}$, $O_{i+1,k}$)。特别地，我们选取部分entity作为link entity（如图\ref{fig: dataset construction pipeline}中的$S_{0}$, $S_{11}$, $S_{12}$），并以这些link entities为核心，收集其相关时间事实，构建出相互关联的TKG。
% 对于TIME-News，我们直接采用\cite{DBLP:conf/acl/Zhang00MLC24TCELongbench}提供的高质量time points作为temporal facts，用于后续处理流程。
% 在TIME-Dial的处理中，我们利用LLM对event graph进行summarize，提取每个session的多个时间事实。这些时间事实由speaker的原子事实及其对应时间戳组成。考虑到原始conversation中存在大量隐式时间表达式（如"two days ago"）和显式时间表达式（如"8:35 am, Feb 23, 2022"），我们通过LLM和人工验证进行统一提取和标准化处理。
\noindent \textbf{Temporal Facts Collection\label{sec: Temporal Facts Collection}}
For \textsc{TimE-Wiki}, we employ the SLING framework to parse Wikidata, extracting temporal fact quadruples and constructing a multi-hop temporal knowledge graph (TKG), with detailed methodology in Appendix \ref{sec:TIME-Wiki TKG construction}, where each temporal fact is formalized as ($S_{i,j}$,$R_{i+1,k}$, $T_{i+1, k}$, $O_{i+1,k}$) and link entities (e.g., $S_{0}$, $S_{11}$, $S_{12}$ in Figure \ref{fig: dataset construction pipeline}) serve as central nodes for collecting related temporal facts; for \textsc{TimE-News}, we directly utilize high-quality time points from \cite{DBLP:conf/acl/Zhang00MLC24TCELongbench} as temporal facts; in \textsc{TimE-Dial}, we leverage LLMs to summarize event graphs, extracting multiple temporal facts per session comprising speaker atomic facts and corresponding timestamps, while standardizing both implicit (e.g., ``\textit{two days ago}'') and explicit (e.g., ``\textit{8:35 am, Feb 23, 2022}'') temporal expressions through LLM processing and manual verification.

% \subsubsection{Timeline Generation\label{sec: Timeline Generation}}

% \texttt{order reasoning}, \texttt{relative reasoning}, \texttt{co-temporality}等子任务的QA synthesis涉及到对同一个实体或者同一组事件的时间线精准梳理。为此，我们整理了三个数据集的timelines。具体而言，
% 对于\textsc{TimE-Wiki}，我们对TKG中的link entity按照时间先后顺序梳理；而对于\textsc{TimE-Dial}，我们分别对conversation中的两个speaker按照时间先后顺序梳理temporal facts。对于\textsc{TimE-News}，我们直接使用现成的每一个temporal complex event的Time points作为Timeline，如图\ref{fig: example_TCE}所示。
\noindent \textbf{Timeline Generation\label{sec: Timeline Generation}}
The synthesis of question-answer pairs for subtasks such as \texttt{order reasoning}, \texttt{relative reasoning}, and \texttt{co-temporality} necessitates precise organization of timelines for individual entities or event groups. To this end, we systematically construct timelines across all three datasets. Specifically, for \textsc{TimE-Wiki}, we chronologically organize the temporal facts associated with link entities in the TKG. For \textsc{TimE-Dial}, we separately arrange the temporal facts for each speaker in the conversation in chronological order. In the case of \textsc{TimE-News}, we directly utilize the pre-existing time points from each temporal complex event as the timeline, as illustrated in Figure \ref{fig: example_TCE}.

\noindent \textbf{Context Collection}
We devise distinct context processing strategies tailored to each dataset's characteristics. For \textsc{TimE-Wiki}, we first aggregate temporal facts centered around link entities, reformulating them into coherent narrative segments, which are then integrated into natural and fluent contexts using \texttt{DeepSeek-V3}. For \textsc{TimE-News}, given that the average token count per TCE exceeds 500,000, as shown in Figure \ref{tab:tce_statistics}, rendering full-text evaluation impractical, we employ Retrieval-Augmented Generation (RAG) to extract the most relevant text segments from associated articles as context, with implementation details elaborated in (\S\ref{sec:experiment setup}). For \textsc{TimE-Dial}, we directly utilize the original conversations as context to preserve their integrity and authenticity.

% 我们针对各任务特性，为每一个子任务设计了不同的QA合成pipeline，并同时结合规则和\texttt{DeepSeek-V3}与\texttt{DeepSeek-R1}来生成QA，如表格\ref{tab: Overview of LLM Utilization Strategies in Question Answering (QA) Construction}所示。
% 具体而言，对于\textsc{TimE-Wiki}和\textsc{TimE-Dial}中的子任务而言，我们主要使用了规则构建方法，基于时间线构建question和gold answer的逻辑关系。然后，我们使用LLM来做question phrasing。
% 对于\textsc{TimE-News}而言，每次构造前首先使用已有的Time Points，以相同的entity为中心，使用\texttt{DeepSeek-V3}合成为与类似timeline特征的context，然后利用这些context再来prompt LLM合成QA，如图\ref{fig: a demo for TIME News synthesize QAs}所示。由于经过工作\cite{DBLP:conf/acl/Zhang00MLC24TCELongbench}的outline生成和质量控制，我们使用的time points已经具有较为明确的时间线和逻辑关系。所以我们参考这篇工作构造QA的方式，直接将合成后的context作为prompt的一部分，用于合成高质量的QA。具体的细节如Appendix \ref{sec: appendix_qa_synthesis}所示。

\noindent \textbf{QA Synthesis and Formats\label{sec: Question-answer Synthesis}}
We design distinct QA synthesis pipelines tailored to each subtask's characteristics, combining rule-based templates with \texttt{DeepSeek-V3} and \texttt{DeepSeek-R1} models, as detailed in Table \ref{tab: Overview of LLM Utilization Strategies in Question Answering (QA) Construction}. For \textsc{TimE-Wiki} and \textsc{TimE-Dial}, we employ a rule-based approach grounded in timeline construction to establish logical relationships between questions and gold answers, followed by LLM-based natural language refinement. In \textsc{TimE-News}, we leverage existing Time Points to generate timeline-aligned contexts centered on specific entities using \texttt{DeepSeek-V3}, which then serve as prompts for LLM-driven QA generation, as illustrated in Figure \ref{fig: a demo for TIME News synthesize QAs}. Building upon the framework proposed in \cite{DBLP:conf/acl/Zhang00MLC24TCELongbench}, our Time Points exhibit well-defined timelines and logical structures, enabling us to directly incorporate these contexts into prompts for high-quality QA synthesis. Implementation details are provided in Appendix \ref{sec: appendix_qa_synthesis}.

We design three primary question formats for different subtasks: free-form, single-choice, and multiple-choice, with their distribution detailed in Table \ref{tab: Overview of Question Answering formats}. However, for \textsc{TimE-News} and \textsc{TimE-Dial}, the free-form format's standard answers often exhibit numerous synonymous expressions, leading to significant calibration errors in direct evaluation. To enhance evaluation accuracy while increasing data diversity\cite{DBLP:conf/acl/CeginPSSBB24misleadingoptionmotivation}, we adopt the STARC framework\cite{DBLP:conf/acl/BerzakML20misleadingqageneration} to synthesize multiple-choice questions with misleading options, as illustrated in Figure \ref{fig: misleading option generation}. The detailed methodology for generating misleading options is presented in Appendix \ref{sec: appendix_misleading_options_generation}.

% The question formats vary across subtasks, primarily encompassing free-form, single-choice, and multiple-choice types, as detailed in Table \ref{tab: Overview of Question Answering (QA) formats across datasetups and task categories.}). However, for \textsc{TimE-News} and \textsc{TimE-Dial}, the free-form format's standard answers often exhibit numerous synonymous expressions, leading to significant calibration errors in direct evaluation. To enhance evaluation accuracy while increasing data diversity\cite{DBLP:conf/acl/CeginPSSBB24misleadingoptionmotivation}, we adopt the STARC framework\cite{DBLP:conf/acl/BerzakML20misleadingqageneration} to synthesize multiple-choice questions with misleading options, as illustrated in Figure \ref{fig: misleading option generation}. The detailed methodology for generating misleading options is presented in Appendix \ref{sec: appendix_misleading_options_generation}.

% \subsubsection{Quality Control\label{sec: Quality Control}}

% 动机：尽管我们主要借助规则合成数据，但LLM的参与使得质量控制变得必要。

% 人工标注部分的写作可以参考：https://arxiv.org/abs/2310.06498
% 描述如何设计原则来进行人工评估、标注
% 1. 若干个标注原则
% 2. 标注人员背景
% 3. 标注平台界面（将其转换成全英文的界面）

% 鉴于数据集规模较大，我们采用分层抽样策略，从每个子数据集的QA任务中随机抽取30个样本作为代表性评估集。为确保数据质量，我们首先进行了初步人工筛选，剔除了明显不符合标注标准的样本。具体抽样方法和随机种子设置详见附录。
% 我们组建了由8名专业标注员组成的团队，并基于标注效率和质量双重指标，选取表现最优的3位标注员的标注结果作为最终人工评估基准。

\noindent \textbf{Quality Control\label{sec: Quality Control}}
To validate the quality of our synthesized data and establish a high-quality subset for evaluation, we conducted a comprehensive manual annotation process. For data sampling, we employed a systematic approach to ensure representativeness across all sub-datasets. Using a fixed random seed of 42, we randomly sampled 30-40 QA pairs from each task within the three sub-datasets (\textsc{TimE-Wiki}, \textsc{TimE-News}, and \textsc{TimE-Dial}). This process yielded a total of 1,071 QA pairs for annotation, with 352 from \textsc{TimE-Wiki}, 359 from \textsc{TimE-News}, and 360 from \textsc{TimE-Dial}. To ensure annotation quality, we recruited annotators through professional forums and conducted rigorous qualification tests. From the initial pool of 8 professional annotators, we selected the top 3 performers based on both efficiency and quality metrics to establish our final human evaluation benchmark. We propose Word-level Similarity as a novel metric for evaluating annotation consistency, achieving a score of 0.6626, which demonstrates the high reliability of our annotated data. Details can be seen in Appendix \ref{sec: quality control}.
% 我们设计了Word-level Similarity作为人工标注一致性新指标，达到0.6626，证明了the high reliability of our annotated data. 
% This rigorous selection process resulted in an annotation consistency (Word-level Similarity) score of 0.6626, demonstrating the high reliability of our annotated data. 

% Given the substantial scale of our dataset, we employed a sampling strategy, randomly selecting some QA pairs from each subset as representative evaluation samples. To ensure data quality, we conducted an initial manual screening to remove samples that significantly deviated from annotation standards. 
% We assembled a team of 8 professional annotators and selected the annotation results from the top 3 performers, based on both efficiency and quality metrics, as our final human evaluation benchmark.
% 最终达到了xxx的准确率。

\begin{table}[t] % You can adjust placement specifiers [htbp] as needed (here, top, bottom, page)
  \centering
  \small% Center the table horizontally
  \caption{Stats of \dataset and \textsc{TimE-Lite}. \# QA indicates the number of question-answer pairs.} % Add a descriptive caption
  \label{tab:dataset_statistics_short} % Add a label for cross-referencing in your text (e.g., \ref{tab:dataset_statistics})
  \begin{tabular}{l r l r} % Define column alignment: l=left, r=right
    \toprule % Top rule from booktabs
    Dataset & \# QA & Dataset & \# QA \\
    \midrule % Middle rule from booktabs
    \textbf{\textsc{TimE}}       & \textbf{38,522} & \textbf{\textsc{TimE-Lite}}       & \textbf{943} \\ % Bold the main dataset names and totals
    \hspace*{1em}\textsc{TimE-Wiki} & 13,848 & \hspace*{1em}\textsc{TimE-Lite-Wiki} & 322 \\ % Indent sub-datasets using \hspace*
    \hspace*{1em}\textsc{TimE-News} & 19,958 & \hspace*{1em}\textsc{TimE-Lite-News} & 299 \\
    \hspace*{1em}\textsc{TimE-Dial} &  4,716 & \hspace*{1em}\textsc{TimE-Lite-Dial} & 322 \\
    \bottomrule % Bottom rule from booktabs
  \end{tabular}
  % \vspace{-mm}
\end{table}

\subsection{\textsc{TimE-Lite}\label{sec: TIME-Lite}}

% 为支持高效、可靠的时间推理能力评估，我们构建了轻量级\textsc{TimE-Lite}数据集。该数据集基于\ref{sec: Quality Control}中采样的人工标注子集，经过多轮专家审查与答案验证，最终精选945条高质量QA对。所有数据均经过严格人工校验，确保了评估结果的准确性与可靠性。

To facilitate efficient and reliable evaluation of temporal reasoning capabilities, we introduce \textsc{TimE-Lite}, a lightweight dataset derived from the manually annotated subset sampled in \S\ref{sec: Quality Control}. Through multiple rounds of expert review and answer verification, we curated 945 high-quality QA pairs, all of which have undergone rigorous manual validation to ensure assessment accuracy and reliability.

% \noindent \textbf{Compare \textsc{TimE-Lite} and \textsc{TimE}}
% 说明TIME-Lite作为从TIME中随机抽取的子数据集，修改比例很小，说明TIME-Lite和TIME的偏差很小。

To assess the quality of our synthesized data, we conducted a systematic comparison between the sampled data from \S\ref{sec: Quality Control} and its manually reviewed counterpart (\textsc{TimE-Lite}). The analysis revealed a high consistency rate of 89.13\%, demonstrating the reliability of our data synthesis pipeline.

% 为了定量评估合成数据的质量，我们系统性地比较了\S\ref{sec: Quality Control}中的采样数据与其人工审查后的版本（\textsc{TimE-Lite}）。分析结果显示两者间的一致性达到89.13%，充分证明了我们数据合成流程的可靠性。

% \subsection{Dataset Statistics\label{sec: Dataset Statistics}}
% 用一个表格展现这个数据集的相关数据（需要包含Lite版本），表格应该插入在右下角，类似于omni-math那样

% \begin{table}[t] % You can adjust placement specifiers [htbp] as needed (here, top, bottom, page)
%   \centering       % Center the table horizontally
%   \caption{Stats of \dataset and \textsc{TimE-Lite}. \# QA indicates the number of question-answer pairs.} % Add a descriptive caption
%   \label{tab:dataset_statistics_short} % Add a label for cross-referencing in your text (e.g., \ref{tab:dataset_statistics})
%   \begin{tabular}{l r l r} % Define column alignment: l=left, r=right
%     \toprule % Top rule from booktabs
%     Dataset & \# QA & Dataset & \# QA \\
%     \midrule % Middle rule from booktabs
%     \textbf{\textsc{TimE}}       & \textbf{38,522} & \textbf{\textsc{TimE-Lite}}       & \textbf{943} \\ % Bold the main dataset names and totals
%     \hspace*{1em}\textsc{TimE-Wiki} & 13,848 & \hspace*{1em}\textsc{TimE-Lite-Wiki} & 322 \\ % Indent sub-datasets using \hspace*
%     \hspace*{1em}\textsc{TimE-News} & 19,958 & \hspace*{1em}\textsc{TimE-Lite-News} & 299 \\
%     \hspace*{1em}\textsc{TimE-Dial} &  4,716 & \hspace*{1em}\textsc{TimE-Lite-Dial} & 322 \\
%     \bottomrule % Bottom rule from booktabs
%   \end{tabular}
% \end{table}

% % 4_experiments
\section{Evaluation \label{sec: experiment}}

% 实验设定
\subsection{Experimental Setup\label{sec:experiment setup}}

% 超参数：（可以参考TimeBench之类的写法）
% 1.greedy search
% 检索器：对于三种检索器设定的分析。
% 1.BM25
% 2.Vector (BGE-3)
% 3.Hybrid（with reranker）：怎么hybrid的，在附录里简单说一下。
% 对于TIME-News而言，没有使用Extract任务进行评估。
% 分别评测了哪些模型？
% ---------------------------------------

% 我们在共计24个模型上面对TIME和TIME-Lite进行评估(细节见Appendix \ref{sec: appendix_eval_models})。对于所有模型，我们使用了greedy search的解码策略。对于\textsc{TimE-News}，我们使用了RAG设定，分别使用BM25, Vector, Hybrid三种检索设定对LLM进行评估（Appendix \ref{sec: appendix_retriever}），由于RAG设定下\texttt{Extract}任务无法有效评估模型检索时间的能力，因此我们未做这个任务的评估。我们分别对不同的任务和评估设定设计了不同的评估prompt模板（Appendix \ref{sec: appendix_evaluation_templates}），评估时，我们将QA和对应的context作为input。

% 我们在24个模型上对TIME和TIME-Lite数据集进行了全面评估（模型细节见附录\ref{sec: appendix_eval_models}）。所有实验均采用greedy search解码策略。针对\textsc{TimE-News}数据集，我们采用RAG框架，分别使用BM25、Vector和Hybrid三种检索策略进行评估（详见附录\ref{sec: appendix_retriever}）。考虑到RAG框架下\texttt{Extract}任务难以有效评估时间检索能力，我们未将其纳入评估范围。我们为不同任务和评估场景设计了专门的prompt模板（附录\ref{sec: appendix_evaluation_templates}），在评估过程中，我们将问题-答案对及其相关上下文作为模型输入。
\paragraph{Settings}
We conducted comprehensive evaluations on the \textsc{TimE} and \textsc{TimE-Lite} datasets across 24 models (see Appendix \ref{sec: appendix_eval_models} for model details). All experiments employed greedy search decoding strategy. For the \textsc{TimE-News} and \textsc{TimE-Lite-News} dataset, we implemented the retrieval augmented generation (RAG) framework with three retrieval strategies: BM25, Vector, and Hybrid (detailed in Appendix \ref{sec: appendix_retriever}). Given the \texttt{Extract} task's limited effectiveness in assessing temporal retrieval capabilities under the RAG framework, we excluded it from our evaluation. 
% We designed specialized prompt templates for different tasks and evaluation scenarios (Appendix \ref{sec: appendix_evaluation_templates}), where we provided question-answer pairs along with their relevant contexts as model inputs.

% 1.每一个任务的指标：（1）怎么设计的？（2）设计的合理性？（多选题和填空题的一致性，可以选择部分task来做分析说明）（主要在附录说明）
% ----------------------------------------
\paragraph{Metrics}
We evaluate free-form QA tasks with token-level metrics: Exact Match (EM) for the \texttt{Timeline} task, and F1 score for other free-form QA tasks. For single-choice and multiple-choice QA tasks, we use option-level F1 scores, emphasizing macro F1 for a comprehensive evaluation across all options.
Details are shown in Appendix \ref{sec: appendix_eval_metrics}.

   \begin{table*}[t]
      \centering
      \caption{
      Results for \textsc{TimE-Wiki}. Abbreviations: Ext.L \texttt{Extract}, Loc.:\texttt{Localization}, Comp.: \texttt{Computation}, DC.: \texttt{Duration Compare}, OC.: \texttt{Order Compare} ER.: \texttt{Explicit Reasoning}, OR.: \texttt{Order Reasoning}, RR.: \texttt{Relative Reasoning}, Co-tmp.: \texttt{Co-temporality}, TL.: \texttt{Timeline} and CTF.: \texttt{Counterfactual}.
      % Local. denotes \texttt{Localization}. Comp. denotes \texttt{Computation}. Dur. Comp. denotes \texttt{Duration Computation}. Expl. Reason. denotes \texttt{Explicit Reasoning}. Ord. Reason. denotes \texttt{Order Reasoning}. Rel. Reason. denotes \texttt{Relative Reasoning}. Co-temp. denotes \texttt{Co-temporality}. Counterf. denotes \texttt{Counterfactual}.
      % 模型等的解释
      Top-1 result for each blank are \textbf{bold}.
      }
      \label{tab: exp on TIME-Wiki for main text}
      \resizebox{\textwidth}{!}{% Resize table to fit within text width
\begin{tabular}{@{}lccccccccccc@{}}
\toprule
\multirow{2}{*}{\textbf{Model}} & \multicolumn{5}{c}{\textbf{Level 1}} & \multicolumn{3}{c}{\textbf{Level-2}} & \multicolumn{3}{c}{\textbf{Level-3}} \\
\cmidrule(lr){2-6} \cmidrule(lr){7-9} \cmidrule(lr){10-12}
& \textbf{Ext.} & \textbf{Loc.} & \textbf{Comp.} & \textbf{DC.} & \textbf{OC.} & \textbf{ER.} & \textbf{OR.} & \textbf{RR.} & \textbf{Co-tmp.} & \textbf{TL.} & \textbf{CTF.} \\
\midrule
\multicolumn{12}{c}{\textit{Non-reasoning Models} (\textsc{TimE-Wiki})}\\
\midrule
Llama-3.1-8B-Instruct & 53.16 & 75.41 & 9.79 & 50.89 & 65.49 & 28.96 & 31.72 & 24.53 & 31.36 & 0.92 & 28.60 \\
Qwen2.5-7B & 35.33 & 67.19 & 24.22 & 23.73 & 65.36 & 10.11 & 14.66 & 5.45 & 2.45 & 0.00 & 0.98 \\
Qwen2.5-14B & 33.58 & 71.26 & 20.53 & 50.64 & 66.49 & 7.42 & 15.37 & 20.68 & 17.11 & 0.00 & 27.95 \\
Qwen2.5-7B-Instruct & 57.58 & 65.30 & 32.34 & 52.22 & 68.75 & 44.76 & 35.48 & 26.79 & 36.68 & 1.08 & 38.42 \\
Qwen2.5-14B-Instruct & 71.02 & 74.49 & 26.37 & 63.50 & 82.76 & 52.93 & 38.94 & 30.34 & 33.68 & 2.62 & 43.16 \\
Qwen2.5-72B-Instruct & \textbf{81.70} & \textbf{83.84} & \textbf{41.37} & \textbf{66.64} & \textbf{84.22} & \textbf{70.13} & \textbf{44.84} & \textbf{35.23} & \textbf{51.17} & \textbf{4.08} & \textbf{50.68} \\
% Qwen2.5-7B-Instruct-1M & 47.43 & 64.56 & 38.34 & 37.19 & 66.51 & 42.21 & 36.53 & 27.23 & 42.57 & 0.69 & 38.75 \\
% Qwen2.5-14B-Instruct-1M & 54.00 & 79.82 & 32.61 & 64.06 & 78.62 & 58.69 & 40.86 & 28.58 & 34.56 & 3.46 & 41.79 \\
\midrule
% \multicolumn{12}{c}{\textit{Test-time Scaled Models} (\textsc{TimE-Wiki})}\\
\multicolumn{12}{c}{\textit{Reasoning Models} (\textsc{TimE-Wiki})}\\
\midrule
Deepseek-R1-Distill-Llama-8B & 66.75 & \textbf{68.82} & \textbf{57.27} & 83.47 & 90.22 & 51.17 & 37.36 & 32.41 & 31.04 & 5.31 & 37.30 \\
Deepseek-R1-Distill-Qwen-7B & 54.89 & 65.04 & 56.63 & 77.85 & 85.71 & 48.88 & 32.53 & 30.57 & 29.74 & 0.54 & 37.38 \\
Deepseek-R1-Distill-Qwen-14B & 67.66 & 66.33 & 51.25 & 81.21 & 92.97 & 58.94 & \textbf{43.49} & 35.63 & 36.30 & 14.54 & 45.69 \\
% Deepseek-R1-Distill-Llama-70B & 74.38 & 70.21 & 73.35 & 88.54 & 93.61 & 65.94 & 45.54 & 38.83 & 43.10 & 21.69 & 45.97 \\
QwQ-32B & \textbf{74.99} & 67.75 & 49.59 & \textbf{88.20} & \textbf{93.53} & \textbf{60.61} & 37.77 & \textbf{36.39} & \textbf{37.76} & \textbf{25.38} & \textbf{53.13} \\
\midrule
\multicolumn{12}{c}{\textit{Advanced Models} (\textsc{TimE-Lite-Wiki})}\\
\midrule
Deepseek-V3 & 93.33 & \textbf{84.51} & 23.76 & 71.43 & 83.33 & 75.69 & 39.77 & 41.76 & 46.62 & 10.00 & 44.82 \\
Deepseek-R1 & 96.67 & 77.61 & 46.39 & 89.29 & \textbf{93.33} & 78.20 & \textbf{57.09} & \textbf{57.79} & 47.45 & \textbf{33.33} & \textbf{55.71} \\
GPT-4o & \textbf{98.89} & 83.24 & 33.82 & 67.86 & 90.00 & 80.68 & 45.83 & 46.56 & 45.45 & 20.00 & 50.72 \\
OpenAI o3-mini & 96.67 & 80.83 & \textbf{49.17} & \textbf{92.86} & \textbf{93.33} & \textbf{82.24} & 52.62 & 48.98 & \textbf{54.34} & \textbf{33.33} & 52.07 \\
\bottomrule
\end{tabular}
      }
      \vspace{-3mm}
   \end{table*}

\subsection{Result}
\subsubsection{Real-world Scenario Analysis}
% 每一个paragraph凝练一个结论，比如第一个就是Knowledge Intensive
\paragraph{Knowledge intensive events makes it challenging for capturing complex temporal expression and relationship.}
% 知识密集的事件对于理解复杂的时间表达式和时间关系更具有挑战性：主要对比Level-2和Level-3
% 如表\ref{tab: exp on TIME-Wiki for main text}所示，理解隐式的时间表达式和事件之间的内在时间关联更具有挑战性，例如o3-mini分别在（如\texttt{Order Reasoning}和\texttt{Relative Reasoning}）上仅有52.62\%和48.98\%的性能，而在\textt{Co-temporality}任务中，也仅有54.34\%的性能。相比之下，它在基本的检索和理解时间能力(Level-1)上，几乎达到了80\%以上的性能。这说明知识密集型的场景中，时间与实体关联更复杂多样，阻碍了模型关联时间和事实的能力。

% 知识密集的事件对于理解复杂的时间表达式和时间关系更具有挑战性：主要对比Level-2和Level-3

% 如表\ref{tab: exp on TIME-Wiki for main text}所示，模型在处理隐式时间表达式和事件间内在时间关联时面临显著挑战。以o3-mini模型为例，其在\texttt{Order Reasoning}和\texttt{Relative Reasoning}任务上的表现分别仅为52.62\%和48.98\%，在\texttt{Co-temporality}任务中也仅达到54.34\%。相比之下，该模型在基础时间检索和理解任务(Level-1)上的表现接近80\%。这一现象表明，在知识密集型场景中，时间与实体间复杂多样的关联关系显著增加了模型准确关联时间与事实的难度。

As shown in Table \ref{tab: exp on TIME-Wiki for main text}, models face significant challenges in comprehending implicit temporal expressions and intrinsic temporal relationships between events. For o3-mini, it achieves only 52.62\% and 48.98\% on \texttt{Order Reasoning} and \texttt{Relative Reasoning} tasks respectively, and merely 54.34\% on the \texttt{Co-temporality} task. In contrast, its performance on basic temporal retrieval and comprehension tasks (Level-1) approaches 80\% for 4 tasks. This phenomenon suggests that the complex and diverse associations between temporal information and entities in knowledge-intensive scenarios substantially hinder models' ability to accurately correlate time with facts.

\paragraph{Complex dynamic events constrain models' ability to comprehend basic temporal relationship and construct coherent timelines.}
% Table \ref{tab: exp on TIME-News for main text}所示，模型在理解时间间隔、时间顺序等基本能力，以及对时间线进行梳理等方面，具有较大的挑战性。例如，推理模型o3-mini在\texttt{Duration Compare}和\texttt{Order Compare}任务上仅能达到最高63.33\%的性能。所有模型在三个事件的排序问题\texttt{Timeline}中只能达到不高于30\%的性能。这说明，复杂事件之间的细节使得模型根据问题定位到多个相似但不确切的时间点，进而产生错误。
As shown in Table \ref{tab: exp on TIME-News for main text} (\textsc{TimE-News}), models is challenged by comprehending fundamental temporal relationships, including time intervals and ordering, as well as constructing coherent timelines. For instance, the reasoning model o3-mini achieves a maximum performance of only 63.33\% on both \texttt{Duration Compare} and \texttt{Order Compare} tasks. Notably, all models demonstrate limited capability in the \texttt{Timeline} task, which requires ordering three events, with performance not exceeding 30\%. This suggests that the intricate details among complex events lead models to identify multiple similar but imprecise temporal points, resulting in erroneous predictions.

% 第三个是在长上下文中对话
% Very-long multi-session dialog削弱了模型的时间检索和定位性能。
% 从Extract, Localization, timeline等任务中去阐述。
% 从very-long和日常对话交互情景中的隐式时间表达两个角度去解释。
\paragraph{Very-long multi-session dialog impairs the capability of time retrieval and event-time localization.}
% 在超长日常对话情景中，模型难以直接定位和准确检索时间信息。如Table \ref{tab: exp on TIME-Dial for main text}所示，开源的vanilla和test-time scaled模型在\texttt{Extract}和\texttt{Localization}任务上仅有最高约40\%的性能，大幅度低于在其他数据集的相同任务上的表现。这可以归因于超长的对话（平均超过15k tokens）、多轮对话使得模型定位更加困难；在日常交互场景中存在多个事件的记忆性表达（例如speaker说"Last saturday, I came up with a familiar."），使得模型准确定位对话中所提到的事件更加困难。

% 在超长多轮对话场景中，模型的时间信息定位与检索能力显著下降。如表\ref{tab: exp on TIME-Dial for main text}所示，开源vanilla模型和test-time scaled模型在\texttt{Extract}和\texttt{Localization}任务上的最高准确率仅为40\%，远低于其他数据集上的表现。这一现象主要源于两个因素：首先，超长对话上下文（平均超过15k tokens）和多轮交互显著增加了时间定位的难度；其次，日常对话中频繁出现的记忆性时间表达（如"上周六我遇到了一个熟人"）进一步加剧了模型准确识别事件时间点的挑战。

As shown in Table \ref{tab: exp on TIME-Dial for main text} (\textsc{TimE-Dial}), the maximum accuracy of open-source vanilla models and test-time scaled models on \texttt{Extract} and \texttt{Localization} tasks is merely 40\%, substantially lower than their performance on other datasets. This phenomenon can be attributed to two primary factors: first, the extensive dialog context (averaging over 15k tokens, as shown in Table \ref{tab:statistics of data source of TIME-Dial, LoCoMo-35 and RealTalk}) and multi-turn interactions significantly increase the difficulty of temporal localization; second, the frequent use of memory-based temporal expressions in daily dialogs (e.g., "Last Saturday"), which necessitate reasoning with the conversation timestamp to pinpoint the precise date, further hinders accurate timestamp identification.

% 2.不同level，不同任务之间的依赖关系？
\subsubsection{Temporal Reasoning Tasks Analysis}

\begin{table*}[t]
 \centering
 \caption{
 Results for \textsc{TimE-News}.
 Top-3 articles are retrieved. Abbreviations follow Table \ref{tab: exp on TIME-Wiki for main text}.
 }
 \label{tab: exp on TIME-News for main text}
 \resizebox{\textwidth}{!}{% Resize table to fit within text width
\begin{tabular}{@{}llcccccccccc@{}} % Original was 13 columns (ll + 11c), now 12 (ll + 10c)
\toprule
 \multirow{2}{*}{\textbf{Model}} & \multirow{2}{*}{\textbf{Retriever}}& \multicolumn{4}{c}{\textbf{Level 1}} & \multicolumn{3}{c}{\textbf{Level-2}} & \multicolumn{3}{c}{\textbf{Level-3}} \\
 \cmidrule(lr){3-6} \cmidrule(lr){7-9} \cmidrule(lr){10-12}
 & & \textbf{Loc.} & \textbf{Comp.} & \textbf{DC.} & \textbf{OC.} & \textbf{ER.} & \textbf{OR.} & \textbf{RR.} & \textbf{Co-tmp.} & \textbf{TL.} & \textbf{CTF.} \\
\midrule
% \multicolumn{12}{c}{\textit{Vanilla Models} (\textsc{TimE-News})} \\
\multicolumn{12}{c}{\textit{Non-reasoning Models} (\textsc{TimE-News})} \\
\midrule
\multirow{3}{*}{Llama3.1-8B-Instruct}&BM25&47.96 &27.12 &39.06 &39.28 &81.72& 66.67& 77.06& 80.50&3.09 &47.17 \\
& Vector&50.99 &32.13 &40.94 &41.17 &81.33& 67.67& 77.67& 81.50&1.94 &46.22\\
& Hybrid&51.81 &34.51 &41.78 &44.11 &82.50& 68.94& 78.39& 82.89&2.55 &46.44\\
% \midrule % Added cline after Llama3.1
\midrule
\multirow{3}{*}{Qwen2.5-14B-Instruct}&BM25&68.53 &70.80 &42.39 &46.17 &83.06& 70.44& 79.61& 82.67&26.13 &59.39 \\
&Vector&71.68 &76.28 &42.22 &45.67 &83.94& 69.33& 80.33& 83.44&23.68 &\textbf{59.67} \\
&Hybrid&71.00 &79.75 &43.61 &48.72 &84.72& 70.39& 81.44& 84.06&26.61 &58.61 \\
\midrule % Added cline after Qwen2.5-14B
\multirow{3}{*}{Qwen2.5-32B-Instruct}&BM25&68.88 &79.48 &\textbf{46.44} &51.22 &84.39& 70.78& 81.56& 85.11&\textbf{27.54} &54.61 \\
&Vector&\textbf{71.76} &84.46 &44.78 &50.61 &85.22& 70.94& 82.11& 84.39&24.16 &55.83 \\
&Hybrid&71.57 &\textbf{86.62} &44.78 &\textbf{54.83} & \textbf{86.28}& \textbf{71.17}& \textbf{82.72}& \textbf{86.11}&25.92 &54.06 \\
\midrule % Existing midrule, no cline added
% \multicolumn{12}{c}{\textit{Test-time Scaled Models} (\textsc{TimE-News})} \\
\multicolumn{12}{c}{\textit{Reasoning Models} (\textsc{TimE-News})} \\
\midrule
\multirow{3}{*}{Deepseek-R1-Distill-Qwen-7B} &BM25&39.66 &60.15 &38.78 &53.33 &76.28& 60.06& 70.17& 74.56&17.94 &37.11 \\
&Vector&41.17 &59.81 &\textbf{41.72} &54.56 &76.44& 61.94& 73.89& 74.67&16.44 &38.78 \\
&Hybrid&41.42 &60.28 &38.22 &54.78 &78.22& 62.67& 72.72& 76.39&17.08 &39.06 \\
\midrule % Added cline after Deepseek-R1-Distill-Qwen-7B
\multirow{3}{*}{Deepseek-R1-Distill-Qwen-14B}&BM25&63.42 &62.36 &39.72 &52.61 &83.39& \textbf{70.33}& 80.83& 83.78&\textbf{21.82} &62.72 \\
&Vector&65.96 &63.56 &39.39 &51.33 &84.89& 69.22& 81.28& 83.89&19.58 &\textbf{63.44} \\
&Hybrid&\textbf{66.11} &\textbf{66.29} &39.39 &\textbf{54.94} & \textbf{85.61}& 69.89& \textbf{82.67}& \textbf{85.00}&21.10 &62.00 \\
\midrule % Existing midrule, no cline added
\multicolumn{12}{c}{\textit{Advanced Models} (\textsc{TimE-Lite-News})} \\
\midrule
\multirow{3}{*}{GPT-4o} &BM25&79.26 &10.56 &43.33 &43.33 &76.67& 70.00& \textbf{93.33}& \textbf{93.33}&24.14 &43.33 \\
&Vector&75.56 &15.00 &40.00 &53.33 &\textbf{80.00}& 66.67& 86.67& 90.00&24.14 &40.00 \\
&Hybrid&\textbf{80.56} &\textbf{20.00} &33.33 &46.67 &73.33& 66.67& 86.67& 90.00&13.79 &\textbf{46.67} \\
\midrule % Added cline after GPT-4o
\multirow{3}{*}{OpenAI o3-mini} &BM25&72.59 &12.78 &56.67 &60.00 &73.33& \textbf{83.33}& 86.67& \textbf{93.33}&\textbf{27.59} &33.33 \\
&Vector&76.67 &18.33 &\textbf{63.33} &\textbf{63.33} &\textbf{80.00}& 66.67& 86.67& 80.00&24.14 &33.33 \\
&Hybrid&77.94 &16.67 &56.67 &\textbf{63.33} &76.67& 63.33& 80.00& 86.67&\textbf{27.59} &36.67 \\
\bottomrule
\end{tabular}}
\end{table*}
% 宏观来看，三个层次的挑战各不相同。对于每个数据集，
% \paragraph{}

% 哪些任务之间有明显的相关关系？（可以考虑放一张任务关系图，但是也可以只用主表描述）
% 检索能力(Extract, Localization)对于全部任务的影响。
\paragraph{Time retrieval ability is significantly correlated with almost all aspects of temporal reasoning tasks.}
% 为了probe基本时间检索能力对其他关系的影响，我们通过计算8个vanilla models在\textsc{TimE-Lite}三个子数据集上的\texttt{Extract} and \texttt{Localization}，与全部任务性能的相关系数热力图，Each task can then be represented with a vector of model performance at various temporal reasoning tasks. The 11 task vectors are then clustered via agglomorative clustering algorithm, using correlation coefficient as the distance metric. 我们能够从这个图看出，\texttt{Extract}和\texttt{Localization}任务与几乎其他所有任务的相关性较高，大部分相关度超过0.5。这说明，基本时间检索能力与其他方面的时间推理能力具有强相关性。

% 为探究基本时间检索能力对时间推理任务的影响，我们基于8个vanilla模型在\textsc{TimE-Lite}三个子数据集上的表现，计算了\texttt{Extract}和\texttt{Localization}任务与其他任务性能的相关系数。具体而言，我们将每个任务表示为模型在不同时间推理任务上的性能向量，并采用凝聚层次聚类算法，以相关系数作为距离度量进行聚类分析。实验结果表明，\texttt{Extract}和\texttt{Localization}任务与几乎所有其他任务均呈现显著相关性（相关系数>0.5），这证实了基本时间检索能力与高层次时间推理能力之间存在强关联性。

\begin{figure}
  \centering
  \includegraphics[width=\textwidth]{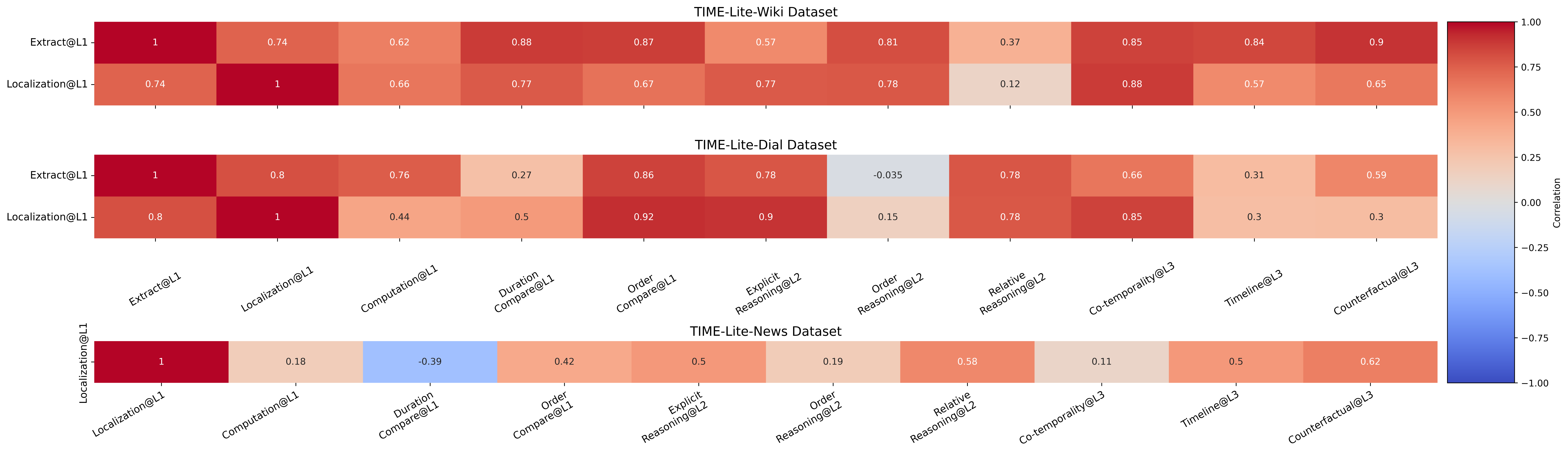}
  
  \caption{
  % Heatmap of \textsc{TimE-Lite-Wiki}, \textsc{TimE-Lite-Dial}, \textsc{TimE-Lite-News}.
  % \texttt{Extract}, \texttt{Localization} 任务与其他任务的相关性热力图。
  Task correlation heatmap highlighting the relationship between \texttt{Extract} and \texttt{Localization} tasks and other temporal reasoning tasks. Note: \texttt{Extract} task is excluded from \textsc{TimE-Lite-News} evaluation.\label{fig: correlation_heatmap}
  }
\end{figure}

To investigate the impact of basic temporal retrieval capabilities on temporal reasoning tasks, we computed correlation coefficients between \texttt{Extract} and \texttt{Localization} tasks and other task performances based on 8 vanilla models across three \textsc{TimE-Lite} subsets. Specifically, we represented each task as a vector of model performance across various temporal reasoning tasks and performed agglomerative clustering using correlation coefficients as the distance metric. The results (shown in Figure \ref{fig: correlation_heatmap}) demonstrate that \texttt{Extract} and \texttt{Localization} tasks exhibit significant correlations (correlation coefficient > 0.5) with nearly all other tasks, confirming a strong relationship between basic temporal retrieval and higher-level temporal reasoning capabilities.

% 具体来看，哪些任务对三个数据集而言，都比较困难，为什么会更困难（case study）？哪些模型在这些数据上表现稍微更有优势。
\paragraph{Grasping timeline over multiple events is much  challenging for long-range contexts.}
% 如Table \ref{tab: exp on TIME-Wiki for main text}, \ref{tab: exp on TIME-News for main text} and \ref{tab: exp on TIME-Dial for main text}所示，在\texttt{Timeline}任务上表现均poorly，尤其是在\textsc{TimE-Wiki}和\textsc{TimE-Dial}数据集上，小规模的Vanill模型性能甚至小于10\%。即使是在\textsc{TimE-News}数据集中，仅对3个事件重排序，也构成了明显的挑战。e\texttt{Timeline}任务是一个综复杂时序信息的间检索和时间先后全面排序的任。它相比于多数模型表现较好的事件先后关系的基本时间概念理解任务（如\texttt{Order Compare}任务），其具备更高的复杂度。

% 如表\ref{tab: exp on TIME-Wiki for main text}, \ref{tab: exp on TIME-News for main text}和\ref{tab: exp on TIME-Dial for main text}所示，所有模型在\texttt{Timeline}任务上的表现均不理想。特别是在\textsc{TimE-Wiki}和\textsc{TimE-Dial}数据集上，小规模Vanilla模型的准确率甚至低于10\%。值得注意的是，即使在相对简单的\textsc{TimE-News}数据集中，仅对3个事件进行时间排序也构成了显著挑战。这主要源于\texttt{Timeline}任务需要同时处理复杂的时间信息检索和全局时间顺序推理，其难度远高于仅需理解两个事件先后关系的\texttt{Order Compare}等基础任务。

As shown in Table \ref{tab: exp on TIME-Wiki for main text}, \ref{tab: exp on TIME-News for main text} and \ref{tab: exp on TIME-Dial for main text}, all models demonstrate suboptimal performance on the \texttt{Timeline} task. Notably, small-scale vanilla models achieve accuracy below 10\% on both \textsc{TimE-Wiki} and \textsc{TimE-Dial} datasets. Even in the relatively simpler \textsc{TimE-News} dataset, merely reordering three events poses a significant challenge. This difficulty stems from the \texttt{Timeline} task's requirement for simultaneous complex temporal information retrieval and global temporal ordering reasoning, which is substantially more challenging than basic tasks like \texttt{Order Compare} that only require understanding the sequence of two events.

% 重点关联任务之间的分析
% 分析两个事件先后和多个事件共同排序的难度差异
% 说明即使模型对两个事件的先后关系捕捉到位，但多个事件的先后顺序捕捉仍然有限。

% 3.Test-time scaling的影响
\subsubsection{Impact of Test-time Scaling}
% test-time scaling对于时间逻辑推理任务是有效的
\paragraph{Test-time scaling benefits temporal logical reasoning.}
% Test-time scaled models主要通过增强思维链来提升反思、验证等逻辑推理能力。为探究test-time scaling对于时间推理的影响，我们对比R1-Distill版本的模型与对应的Vanilla模型在各任务上的性能差异。如Table \ref{tab: exp on TIME-Wiki for main text} and \ref{tab: exp on TIME-Dial for main text}所示。以\texttt{Order Compare}, \texttt{Duration Compare}为代表的等涉及推理的时间理解任务，以及\texttt{Counterfactual}为代表的复杂时间-事件关系处理的任务上，Deepseek-R1-Distill-Qwen-14B均以明显margin优势超过Qwen2.5-14B-Instruct模型，分别在\textsc{TimE-Dial}的这三个任务上具有24.44\%, 11.33\%, 12.0\%的优势。

% Test-time scaling通过增强模型的思维链推理能力，显著提升了其在复杂逻辑推理任务中的表现。为系统评估该方法对时间推理任务的影响，我们对比分析了R1-Distill模型与其对应的Vanilla模型在多个任务上的性能差异（详见表\ref{tab: exp on TIME-Wiki for main text}和\ref{tab: exp on TIME-Dial for main text}）。实验结果表明，在需要时间推理的\texttt{Order Compare}和\texttt{Duration Compare}任务，以及处理复杂时间-事件关系的\texttt{Counterfactual}任务上，Deepseek-R1-Distill-Qwen-14B模型相比Qwen2.5-14B-Instruct展现出显著优势，在\textsc{TimE-Dial}数据集上分别取得了24.44\%, 11.33\%, 12.0\%的性能提升。

Test-time scaling enhances models' performance in complex logical reasoning tasks by strengthening their chain-of-thought capabilities. To systematically evaluate its impact on temporal reasoning, we compare R1-Distill models and their vanilla counterparts across multiple tasks (see Table \ref{tab: exp on TIME-Wiki for main text} and \ref{tab: exp on TIME-Dial for main text}). Experimental results demonstrate that Deepseek-R1-Distill-Qwen-14B significantly outperforms Qwen2.5-14B-Instruct in temporal reasoning tasks such as \texttt{Order Compare} and \texttt{Duration Compare}, as well as in handling complex temporal-event relationships in the \texttt{Counterfactual} task, achieving performance improvements of 24.44\%, 11.33\%, and 12.0\% respectively on the \textsc{TimE-Dial} dataset. 
Our analysis further reveals that advanced test-time scaled models, including o3-mini and Deepseek-R1, consistently outperform their non-reasoning counterparts in logical reasoning-based tasks, demonstrating the effectiveness of test-time scaling in enhancing complex reasoning capabilities.

% test-time scaling对于xxx任务不具有优势【尝试结合case来分析】
\paragraph{Test-time scaling is not consistently effective for time retrieval and event location.}
% \textsc{TimE-Wiki}数据集上，Deepseek-R1-Distill-Qwen-14B相比Qwen2.5-14B-Instruct在\texttt{Extract}和\texttt{Localization}任务上分别性能低了3.36\%和8.16\%。而相比之下，在\textsc{TimE-Dial}数据集上，Deepseek-R1-Distill-Qwen-14B相比Qwen2.5-14B-Instruct在\texttt{Extract}任务上具有1.55\%的优势，而在\texttt{Localization}任务上却减弱了12.49\%。
% test-time scaling models在提取时间时，具有遍历查询context中时间信息的思维过程，对于\textsc{TimE-Dial}这类multi-session对话场景而言更具有优势，但可能会过度思考，在遍历检索时间错误后陷入逻辑上的循环反思，无法纠正自己的错误，如case \ref{box: overthinking for distill-r1-qwen-14b}所示。

% 实验结果表明，test-time scaling模型在不同数据集上的表现存在显著差异。在\textsc{TimE-Wiki}数据集中，Deepseek-R1-Distill-Qwen-14B在\texttt{Extract}和\texttt{Localization}任务上的表现分别比Qwen2.5-14B-Instruct低3.36\%和8.16\%。然而，在\textsc{TimE-Dial}数据集中，该模型在\texttt{Extract}任务上取得了1.55\%的性能提升，但在\texttt{Localization}任务上却下降了12.49\%。这种差异主要源于test-time scaling模型采用的时间信息检索机制：它们通过系统性地遍历上下文来提取时间信息，这种策略在multi-session对话场景中具有优势，但也可能导致模型在检索错误后陷入过度思考的循环，无法及时纠正错误（详见case \ref{box: overthinking for distill-r1-qwen-14b}）。

Experimental results reveal significant performance variations of test-time scaling models across different datasets. On \textsc{TimE-Wiki}(shown in Table \ref{tab: exp on TIME-Wiki for main text}), Deepseek-R1-Distill-Qwen-14B underperforms Qwen2.5-14B-Instruct by 3.36\% and 8.16\% in \texttt{Extract} and \texttt{Localization} tasks respectively. Conversely, on \textsc{TimE-Dial} (shown in \ref{tab: exp on TIME-Dial for main text}), it achieves a 1.55\% improvement in \texttt{Extract} but suffers a 12.49\% decline in \texttt{Localization}. This discrepancy stems from the temporal information retrieval mechanism of test-time scaling models: their systematic context traversal strategy benefits multi-session dialog scenarios but may lead to overthinking cycles after retrieval errors, hindering error correction (see case in Appendix \ref{box: overthinking for distill-r1-qwen-14b}).

% 4.Retriever 的影响
\subsubsection{Impact of Retrievers in \textsc{TimE-News}}
% 检索本身对于寻找事件细节更有优势，而根据时间计算、
% 检索能力：主要针对TIME-News来说，一些任务使用不同的检索器，性能差异比较大；而有些任务，不同规模的模型对于这个任务的性能影响较大。总体来讲，不同检索器对于性能的影响相比于不同的模型更大。
% 对于RAG设定
% 1.不同模型下，性能相似的，说明这个任务检索器影响大
% 2.相同模型，不同检索器性能差异大，说明这个任务检索器影响大
% 3.不同模型（尤其是推理模型相比于非推理模型而言），性能差异大，说明这个任务推理能力影响大

% 检索器选择对时间推理性能具有显著影响。对于相同模型而言，不同检索器下的评估性能差距较大。以GPT-4o为例，在Timeline任务中，Hybrid检索器下的性能较BM25和Vector检索器低10%以上；在\texttt{Order Compare}任务中，不同检索器间的性能差异也达到10%。这一现象表明，准确的时间事实检索是处理真实世界动态信息的关键，直接影响复杂事件推理的效果。另外，对于\texttt{Explicit Reasoning}、\texttt{Order Reasoning}任务，不同的模型在相同检索器设定下的性能差异较小，说明检索器对这些任务上的时间推理影响较大，掩盖掉了模型之间本来的差异。

% 实验结果表明，检索器选择对时间推理任务性能具有显著影响。以GPT-4o为例，在\texttt{Timeline}任务中，Hybrid检索器相比BM25和Vector检索器导致性能下降超过10%；在\texttt{Order Compare}任务中，不同检索器间的性能差异也达到10%。这一发现表明，准确的时间事实检索是处理动态信息的关键，直接影响复杂事件推理的效果。值得注意的是，在\texttt{Explicit Reasoning}和\texttt{Order Reasoning}任务中，相同检索器设定下不同模型的性能差异显著减小，这表明检索器对这些任务的时间推理能力具有主导性影响，甚至掩盖了模型本身的能力差异。

Experimental results (shown in Table \ref{tab: exp on TIME-News for main text}) demonstrate that the choice of retriever significantly impacts temporal reasoning performance. Taking GPT-4o as an example, its performance with the Hybrid retriever is over 10\% lower than with BM25 and Vector retrievers in the \texttt{Timeline} task. Similarly, a 10\% performance gap exists across different retrievers in the \texttt{Order Compare} task. This finding suggests that accurate temporal fact retrieval is crucial for processing dynamic information, directly affecting the effectiveness of complex event reasoning. Notably, in \texttt{Explicit Reasoning} and \texttt{Order Reasoning} tasks, the performance differences among models under the same retriever setting are significantly reduced, indicating that the retriever plays a dominant role in temporal reasoning for these tasks, even overshadowing the inherent capabilities of different models.

   \begin{table*}[t]
      \centering
      \caption{
      Results for \textsc{TimE-Dial}. Abbreviations follow Table \ref{tab: exp on TIME-Wiki for main text}.
      }
      \label{tab: exp on TIME-Dial for main text}
      \resizebox{\textwidth}{!}{% Resize table to fit within text width
\begin{tabular}{@{}lccccccccccc@{}}
    \toprule
    \multirow{2}{*}{\textbf{Model}} & \multicolumn{5}{c}{\textbf{Level 1}} & \multicolumn{3}{c}{\textbf{Level-2}} & \multicolumn{3}{c}{\textbf{Level-3}} \\
    \cmidrule(lr){2-6} \cmidrule(lr){7-9} \cmidrule(lr){10-12}
    & \textbf{Ext.} & \textbf{Loc.} & \textbf{Comp.} & \textbf{DC.} & \textbf{OC.} & \textbf{ER.} & \textbf{OR.} & \textbf{RR.} & \textbf{Co-tmp.} & \textbf{TL} & \textbf{CTF.} \\
    \midrule
    % \multicolumn{12}{c}{\textit{Vanilla Models} (\textsc{TimE-Dial})} \\
    \multicolumn{12}{c}{\textit{Non-reasoning Models} (\textsc{TimE-Dial})} \\
    \midrule
    Llama-3.1-8B-Instruct & 27.45 & \textbf{38.61} & 9.05 & \textbf{48.44} & \textbf{52.67} & 38.22 & \textbf{46.22} & \textbf{57.33} & \textbf{72.00} & 0.00 & 38.00 \\
    Qwen2.5-7B-Instruct & 36.51 & 30.91 & 23.25 & 41.11 & 41.33 & 31.11 & 34.22 & 44.44 & 58.00 & 0.22 & \textbf{46.44} \\
    Qwen2.5-14B-Instruct & 38.85 & 30.83 & 16.35 & 42.00 & 47.78 & 38.22 & 38.67 & 49.11 & 57.33 & 0.00 & 34.89 \\
    Qwen2.5-32B-Instruct & \textbf{40.67} & 33.56 & \textbf{23.45} & 40.89 & \textbf{52.67} & \textbf{43.33} & 36.67 & 46.00 & 63.11 & \textbf{0.67} & 40.44 \\
    \midrule
    % \multicolumn{12}{c}{\textit{Test-time Scaled Models} (\textsc{TimE-Dial})} \\
    \multicolumn{12}{c}{\textit{Reasoning Models} (\textsc{TimE-Dial})} \\
    \midrule
    Deepseek-R1-Distill-Llama-8B & 40.21 & 36.37 & 14.69 & 40.89 & 57.11 & 34.89 & 34.00 & 40.44 & 54.67 & \textbf{0.44} & 42.22 \\
    Deepseek-R1-Distill-Qwen-14B & \textbf{40.40} & 18.34 & 12.98 & 53.33 & 72.22 & 54.67 & 40.44 & 53.33 & 66.89 & 0.22 & 46.89 \\
    Deepseek-R1-Distill-Qwen-32B & 39.28 & \textbf{35.79} & \textbf{22.87} & \textbf{58.22} & \textbf{75.33} & \textbf{57.56} & \textbf{41.78} & \textbf{54.89} & \textbf{72.67} & 0.22 & \textbf{49.78} \\
    \midrule
    \multicolumn{12}{c}{\textit{Advanced Models} (\textsc{TimE-Lite-Dial})} \\
    \midrule
    Deepseek-V3 & 52.63 & 42.67 & 13.00 & 70.00 & 73.33 & 40.00 & 26.67 & 60.00 & 56.67 & 3.33 & 43.33 \\
    Deepseek-R1 & \textbf{65.00} & 48.56 & 22.61 & \textbf{73.33} & \textbf{86.67} & \textbf{76.67} & 53.33 & 66.67 & \textbf{76.67} & \textbf{10.00} & \textbf{53.33} \\
    GPT-4o & 61.08 & \textbf{52.98} & 14.00 & 40.00 & 76.67 & 60.00 & 43.33 & 66.67 & 76.67 & 0.00 & 46.67 \\
    OpenAI o3-mini & 41.41 & 45.30 & \textbf{29.90} & 56.67 & \textbf{86.67} & \textbf{76.67} & \textbf{60.00} & \textbf{70.00} & 70.00 & \textbf{10.00} & 46.67 \\
    \bottomrule
\end{tabular}
      }
   \end{table*}

\section{Conclusion}

\dataset presents a comprehensive benchmark for evaluating temporal reasoning in LLMs, featuring three hierarchical levels with 11 subtasks that systematically assess temporal understanding. Our benchmark captures real-world complexities through knowledge associations, temporal dynamics, and long-term interactions. We introduce \textsc{TimE-Lite}, a fully human-annotated subset for efficient evaluation. Extensive experiments across diverse models reveal that test-time scaling significantly enhances logical reasoning while showing varied effects on time retrieval. These findings provide critical insights for advancing temporal reasoning capabilities. \dataset establishes a foundation for rigorous evaluation and deeper understanding of temporal reasoning, paving the way for future advancements in this essential NLP capability.

% acknowledgement
\section*{Acknowledgments}
This work was supported by Beijing Natural Science Foundation (No. L253020)  and National Natural Science
Foundation of China (62036001). The corresponding author is
Houfeng Wang.

% \section*{References}
\bibliographystyle{plain}
\bibliography{ref}{}

% References follow the acknowledgments in the camera-ready paper. Use unnumbered first-level heading for
% the references. Any choice of citation style is acceptable as long as you are
% consistent. It is permissible to reduce the font size to \verb+small+ (9 point)
% when listing the references.
% Note that the Reference section does not count towards the page limit.
% \medskip

% {
% \small

% [1] Alexander, J.A.\ \& Mozer, M.C.\ (1995) Template-based algorithms for
% connectionist rule extraction. In G.\ Tesauro, D.S.\ Touretzky and T.K.\ Leen
% (eds.), {\it Advances in Neural Information Processing Systems 7},
% pp.\ 609--616. Cambridge, MA: MIT Press.

% [2] Bower, J.M.\ \& Beeman, D.\ (1995) {\it The Book of GENESIS: Exploring
%   Realistic Neural Models with the GEneral NEural SImulation System.}  New York:
% TELOS/Springer--Verlag.

% [3] Hasselmo, M.E., Schnell, E.\ \& Barkai, E.\ (1995) Dynamics of learning and
% recall at excitatory recurrent synapses and cholinergic modulation in rat
% hippocampal region CA3. {\it Journal of Neuroscience} {\bf 15}(7):5249-5262.
% }

% checklist
% \newpage
% \input{main_content/7_G_Checklist}

%%%%%%%%%%%%%%%%%%%%%%%%%%%%%%%%%%%%%%%%%%%%%%%%%%%%%%%%%%%%
\newpage
\appendix

% % Appendix A
\section{Benchmark Construction\label{apdx: Benchmark Construction}}

\subsection{\textsc{\textsc{TimE-Wiki}} Construction}

\subsubsection{Data Source}
% 我们使用Wikidata作为TIME-Wiki的数据构建来源。我们首先从WikiMedia网站\footnote{WikiMedia: \url{https://dumps.wikimedia.org}}中下载了2024年11月1日版本的WikiData Dump\footnote{\url{https://dumps.wikimedia.org/enwiki/20241101/enwiki-20241101-pages-articles-multistream.xml.bz2}}\footnote{\url{https://dumps.wikimedia.org/enwiki/20241101/enwiki-20241101-pages-articles-multistream-index.txt.bz2}}，然后使用SLING\footnote{SLING: \url{https://github.com/ringgaard/sling}}，基于规则解析出给定relation下的时间事实，最后生成结构化的时间知识图谱数据。最后，我们对时间知识图谱数据进行增强，得到1300条多跳时间知识图谱作为后续构造QA的数据源。
We utilize Wikidata as the data source for constructing \textsc{TimE-Wiki}. We downloaded the WikiData Dump\footnote{\url{https://dumps.wikimedia.org/enwiki/20241101/enwiki-20241101-pages-articles-multistream.xml.bz2}}\footnote{\url{https://dumps.wikimedia.org/enwiki/20241101/enwiki-20241101-pages-articles-multistream-index.txt.bz2}} from November 1, 2024 from the WikiMedia\footnote{WikiMedia: \url{https://dumps.wikimedia.org}}.

\paragraph{WikiData} WikiData is a free and open collaborative knowledge base that serves as a central storage for the structured data of Wikimedia projects. It organizes information into items, each identified by a unique Q-number, and describes them using statements composed of properties and values. This structured data, akin to a large-scale knowledge graph, provides a valuable resource for Natural Language Processing tasks such as entity linking, relation extraction, and building comprehensive linguistic resources by connecting text to real-world entities and their relationships. 

Wikidata contains abundant real-world temporal facts that can be extracted to form multi-hop structured temporal knowledge graphs, which can then be transformed into unstructured temporal contexts for evaluating models' temporal fact comprehension capabilities.

\subsubsection{Temporal Knowledge Graph Construction\label{sec:TIME-Wiki TKG construction}}

Using SLING\footnote{SLING: \url{https://github.com/ringgaard/sling}}, we parsed temporal facts under given relations based on predefined rules, ultimately generating structured temporal knowledge graph data. Subsequently, we enhanced the temporal knowledge graph data to obtain 1,300 multi-hop temporal knowledge graphs as the data source for QA construction.

To comprehensively reflect world knowledge, we selected six categories of relations when constructing temporal facts from Wikidata: (1) Education, employment, and organizational affiliation, (2) Family relations, (3) Geographical location relations, (4) Naming relations, (5) Significant event, and (6) Role/Identity relations, as shown in Table \ref{tab:wikidata_relations}.
For each relation, we meticulously designed corresponding natural language templates to facilitate the subsequent synthesis of contexts containing complex temporal facts using LLMs.

\paragraph{SLING} SLING is a task processing pipeline designed for downloading and processing Wikipedia and Wikidata dumps. It leverages freely available dump files from Wikimedia and converts them into the SLING frame format through its workflow task system. Specifically, following the construction process of TempLAMA\cite{DBLP:journals/tacl/DhingraCEGEC22templama}, we first installed the SLING toolkit (version 3.0.0) and downloaded two key datasets: the SLING-formatted WikiData KB and the WikiData to Wikipedia mapping. We then used SLING to extract facts and relations with temporal properties \textbf{P580 (start time)}\footnote{P580 \textit{start time}: \url{https://www.wikidata.org/wiki/Property:P580}}, \textbf{P582 (end time)}\footnote{P582 \textit{end time}: \url{https://www.wikidata.org/wiki/Property:P582}}, and \textbf{P585 (point in time)}\footnote{P585 \textit{point in time}: \url{https://www.wikidata.org/wiki/Property:P585}}, filtering out non-entity objects and null objects. We mapped entity names to Wikipedia page titles, generating structured temporal knowledge graph data. The resulting temporal knowledge graph data contains fact relation sets from both subject and object perspectives.

\paragraph{Data Augmentation of Temporal Knowledge Graph} We first filtered entities associated with at least three facts from the SLING-generated temporal knowledge graph. We then removed facts with duplicate temporal information and alternately combined (s, r, o) relations from both subject and object perspectives. Specifically, we constructed a 2-hop temporal knowledge graph where we randomly selected 2-3 object entities as \textit{intermediate linking entities} at each hop. Each \textit{intermediate linking entity} then served as a new subject entity, concatenating new temporal fact relations. Finally, we selected 1,300 high-quality multi-hop temporal knowledge graphs as the data source for subsequent QA construction.

\begin{figure}
   \centering
   \includegraphics[width=\textwidth]{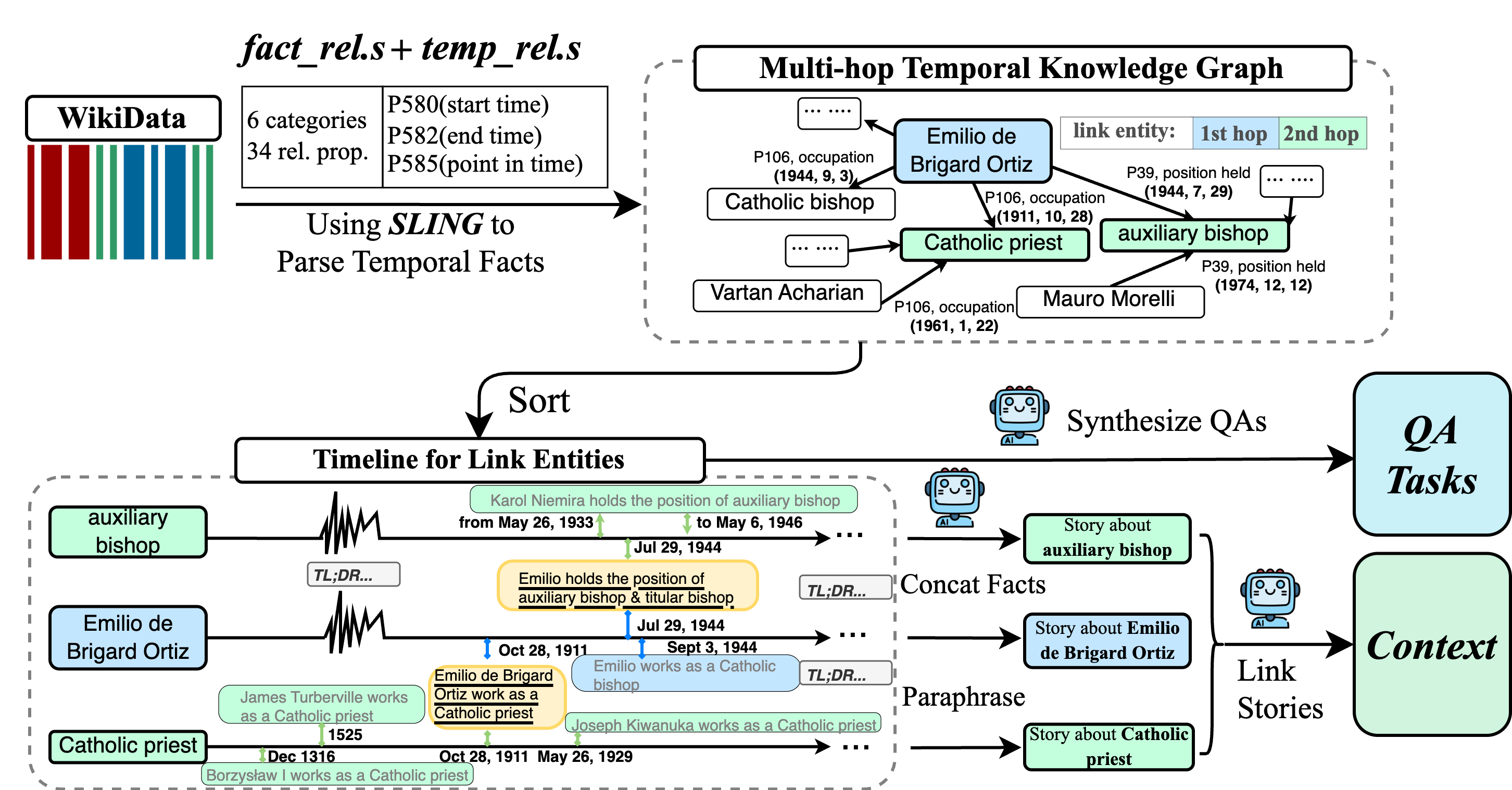}
   \caption{An overview of the \textsc{TimE-Wiki} benchmark construction pipeline. Beginning with Wikidata as the data source, temporal facts are parsed using SLING. These facts are then used to construct multi-hop temporal knowledge graphs. Timelines for link entities are generated from these graphs by sorting temporal facts. Finally, these timelines are used to synthesize question-answer (QA) pairs, and corresponding context is generated by concatenating and paraphrasing stories derived from the timelines, forming the final QA tasks.}
   \label{fig: example_wiki_data_construction}
\end{figure}

\begin{table}[htbp]
 \centering
 % \captionsetup{width=\linewidth} % Set caption width to line width
 % \caption{Selected Wikidata Properties Categorized by Semantic Type, including Property IDs, Query Frequencies, and Natural Language Templates} % Updated Caption
 \caption{Selected Wikidata Properties Categorized by Semantic Type, including Property IDs, Query Frequencies, and Natural Language Templates. Categories are highlighted by different colors: \colorbox{cyan!5}{Education, employment, and organizational affiliation}, 
 \colorbox{magenta!5}{Family relations}, \colorbox{yellow!8}{Geographical location relations}, \colorbox{green!5}{Naming relations}, \colorbox{orange!8}{Significant event}, and \colorbox{teal!5}{Role/Identity relations}. \textit{Num.} represents the number of facts with the corresponding relation that are extracted using SLING.} % Updated Caption with color explanations
 \label{tab:wikidata_relations}
 \begin{tabular}{@{}lp{4cm}rp{6cm}@{}}
    \toprule
    WikiData ID & Relation & Num. & Template \\
    \midrule
    % P39-P1416: Education, employment, organizational affiliation (Using cyan!5 - lighter)
    \rowcolor{cyan!5} P39 & position held & 165479 & <subject> holds the position of <object> \\
    \rowcolor{cyan!5} P54 & member of sports team & 903683 & <subject> plays for <object> \\
    \rowcolor{cyan!5} P69 & educated at & 68121 & <subject> attended <object> \\
    \rowcolor{cyan!5} P102 & political party & 16871 & <subject> is a member of <object> \\
    \rowcolor{cyan!5} P108 & employer & 63298 & <subject> works for <object> \\
    \rowcolor{cyan!5} P127 & owned by & 8065 & <object> owns <subject> \\
    \rowcolor{cyan!5} P286 & head coach & 10389 & <object> becomes the head coach of <subject> \\
    \rowcolor{cyan!5} P106 & occupation & 33209 & <subject> works as a <object> \\
    \rowcolor{cyan!5} P463 & member of & 41425 & <subject> becomes a member of <object> \\
    \rowcolor{cyan!5} P1416 & affiliation & 938 & <subject> becomes affiliated with <object> \\
    \midrule % Optional: Add midrule to visually separate categories
    % P22-P101: Family relations (Using magenta!5 - lighter)
    \rowcolor{magenta!5} P22 & father & 6 & <object> becomes the father of <subject> \\
    \rowcolor{magenta!5} P40 & child & 10 & <subject> has a child <object> \\
    \rowcolor{magenta!5} P101 & spouse & 175 & <subject> is the spouse of <object> \\
    \midrule % Optional: Add midrule to visually separate categories
    % P17-P159: Geographical location relations (Using yellow!8 - lighter)
    \rowcolor{yellow!8} P17 & country & 37344 & <subject> becomes a part of the country: <object> \\
    \rowcolor{yellow!8} P131 & located in the administrative territorial entity & 82221 & <subject> becomes a part of <object> \\
    \rowcolor{yellow!8} P276 & location & 2065 & <subject> becomes a part of <object> \\
    \rowcolor{yellow!8} P551 & residence & 7771 & <subject> resides in <object> \\
    \rowcolor{yellow!8} P740 & location of formation & 5 & <subject> was formed in <object> \\
    \rowcolor{yellow!8} P159 & headquarters location & 5619 & <subject> has its headquarters in <object> \\
    \midrule % Optional: Add midrule to visually separate categories
    % P138-P156: Naming relations (Using green!5 - lighter)
    \rowcolor{green!5} P138 & named after & 3558 & <object> names after <subject> \\
    \rowcolor{green!5} P155 & follows & 272 & <subject> follows <object> \\
    \rowcolor{green!5} P156 & followed by & 290 & <object> follows <subject> \\
    \midrule % Optional: Add midrule to visually separate categories
    % P793: Significant event (Using orange!8 - lighter)
    \rowcolor{orange!8} P793 & significant event & 4436 & <subject> experienced the significant event <object> \\
    \midrule % Optional: Add midrule to visually separate categories
    % P50-P1431: Role/Identity relations (Using teal!5 - lighter)
    \rowcolor{teal!5} P50 & author & 47 & <object> writes <subject> \\
    \rowcolor{teal!5} P57 & director & 62 & <object> directes <subject> \\ % Note: 'directes' seems like a typo, probably should be 'directs'
    \rowcolor{teal!5} P86 & composer & 92 & <object> composes <subject> \\
    \rowcolor{teal!5} P170 & creator & 21 & <object> creates <subject> \\
    \rowcolor{teal!5} P175 & performer & 132 & <object> performs <subject> \\
    \rowcolor{teal!5} P676 & lyrics by & 8 & <object> wrote the lyrics for <subject> \\
    \rowcolor{teal!5} P110 & illustrator & 11 & <object> illustrates <subject> \\
    \rowcolor{teal!5} P162 & producer & 31 & <object> produces <subject> \\
    \rowcolor{teal!5} P58 & screenwriter & 29 & <object> writes <subject> as a screenwriter \\
    \rowcolor{teal!5} P161 & cast member & 495 & <object> plays the role of <subject> \\
    \rowcolor{teal!5} P1431 & executive producer & 32 & <object> produces <subject> as the executive producer \\
    \bottomrule
 \end{tabular}
\end{table}
% P39-P1416是教育经历、工作单位和组织从属关系
% P22-P101是家庭关系
% P17-P159是地理位置关系
% P138-P156是命名关系
% P793是表达典型经历的关系
% P50-P1431是身份关系

% ---------------------补充材料：将TKG一步步转化到用于评估的context----------------
\subsubsection{Collect Context for Evaluation for \textsc{TimE-Wiki}}

% 为了评估时提供包含复杂的世界知识context，我们利用上一步构建的Temporal Knowledge Graph合成用于评估时提供给LLM的context。
% 具体而言，我们首先利用我们构建的temporal knowledge graph，构建link entity的timeline，作为结构化原始数据。然后，对于每一个link entity，我们按照时间顺序拼接各自的时间事实，然后paraphrase成对于三个link entity的story。最后，我们prompt LLM，生成以三个link entity为线索的的context。
% 下面是我们的few-shot prompt for paraphrasing facts into stories, and prompt for concatenating stories for \textsc{TimE-Wiki}.
% 最后，我们展示了context(即a concatenated story)。

To provide comprehensive world knowledge contexts for evaluation, we synthesize evaluation contexts for LLMs by leveraging the previously constructed Temporal Knowledge Graph. Specifically, we first build timelines for link entities from the temporal knowledge graph as structured raw data. For each link entity, we chronologically concatenate temporal facts and paraphrase them into coherent stories. Finally, we prompt LLMs to generate contexts centered around three interconnected link entities. Below we present our few-shot prompts for paraphrasing facts into stories and concatenating stories for \textsc{TimE-Wiki}, followed by an example of the final concatenated context.

% 下面1.1 插入转换stories to 
\begin{tcolorbox}[title=Few-shot Prompt for Paraphrasing Facts into Stories for \textsc{TimE-Wiki}, label={box: prompt_paraphrase_facts_into_stories}]
\textbf{[Instruction:]}  
You are a very skilled storyteller. Below are the facts I've extracted from WikiData. Please craft a complete story centered around \{link\_entity\}.

\textbf{[Requirements:]}

1. Generate content that accurately conveys all the facts I have provided without omitting any of them.

2. Particular attention must be paid to the accuracy of time expressions, and the temporal relationships between sentences must be coherent and natural.

3. The story should be coherent, natural, and smooth, consistently revolving around the core facts, with concise and precise language, avoiding any unnecessary digressions.

\textbf{[Example:]}
[Core entity:] Mauro Morelli

[Facts:] Mauro Morelli holds the position of diocesan bishop from May 25, 1981 to March 30, 2005.. Mauro Morelli holds the position of titular bishop in December 12, 1974.. Mauro Morelli holds the position of auxiliary bishop in December 12, 1974.. Mauro Morelli works as a Catholic priest in April 28, 1965.. Mauro Morelli works as a Catholic bishop in January 25, 1975.

[Generated Story:] Mauro Morelli's journey in the Catholic Church is a testament to his unwavering dedication and service. Born into a world of faith, his path was set early on, leading him to become a significant figure in the Church's hierarchy.

On April 28, 1965, Mauro Morelli took his first major step in his religious career by being ordained as a Catholic priest. This ordination marked the beginning of his lifelong commitment to serving the Church and its followers. For nearly a decade, he devoted himself to his priestly duties, gaining the respect and admiration of his parishioners.
... ...

\textbf{[Example:]}
(Here is another example with the same format of the example above.)

\textbf{[TODO:]}
Below are the facts and the link entity. Please generate a story in accordance with the aforementioned requirements and with reference to the examples provided.

\textbf{[Link entity:]}\{link\_entity\}

\textbf{[Facts:]}\{raw\_text\}

\textbf{[Generated story:]}
\end{tcolorbox}

% 下面1.2 插入prompt for concat stories
\begin{tcolorbox}[title=Prompt for Concatenating Stories for \textsc{TimE-Wiki}, label={box: prompt_concat_stories}]
\textbf{[Instruction:]}  
You are adept at weaving multiple complete stories together to form a coherent and comprehensive narrative. I will provide you with several stories and the connections between them. I hope you can retain all the facts from the stories without any omissions or errors in time or events.

\textbf{[Requirements:]}  

1. The content you generate must not omit any time-related information from the texts being concatenated.

2. Particular attention must be paid to the accuracy of time expressions, and the temporal relationships between sentences must be coherent and natural.

3. When concatenating, pay attention to the inherent logical connections between the given multiple stories, and refer to the provided linking facts to generate transitional paragraphs that seamlessly connect the different stories.

\textbf{[Link facts:]}
\{linked\_facts\}

\textbf{[Stories:]}
\{stories\}

\textbf{[Generated Story:]}

\end{tcolorbox}

% 2.插入context示例

\begin{tcolorbox}[title=One Example for the Concatenated Story for \textsc{TimE-Wiki}, label={box: example_concated_story}]
Mauro Morelli’s life was a testament to unwavering dedication and service to the Catholic Church. His journey began with a profound commitment to his faith, which shaped his path toward becoming a respected leader within the Church’s hierarchy. On April 28, 1965, Mauro Morelli was ordained as a Catholic priest, marking the beginning of his spiritual and pastoral vocation. For nearly a decade, he devoted himself to his priestly duties, guiding and nurturing the communities he served. His compassion, wisdom, and deep connection to his faith earned him the trust of his parishioners and the attention of his superiors.\\ \\Recognizing his leadership qualities, the Church appointed Mauro Morelli as both a titular bishop and an auxiliary bishop on December 12, 1974. These dual roles signified a new chapter in his journey, as he took on greater responsibilities in assisting the diocesan bishop and overseeing pastoral and administrative tasks. His dedication to these roles showcased his ability to balance spiritual guidance with effective governance. Just over a month later, on January 25, 1975, Mauro Morelli was consecrated as a Catholic bishop. This formal consecration solidified his position as a key figure within the Church, empowering him to lead with authority and grace. His work as a bishop further deepened his impact on the communities he served, as he continued to champion the values of the Church.\\ \\
\textbf{\textit{Skipping 3 paragraphs here...}} \\ \\
As we reflect on Mauro Morelli’s life and the broader history of auxiliary and titular bishops, we are reminded of the importance of service, humility, and faith. The roles of these bishops, though sometimes overlooked, are a testament to the Church's enduring commitment to its mission of love, guidance, and spiritual care. Their contributions, woven together, form a testament to the enduring legacy of leadership within the Catholic tradition.
\end{tcolorbox}

% ---------------------补充材料：将TKG一步步转化到用于评估的context----------------

\subsection{\textsc{TimE-News} Construction}
\subsubsection{Data Source}
% TIME-News旨在探究现实世界中的复杂temporal dynamics下的时间推理。现有工作通常使用news articles来分析复杂的时序动态事件，如\citeN{ma2023sctc,DBLP:conf/acl/Zhang00MLC24TCELongbench}。
% 我们选择基准\cite{DBLP:conf/acl/Zhang00MLC24TCELongbench}的开源数据来作为TIME-News的数据源，考虑到它的数据库数量足够庞大，并且容易获取，并且具有较为丰富的时间信息。该数据源使用LLMs来系统性地提取和分析event chain within temporal complex events(TCEs)，其中each temporal complex event is composed of many news articles over an extended period.
% 具体而言，\cite{DBLP:conf/acl/Zhang00MLC24TCELongbench}一共包含了2289个temporal complex events(TCEs)，并且每一个复杂事件平均有29.31个articles和17.44 days的跨度。在开源数据中，我们可以获取到每一个TCE的corpus和对应的outline，如图xx?所示。我们从中随机挑选了600个TCEs。对于每一个TCE，我们使用outline来合成QA；在评估阶段，我们使用corpus作为检索数据源，对于每一个question（即query），特定检索top-k篇news articles。
\textsc{TimE-News} investigates temporal reasoning in real-world complex dynamics. Prior work typically employs news articles to analyze intricate temporal event sequences \cite{ma2023sctc,DBLP:conf/acl/Zhang00MLC24TCELongbench}. We utilize the open-source dataset from \cite{DBLP:conf/acl/Zhang00MLC24TCELongbench} as our data source, selected for its extensive scale, accessibility, and rich temporal information. This dataset leverages LLMs to systematically extract and analyze event chains within temporal complex events (TCEs), where each TCE comprises multiple news articles spanning extended periods.

\begin{figure}
   \centering
   \includegraphics[width=\textwidth]{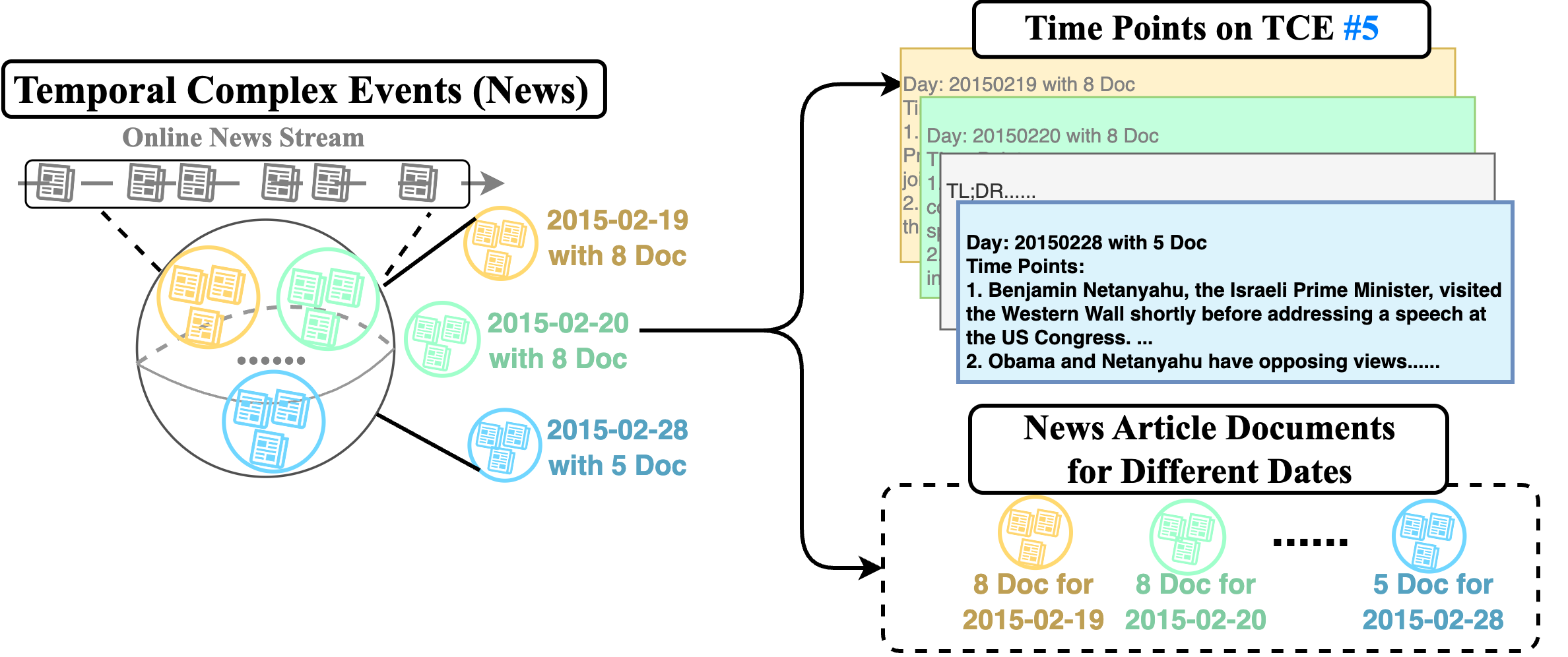}
   \caption{One example of temporal complex events in dataset\cite{DBLP:conf/acl/Zhang00MLC24TCELongbench}}
   \label{fig: example_TCE}
\end{figure}

\begin{figure}
   \centering
   \includegraphics[width=0.65\textwidth]{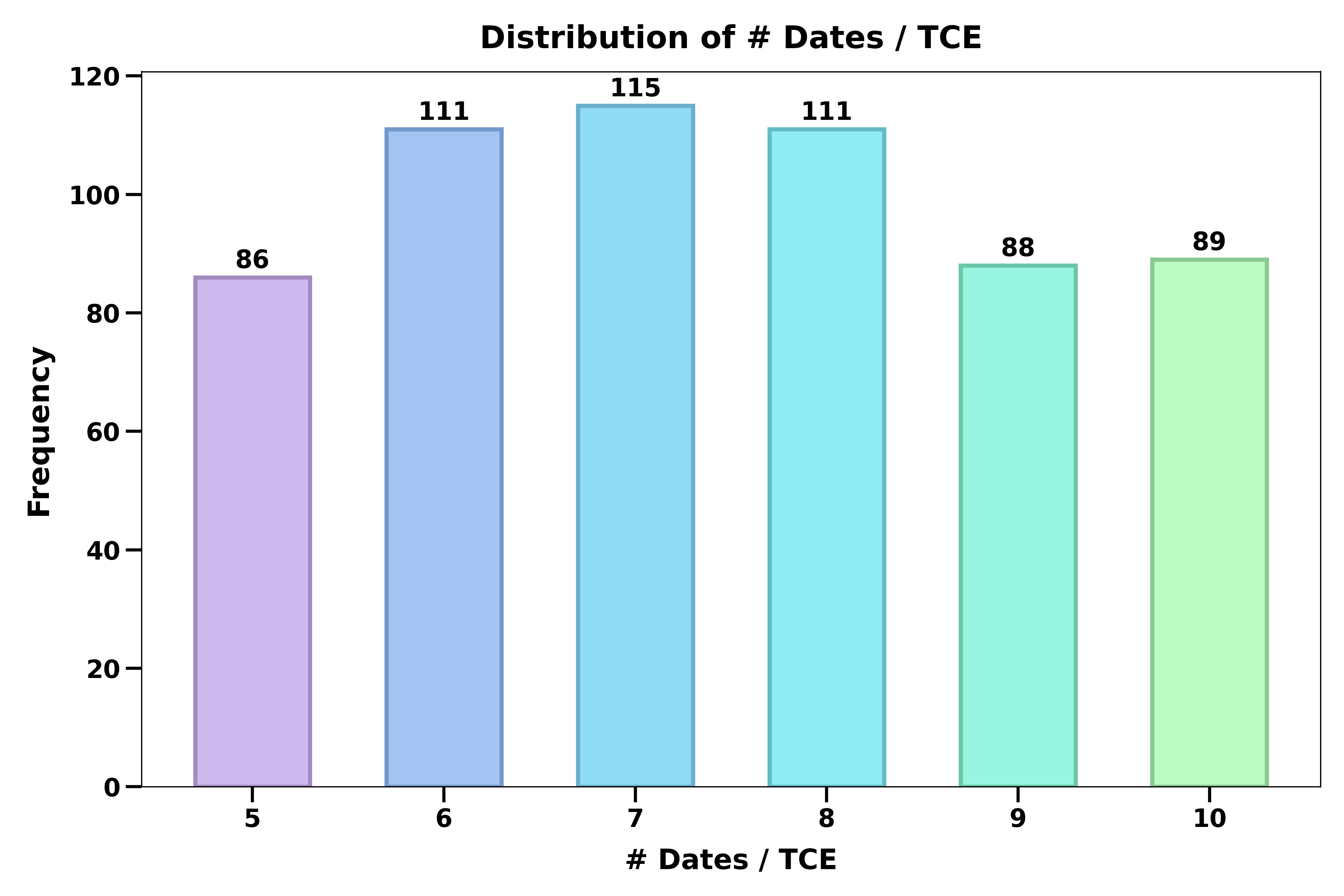}
   \caption{Frequency distribution of length of days in 600 selected Temporal Complex Events (TCEs) in \textsc{TimE-News}}
   \label{fig: tcelongbench_days_num_distribution}
\end{figure}

Specifically, dataset \cite{DBLP:conf/acl/Zhang00MLC24TCELongbench} contains 2,289 TCEs, with each event averaging 29.31 articles and spanning 17.44 days. The open-source data provides both the corpus and corresponding outline for each TCE, as shown in Figure \ref{fig: example_TCE}. 
We randomly sampled 600 temporal complex events (TCEs) with at least 5 distinct dates and at most 10 distinct dates as the data source for \textsc{TimE-News}, with detailed statistics presented in Table \ref{tab:tce_statistics} and the number of dates distribution presented in Figure \ref{fig: tcelongbench_days_num_distribution}. For each TCE, we utilize its outline for QA synthesis. During evaluation, we employ the corpus as the retrieval source, where for each question (i.e., query), we specifically retrieve the top-k relevant news articles.

% TODO:画图:来表达Temporal Complex Events是什么、提取的时间线是怎样的形式，用例子形式比较好（可以参考pipeline的图）。

% TODO:表格: 来说明我们选择的statistics，比如说平均每一个Temporal Complex Events对应的article数量、长度等数据。

% 统计结果：
% 每个复杂事件平均文章数量: 33.67
% 每个复杂事件平均token长度: 21018.01
% 最大文章数量: 166
% 最大token长度: 146300
% 最小文章数量: 5
% 最小token长度: 571
% 每个复杂事件平均天数跨度: 405.87
% 最大天数跨度: 8900
% 最小天数跨度: 4
% 每个复杂事件平均不同日期数量: 9.47
% 最大不同日期数量: 25
% 最小不同日期数量: 5

% TODO:修改这里的数据，因为每一个TCE的article数量和token长度是有问题的。
\begin{table}[htbp]
    \centering
    \caption{Statistics of the 600 selected Temporal Complex Events (TCEs) in \textsc{TimE-News}. Note that "Time Span" means the span between the earliest and the latest date of the TCE. "Art." and "Tok." are abbreviations for "Article" and "Token" respectively. Token counts are calculated using \textit{tiktoken's cl100k\_base} encoder.}
    \label{tab:tce_statistics}
        \resizebox{\textwidth}{!}{% Resize table to fit within text width
    \begin{tabular}{lcccccc}
        \toprule
        \textbf{} & \textbf{\# Art. / TCE} & \textbf{\# Art. / Date} & \textbf{\# Tok. / TCE} & \textbf{\# Tok. / Art.} & \textbf{Time Span / TCE} & \textbf{\# Dates / TCE} \\
        \midrule
        Avg. & 871.46 & 116.95 & 527,418.05 & 605.21 & 405.87 & 7.45 \\
        Max. & 2068 & 304 & 1,232,706 & 3,405 & 8,900 & 10 \\
        Min. & 344 & 18 & 236,521 & 31 & 4 & 5 \\
        \bottomrule
    \end{tabular}
    }
\end{table}

\subsection{\textsc{TimE-Dial} Construction}
\begin{figure}
   \centering
   \includegraphics[width=\textwidth]{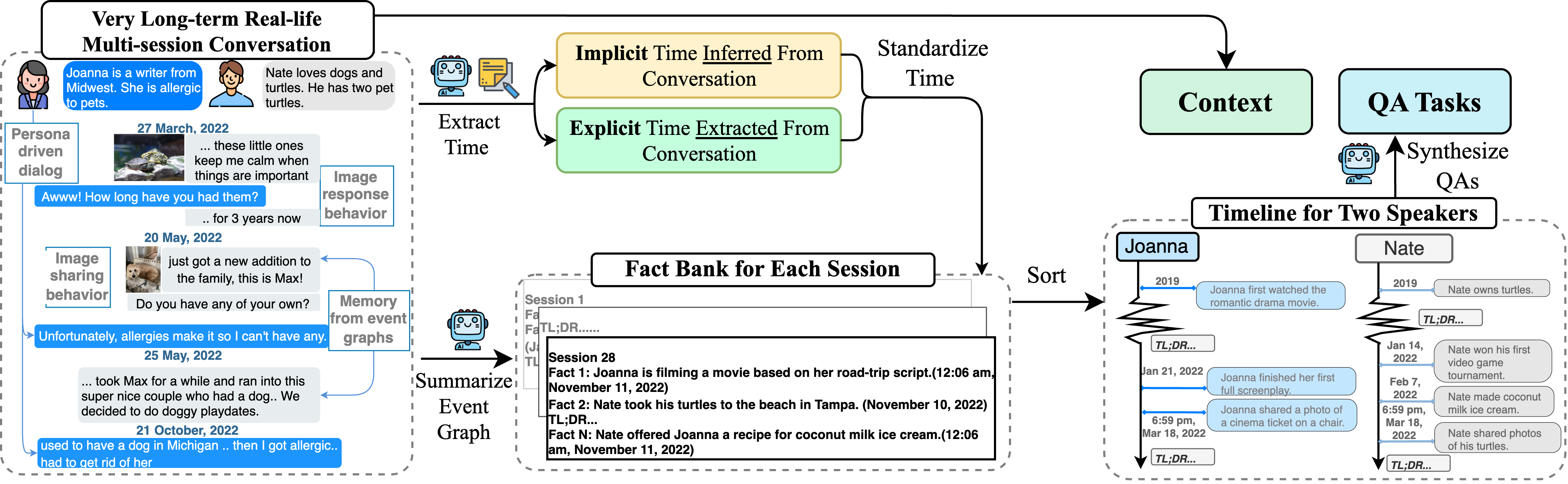}
   \caption{
      An overview of the construction pipeline of \textsc{TimE-Dial}. We employ real-world long-term multi-turn conversations as evaluation contexts. First, we extract and summarize event graphs from the conversations. Then, through a combination of LLM extraction and manual verification, we identify explicit temporal information and infer implicit temporal relations, standardizing them into unified temporal expressions. Subsequently, we temporally order the event graphs to construct individual timelines for each speaker. Finally, we generate question-answer pairs covering 11 subtasks based on these timeline representations.
   }
   \label{fig: TIME-Dial data construction pipeline}
\end{figure}

% \textsc{TIME-Dial}的合成数据流程中，我们选择数据源very long-term real-life multi-session conversations\cite{DBLP:conf/acl/MaharanaLTBBF24locomo,DBLP:journals/corr/abs-2502-13270realtalk}，直接作为评估时使用的context。我们首先从conversation中summarize event graph。然后，我们使用LLM提取和人工验证的方式，提取explicit time并推断implicit time，并standardize time得到统一的时间表达形式。然后，我们将event graph进行排序，得到两个speaker各自的时间线。最后，用时间线信息来合成11个子任务的quesiton-answer pairs。\textsc{TIME-Dial}的合成数据流程如图\ref{fig: TIME-Dial data construction pipeline}所示。
% \textsc{TIME-Dial}的合成数据流程如下：我们以真实场景中的长期多轮对话作为评估上下文。首先，从对话内容中提取并总结事件图谱。随后，结合大语言模型（LLM）和人工验证，我们提取显式时间信息并推断隐式时间关系，将其统一为标准化的时间表达。接着，对事件图谱进行时序排序，构建出对话双方各自的时间线。最终，基于这些时间线信息，我们生成了涵盖11个子任务的问答对。
% ---------------------------------------正文-------------------------------------------
In the data construction pipeline of \textsc{TimE-Dial}, we utilize very long-term real-life multi-session conversations \cite{DBLP:conf/acl/MaharanaLTBBF24locomo,DBLP:journals/corr/abs-2502-13270realtalk} as the evaluation context. Our pipeline begins with summarizing event graphs from conversations. We then employ a combination of LLM extraction and manual verification to extract explicit temporal information and infer implicit temporal relations, which are subsequently standardized into unified temporal expressions. The event graphs are then temporally ordered to construct individual timelines for each speaker. Finally, we synthesize question-answer pairs for 11 subtasks based on these timeline representations. The complete construction pipeline is illustrated in Figure \ref{fig: TIME-Dial data construction pipeline}.

\subsubsection{Data Source}
% 这里需要添加很多对数据源本身的分析
% TODO:
% 1. 强调为什么这个很重要？：贴近现实世界的真实对话场景、长文本挑战，另外还可以参考TReMu的故事。（已完成）
% 2. 强调长文本，并且画表————统计总数据源+LoCoMo+RealTalk的统计数据。（包括session数量、token数量、turn数量、image数量等）TODO:
% 3. 对比LoCoMo和RealTalk和之前的各数据集的统计数据（这里可以直接照搬RealTalk里面的statistics表格）TODO:
% 4. 强调案例，并且画图————画一个体现这个数据源突出特征的图（可以考虑直接从pipeline的图里面搬）（已完成）

% 在现实世界中，复杂的人际交互关系是时间推理的重要场景。从数据特征上讲，人与人之间的交流往往具有长期性、持续性特点：对话人之间存在共同记忆，在对话中往往依赖于之前已经交谈过或者共同已知的信息，并能够延续话题。此外，现实世界中的会话次数是非常多的，不同会话之间存在着各种时间跨度，并且含有大量的琐碎内容，如语气词、每turn的发言比较简短，主观信息丰富。更重要的是，超长对话上下文中的时间信息处理，仍然未被充分地探究。
% 我们选择了LoCoMo和RealTalk作为数据源，如图\ref{fig: dial data source demo}所示。这是因为它们不仅是超长的、由多次会话组成的conversation数据集，而且他们通过数据合成方式，生成了persona driven dialog，非常符合真实的人类特征。此外，在这两个数据集中，对话人之间会交换除了自然语言，例如图片等，并且会对图片信息进行回复，符合现实场景中的对话需求。然而，由于我们只考虑评估LLM对于自然语言环境下的时间推理，所以我们将图片的caption信息而不是直接使用原始数据集中的image作为图片信息。
% --------------------正文-------------------
Real-world interpersonal interactions provide a crucial context for temporal reasoning research. Human conversations exhibit distinct characteristics of long-term continuity and persistence: interlocutors frequently reference previously discussed or mutually known information based on shared memory, enabling natural topic continuation. Moreover, authentic dialogues typically involve numerous sessions with significant temporal spans, containing rich fine-grained information such as discourse markers, brief turns, and subjective expressions. Notably, the processing of temporal information in ultra-long conversational contexts remains an under-explored research area.

\begin{figure}
   \centering
   \includegraphics[width=0.8\textwidth]{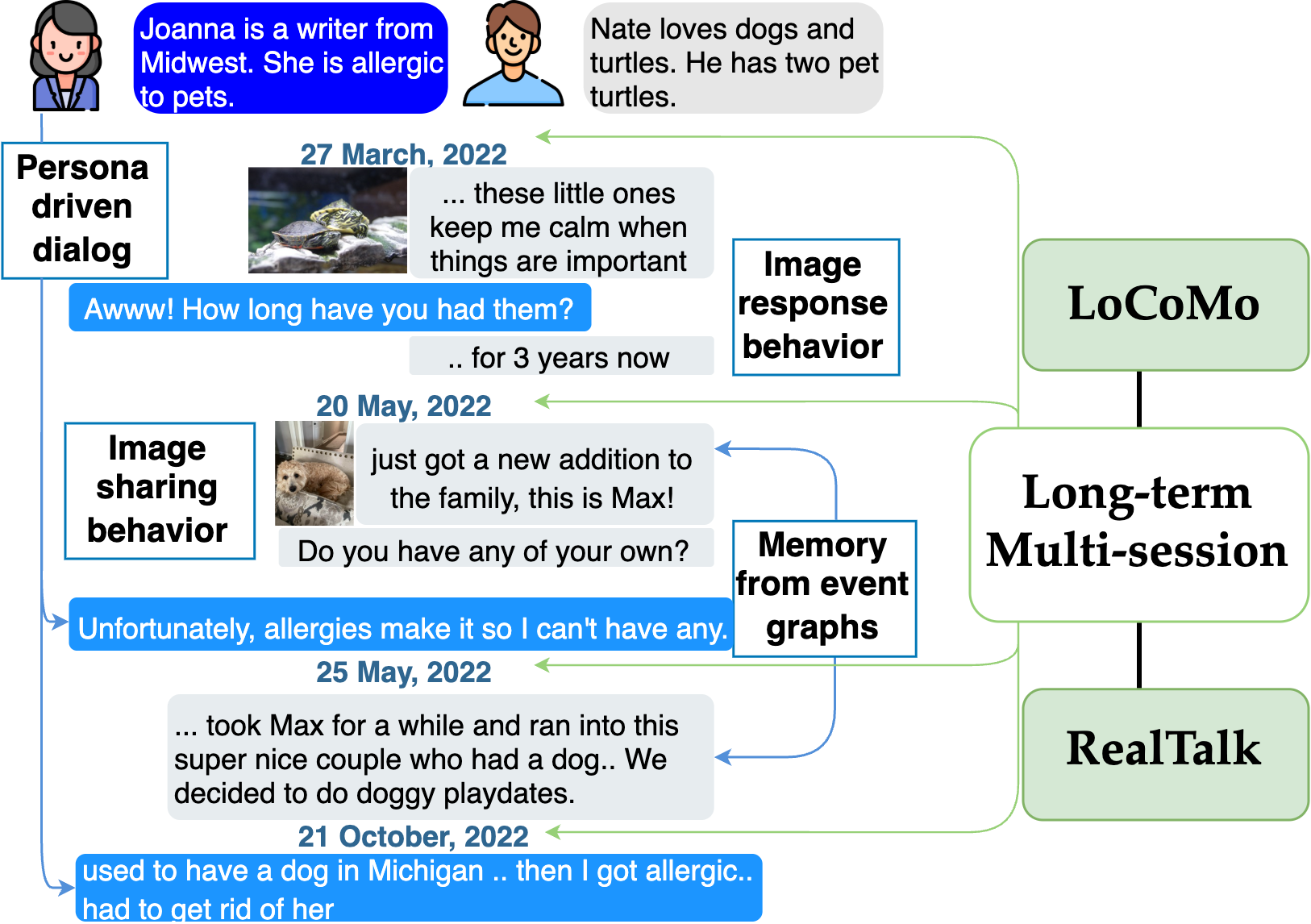}
   \caption{
      A long-term multi-turn conversation example from the \textsc{LoCoMo} dataset \cite{DBLP:conf/acl/MaharanaLTBBF24locomo}. The interlocutors, Joanna and Nate, are each assigned distinct persona characteristics. The conversation consists of multiple session units, each timestamped and containing several dialogue turns. Throughout the interaction, the participants engage in image sharing and response behaviors. Notably, certain dialogue segments require the speakers to leverage previously established shared memory to facilitate conversational progression. \textsc{RealTalk} \cite{DBLP:journals/corr/abs-2502-13270realtalk} maintains an identical data format to the example illustrated in this figure.
   }
   \label{fig: dial data source demo}
\end{figure}

We select LoCoMo\cite{DBLP:conf/acl/MaharanaLTBBF24locomo} and RealTalk\cite{DBLP:journals/corr/abs-2502-13270realtalk} as our data sources (illustrated in Figure \ref{fig: dial data source demo}) based on the following considerations: First, both datasets contain ultra-long conversations composed of multiple sessions, with persona-driven dialogues generated through data synthesis techniques that align with human characteristics. Second, the datasets feature not only natural language exchanges but also multimodal content such as images, with corresponding responses, closely mirroring real-world conversational scenarios. Given our focus on temporal reasoning evaluation in natural language contexts, we utilize image captions rather than raw image data for analysis.

\paragraph{\textsc{LoCoMo}} \textsc{LoCoMo} \cite{DBLP:conf/acl/MaharanaLTBBF24locomo} addresses the scarcity of datasets for evaluating long-term memory in open-domain dialogues, which traditionally span few sessions. It was created using a machine-human pipeline where LLM-based agents generate dialogues grounded in personas and, crucially for our work, temporal event graphs. These conversations are extensive, averaging 300 turns and 9K tokens over as many as 35 sessions. This characteristic makes \textsc{LoCoMo} a valuable source for \textsc{TimE-Dial}, providing the very long-term, multi-session conversational data needed to construct complex temporal reasoning tasks. While \textsc{LoCoMo} agents can share and react to images, for \textsc{TimE-Dial}, we utilize the textual captions of these images to maintain focus on natural language-based temporal reasoning.
As of February 2025, \textsc{LoCoMo} has released only 35 conversations in \footnote{\textsc{LoCoMo-35} is downloaded from Google Drive: \url{https://drive.google.com/file/d/1JimNy04ery0Ijz6dZwLaqWs7glumE2v-/view}}, which we incorporate as part of our data source. 
% Existing works on long-term open-domain dialogues focus on evaluating model responses within contexts spanning no more than five chat sessions.

% We introduce a machine-human pipeline to generate high-quality, very long-term dialogues by leveraging LLM-based agent architectures and grounding their dialogues on personas and temporal event graphs. Moreover, we equip each agent with the capability of sharing and reacting to images. The generated conversations are verified and edited by human annotators for long-range consistency and grounding to the event graphs. Using this pipeline, we collect LoCoMo, a dataset of very long-term conversations, each encompassing 300 turns and 9K tokens on avg., over up to 35 sessions.

% Based on LOCOMO, we present a comprehensive evaluation benchmark to measure long-term memory in models, encompassing question answering, event summarization, and multi-modal dialogue generation tasks.

\paragraph{\textsc{RealTalk}} \textsc{RealTalk} \cite{DBLP:journals/corr/abs-2502-13270realtalk} provides a corpus of authentic, multi-session dialogues collected from real-world messaging app interactions over a 21-day period. This dataset captures genuine human conversational dynamics and long-term interaction patterns, offering a benchmark against true human interactions rather than synthetic data. For the construction of \textsc{TimE-Dial}, \textsc{RealTalk} serves as an invaluable source of very long-term, real-life conversational contexts. These characteristics are instrumental for developing tasks that require understanding complex temporal relationships and recalling information across extended dialogues, aligning with our objective to evaluate nuanced temporal reasoning capabilities within multi-session conversations.
% Persona simulation: where a model continues a conversation on be- half of a specific user given prior dialogue con- text和memory probing: where a model answers targeted questions requiring long-term memory of past interactions.

\begin{table}[h]
\centering
\caption{
   Statistics of \textsc{LoCoMo-35}, \textsc{RealTalk}, and \textsc{TimE-Dial} datasets. Note: Token counts are calculated using \textit{tiktoken's cl100k\_base} encoder. \textsc{LoCoMo-35} is the open-source subset of \textsc{LoCoMo} as of February 2025. $C$ represents "Conversation".
}
\begin{tabular}{lccccc}
\toprule
\textbf{Dataset} & \textbf{\# $C$} & \textbf{\# Session / $C$} & \textbf{\# Token / $C$} & \textbf{\# Turn / $C$} & \textbf{\# Image / $C$} \\
\midrule
\textsc{LoCoMo-35}\cite{DBLP:conf/acl/MaharanaLTBBF24locomo} & 35 & 20.49 & 14509.91 & 431.23 & 94.94 \\
\textsc{RealTalk}\cite{DBLP:journals/corr/abs-2502-13270realtalk} & 10 & 21.90 & 20581.60 & 894.40 & 31.30 \\
\textsc{TimE-Dial} & 45 & 20.80 & 15859.18 & 534.16 & 80.80 \\
\bottomrule
\end{tabular}
\label{tab:statistics of data source of TIME-Dial, LoCoMo-35 and RealTalk}
\end{table}

% \subsubsection{Fact Bank Generation}

% 这里是dialog data construction的例子

% \subsubsection{Timeline Generation}

\subsection{QA Synthesis\label{sec: appendix_qa_synthesis}}

% 我们结合时间线和规则，使用\texttt{DeepSeek-V3} and \texttt{DeepSeek-R1}来合成QA。然而，具体到每一个任务而言，合成数据所使用的方法（如表格\ref{tab: Overview of LLM Utilization Strategies in Question Answering (QA) Construction}）和任务形式（如表格\ref{tab: Overview of Question Answering (QA) formats across datasets and task categories.}）均有所不同。在\S\ref{sec: synthesize QA for Level 1}, \S\ref{sec: synthesize QA for Level 2}以及\S\ref{sec: synthesize QA for Level 3}中分别具体阐述QA synthesis的思路和细节。

We synthesize QA pairs by integrating timelines and rules, utilizing \texttt{DeepSeek-V3} and \texttt{DeepSeek-R1}. However, the methodologies employed for data synthesis (as shown in Table~\ref{tab: Overview of LLM Utilization Strategies in Question Answering (QA) Construction}) and task formats (as presented in Table~\ref{tab: Overview of Question Answering formats}) vary across different tasks. The approaches and details of QA synthesis are elaborated in \S\ref{sec: synthesize QA for Level 1}, \S\ref{sec: synthesize QA for Level 2}, and \S\ref{sec: synthesize QA for Level 3}, respectively.

\begin{table*}[htbp]
   \centering
   \caption{Overview of LLM Utilization Strategies in Question Answering (QA) Construction.
   The symbols in the cells denote the construction methodology:
   $\mathcal{R}$: Purely \textbf{r}ule-based and template-driven QA generation.
   $\mathcal{L}$: QA generation fully reliant on \textbf{L}arge Language Models (LLMs).
   $\mathcal{H_Q}$: \textbf{H}ybrid for \textbf{Q}uestion generation, i.e. rule-based extraction of question logic and answers, with LLMs employed for question phrasing.
   The suffix '{$+\mathcal{M}$}' indicates the additional use of LLMs specifically for generating \textbf{m}isleading options.
   \colorbox{cyan!15}{Light blue} shaded cells denote data generated by the \texttt{DeepSeek-R1}, while uncolored cells correspond to the \texttt{DeepSeek-V3}.
   }
   \label{tab: Overview of LLM Utilization Strategies in Question Answering (QA) Construction}
   \resizebox{\textwidth}{!}{% Resize table to fit within text width
   \begin{tabular}{@{}lccccccccccc@{}}
   \toprule
   \multirow{2}{*}{\textbf{Dataset}} & \multicolumn{5}{c}{\textbf{Level 1}} & \multicolumn{3}{c}{\textbf{Level-2}} & \multicolumn{3}{c}{\textbf{Level-3}} \\
   \cmidrule(lr){2-6} \cmidrule(lr){7-9} \cmidrule(lr){10-12}
   & \textbf{Extract} & \textbf{Local.} & \textbf{Comp.} & \textbf{Dur. Comp.} & \textbf{Ord. Comp.} & \textbf{Expl. Reason.} & \textbf{Ord. Reason.} & \textbf{Rel. Reason.} & \textbf{Co-temp.} & \textbf{Timeline} & \textbf{Counterf.} \\
   \midrule
   \textsc{TimE-Wiki} & $\mathcal{H_Q}+\mathcal{M}$ & $\mathcal{H_Q}$ & $\mathcal{H_Q}$ & $\mathcal{R}$ & $\mathcal{R}$ & $\mathcal{H_Q}$ & $\mathcal{H_Q}$ & $\mathcal{H_Q}$ & $\mathcal{H_Q}$ & $\mathcal{R}$ & $\mathcal{H_Q}$ \\
   \textsc{TimE-News} & $\mathcal{L}+\mathcal{M}$ & $\mathcal{L}$ & $\mathcal{L}$ & $\mathcal{L}$ & $\mathcal{L}$ & $\mathcal{L}+\mathcal{M}$ & $\mathcal{L}+\mathcal{M}$ & $\mathcal{L}+\mathcal{M}$ & $\mathcal{L}+\mathcal{M}$ & \textsuperscript{*} & $\mathcal{H_Q}+\mathcal{M}$ \\
   \textsc{TimE-Dial} & $\mathcal{H_Q}+\mathcal{M}$ & $\mathcal{H_Q}$ & $\mathcal{H_Q}$ & $\mathcal{R}$ & $\mathcal{R}$ & $\mathcal{H_Q}$ & \cellcolor{cyan!15}{$\mathcal{H_Q}+\mathcal{M}$} & \cellcolor{cyan!15}{$\mathcal{H_Q}+\mathcal{M}$} & \cellcolor{cyan!15}{$\mathcal{H_Q}+\mathcal{M}$} & $\mathcal{R}$ &  \cellcolor{cyan!15}{$\mathcal{H_Q}+\mathcal{M}$} \\
   \bottomrule
   \multicolumn{12}{@{}l}{\textsuperscript{*}\textit{Not applicable; data directly reused from TCELongBench\cite{DBLP:conf/acl/Zhang00MLC24TCELongbench} examples.}} \\
   \end{tabular}%
   }
   \end{table*}

   \begin{table*}[htbp]
      \centering
      \caption{Overview of Question Answering (QA) formats across datasets and task categories. The QA formats are denoted by calligraphic letters: $\mathcal{F}$ (\textbf{f}ree-form), $\mathcal{S}$ (\textbf{s}ingle-choice), and $\mathcal{M}$ (\textbf{m}ultiple-choice). A \colorbox{green!15}{light green} cell background indicates that the task's gold answer includes a refusal option, and models are permitted to decline answering during evaluation. Refusal options include answers such as \textit{"There is no answer.", "None of the above."}}
      \label{tab: Overview of Question Answering formats}
      \resizebox{\textwidth}{!}{% Resize table to fit within text width
      \begin{tabular}{@{}lccccccccccc@{}}
      \toprule
      \multirow{2}{*}{\textbf{Dataset}} & \multicolumn{5}{c}{\textbf{Level 1}} & \multicolumn{3}{c}{\textbf{Level-2}} & \multicolumn{3}{c}{\textbf{Level-3}} \\
      \cmidrule(lr){2-6} \cmidrule(lr){7-9} \cmidrule(lr){10-12}
      & \textbf{Extract} & \textbf{Local.} & \textbf{Comp.} & \textbf{Dur. Comp.} & \textbf{Ord. Comp.} & \textbf{Expl. Reason.} & \textbf{Ord. Reason.} & \textbf{Rel. Reason.} & \textbf{Co-temp.} & \textbf{Timeline} & \textbf{Counterf.} \\
      \midrule
      \textsc{TimE-Wiki} & \cellcolor{green!15}$\mathcal{M}$ & $\mathcal{F}$ & $\mathcal{F}$ & $\mathcal{S}$ & $\mathcal{S}$ & \cellcolor{green!15}$\mathcal{F}$ & $\mathcal{F}$ & \cellcolor{green!15}$\mathcal{F}$ & \cellcolor{green!15}$\mathcal{F}$ & $\mathcal{F}$ & \cellcolor{green!15}$\mathcal{F}$ \\
      \textsc{TimE-News} & $\mathcal{M}$ & $\mathcal{F}$ & $\mathcal{F}$ & $\mathcal{S}$ & $\mathcal{S}$ & $\mathcal{S}$ & $\mathcal{S}$ & $\mathcal{S}$ & $\mathcal{S}$ & $\mathcal{F}$ & \cellcolor{green!15}$\mathcal{S}$ \\
      \textsc{TimE-Dial} & $\mathcal{M}$ & $\mathcal{F}$ & $\mathcal{F}$ & $\mathcal{S}$ & $\mathcal{S}$ & $\mathcal{S}$ & $\mathcal{S}$ & $\mathcal{S}$ & $\mathcal{S}$ & $\mathcal{F}$ & \cellcolor{green!15}$\mathcal{S}$ \\
      \bottomrule
      \end{tabular}%
      }
   \end{table*}

\paragraph{DeepSeek-V3 and DeepSeek-R1}
\texttt{DeepSeek-V3}\cite{DBLP:journals/corr/abs-2412-19437deepseek-v3} and \texttt{DeepSeek-R1}\cite{DBLP:journals/corr/abs-2501-12948deepseek-r1} are both released by DeepSeek, representing state-of-the-art non-reasoning and reasoning models, respectively. These models offer superior performance at cost-effective rates, making them widely adopted for data synthesis and various downstream applications.
After evaluating both performance and cost considerations, we primarily employed \texttt{DeepSeek-V3} for synthesizing QA pairs across most tasks. However, for certain Level-2 and Level-3 tasks in the \textsc{TimE-Dial} dataset, which involve processing numerous input instances and require complex reasoning capabilities, we utilized the advanced Large Reasoning Model \texttt{DeepSeek-R1} to ensure high-quality QA synthesis. Specifically, \texttt{DeepSeek-R1} was deployed for generating the \texttt{Order Reasoning}, \texttt{Relative Reasoning}, \texttt{Co-temporality}, and \texttt{Counterfactual} tasks within the \textsc{TimE-Dial} dataset. Note that the DeepSeek-V3 used in data synthesis is the version released in December 2024.

\paragraph{QA Formats} 
The \textsc{TimE} dataset incorporates three distinct QA formats: \textbf{free-form}, \textbf{single-choice}, and \textbf{multiple-choice}. The free-form format requires short-answer generation, exemplified by questions like "\textit{When did Nicola Agnozzi become an auxiliary bishop?}" with concise answers such as "\textit{April 2, 1962.}" This format is particularly suitable for questions with limited synonymous answer variations, as demonstrated by temporal expressions which inherently possess low ambiguity. The single-choice format presents exactly one correct option among the provided choices, while the multiple-choice format may contain one or more correct options. For multiple-choice questions, we expect evaluated models to comprehensively identify all correct options in their responses.
% TIME数据集中有3种不同形式的QA格式，包括\textbf{free-form}, \textbf{single-choice}和\textbf{multiple-choice}三种类型。Free-form是短答案生成的格式，例如“When did Nicola Agnozzi become an auxiliary bishop?”，答案为短答案形式"April 2, 1962."。Free-form格式的答案形式较为自由，只适用于同义回答类型较少的问题，如这个例子当中，答案是时间表达式，具有较少的歧义。Single-choice的答案只可能有一个正确选项；而multiple-choice的答案可能有1个或多个正确选项。对于multiple-choice的答案，我们希望被评估的模型能够完整地回答出正确的选项。

\begin{figure}
   \centering
   \includegraphics[width=\textwidth]{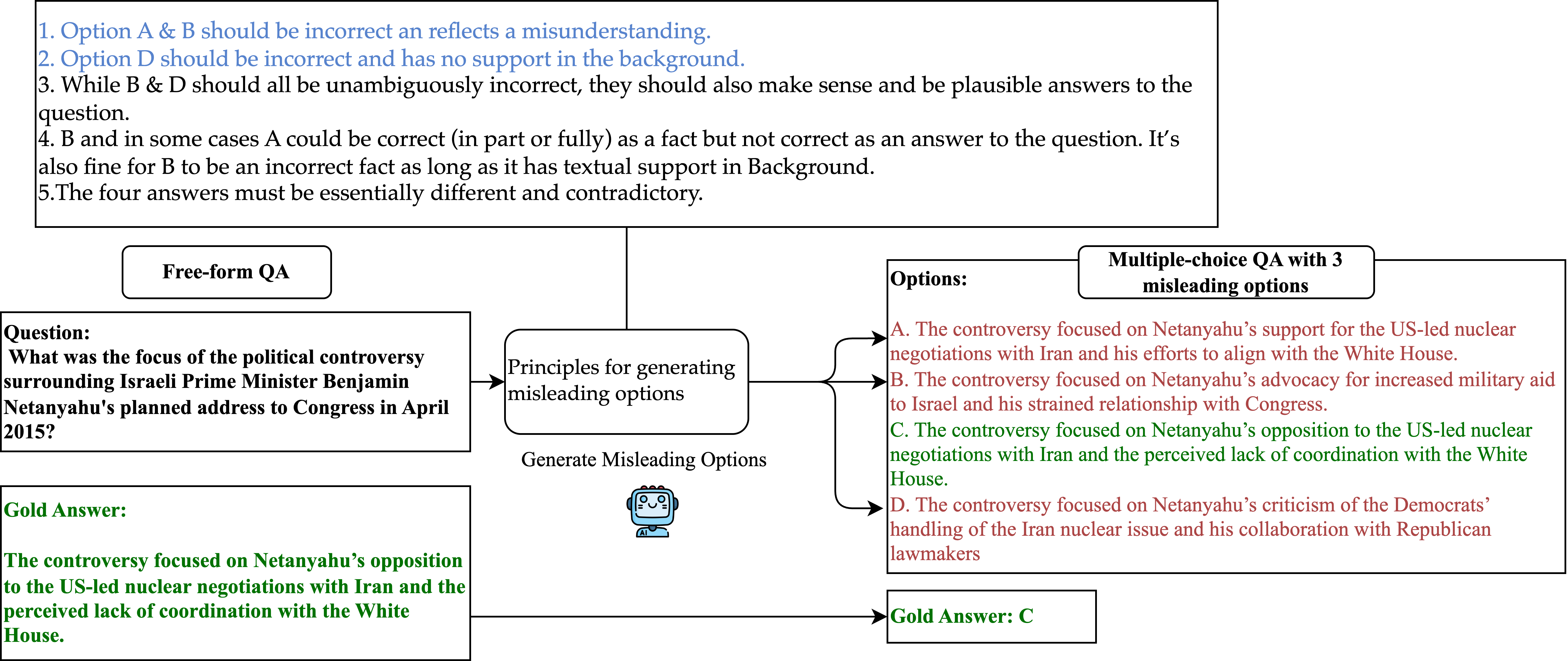}
   \caption{
      % 生成misleading option的过程示例。这是一个从\textsc{TIME-News}中关于Netanyahu's planned address to Congress in April 2015的TCE问题，属于explicit reasoning的问题类型。我们基于free-form形式的QA中的gold answer，按照我们预先设定的原则使用LLM生成misleading option，得到single-choice的QA形式。
      An illustration of the misleading option generation process. This example demonstrates a TCE question from \textsc{TimE-News} regarding Netanyahu's planned address to Congress in April 2015, which falls under the explicit reasoning task category. Based on the gold answer from the free-form QA format, we employ an LLM to generate misleading options following our predefined principles, resulting in a single-choice QA format.
   }
   \label{fig: misleading option generation}
\end{figure}

\paragraph{Misleading Options Generation\label{sec: appendix_misleading_options_generation}}
Building upon the STARC framework \cite{DBLP:conf/acl/BerzakML20misleadingqageneration} for generating diverse misleading options, we modify the principles for creating such distractors. The specific guidelines are illustrated in Figure \ref{fig: misleading option generation}. Each time we prompt the model, it generates three misleading options based on the given principles, the original question, and the gold answer. These options are designed to satisfy distinct types of misleading requirements according to the specified guidelines.

\begin{figure}
   \centering
   \includegraphics[width=\textwidth]{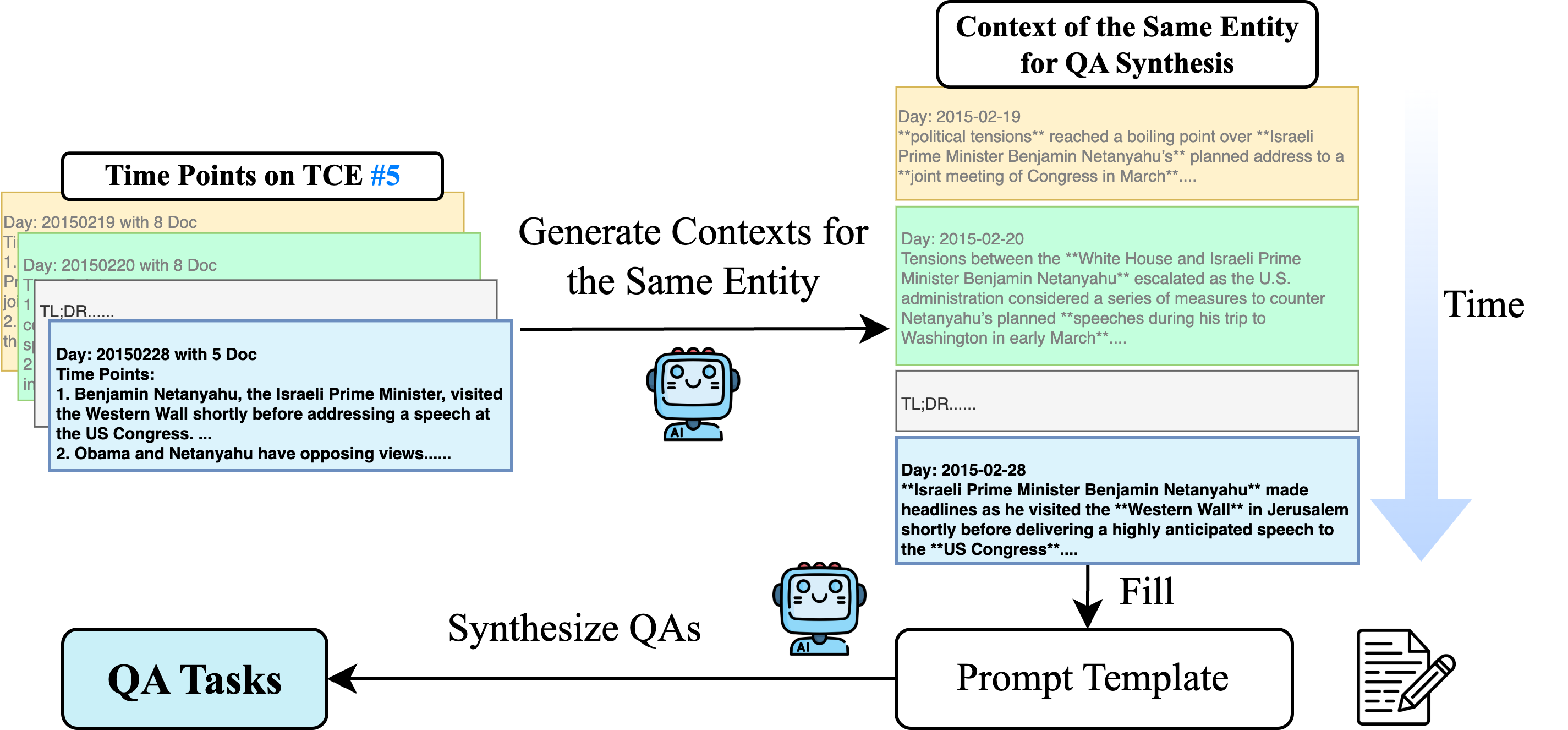}
   \caption{
      % \textsc{TIME-News}基于Time Points合成QA的过程示例。我们首先使用LLM生成对于the same entity的contexts，并组织成时间线contexts。然后，对于每一个QA任务生成时，选择其中的一些时间点对应的context fill到prompt template中，合成得到QAs。
      An illustration of the QA synthesis process based on time points in \textsc{TimE-News}. We first utilize an LLM to generate contexts for the same entity, which are then organized into timeline contexts. Subsequently, for each QA task generation, we select relevant time points and their corresponding contexts to populate the prompt template, thereby synthesizing the QAs.
   }
   \label{fig: a demo for TIME News synthesize QAs}
\end{figure}

% ---------------------------------补充材料1：合成QA的prompt模板----------------------------------
% \colorbox{teal!5}
% 下面是我们在合成每一个QA任务中的过程和prompt模板（以\textsc{TimE-Wiki}和\textsc{TimE-Dial}为例，\textsc{TimE-News}与其相似）。
\subsubsection{Level 1: Basic Temporal Understanding and Retrieval\label{sec: synthesize QA for Level 1}}

At this level, we primarily evaluate models' capabilities in temporal information retrieval and comprehension. We design tasks focusing on temporal information extraction, including directly extracting time expressions from context (\texttt{Extract}) and determining event occurrence times given specific events (\texttt{Localization}). Additionally, we assess fundamental temporal understanding through tasks such as duration calculation (\texttt{Computation}), duration comparison (\texttt{Duration Compare}), and event ordering (\texttt{Order Compare}).

\paragraph{Extract} For the \texttt{Extract} task, we first provide a set of authentic time points extracted from the source data. We then instruct the LLM to generate five novel time expressions that are distinct from all existing ones. Subsequently, we randomly select 0-4 authentic time points to form multiple-choice options, ensuring their randomness. This approach combines rule-based selection with LLM-generated distractors, thereby maintaining high QA quality. The task is formulated using a fixed question template: \textit{Which of the following are time expressions mentioned in the context? (Note: There may be one or more correct options. If you think NONE of the time expressions in options A/B/C/D are mentioned, then you can choose E. Do not choose E together with other options.)}

\begin{tcolorbox}[title= Prompt for generating fake time expressions (\texttt{Extract}) (\textsc{TimE-Wiki}), label={box: gen_fake_time}]
\textbf{[Rules:]} Given a list of time expressions, please generate FIVE new time expressions randomly. You should follow the instructions below:  \\
1. The 5 new time expressions should totally different from all of the the given time expressions. \\
2. Each new time expression should be in the format of <Month> <Day>, <Year>. For example: "May 4, 1998", "April 1992", "1934". \\
3. The 5 new time expressions should closely resemble the given time expression in format and structure, creating a high level of confusion, yet they must represent entirely different times. For instance, you could alter the year while keeping the month and day unchanged, such as changing May 2, 1922 to May 2, 1923; or you could modify the month and day while retaining the same year, for example, transforming May 2, 1922 into July 2, 1922. Additionally, you might consider changing the day while keeping the month and year consistent, like adjusting May 2, 1922 to May 3, 1922. Another approach could involve altering the month and year but keeping the day the same, such as changing May 2, 1922 to May 2, 1921. Lastly, you could shift the entire date by a consistent interval, for example, moving May 2, 1922 to June 3, 1923, ensuring that each change introduces a subtle yet distinct variation from the original time expression. \\
\textbf{[Example:]} Here are some examples showing the writing style. NOTE that the content of the examples are irrelevant to the question you will generate.  \\
\textbf{[Given time expressions:]} ['1656', '1763', '1782', '1784', '1787', '1815', 'September 20, 1817', 'January 20, 1848', ..., 'March 30, 2005', 'June 27, 1965', '1973', '2009', 'May 25, 1963', '1980', '2008', 'June 11, 1949', '1983', '1984', '2007', 'August 14, 2004', 'October 7, 2014', 'November 11, 2021', 'October 14, 2024', 'May 31, 1971', 'October 28, 1975'] \\
\textbf{[New 5 time expressions:]} ['1682', 'June 12, 1974', 'May 8, 1907', 'May 11, 1949', 'May 25, 2011'] \\
\textbf{\textit{Skipping 2 examples here......}} \\
\textbf{[Output:]} Now please write 5 new time expressions following the instructions and examples above. You should output the 5 new time expressions along with its answer, in the format of "["YY", "MM, YY", "MM DD, YY"]". NOTE that the time expressions should be in chronological order. Now the given time expressions are: \\
\textbf{[Given time expressions:]} \{Given\_time\_expressions\} \\
\textbf{[New 5 time expressions:]}
\end{tcolorbox}

\paragraph{Localization}
We select three facts from the temporal knowledge graph and employ LLMs to generate corresponding QA pairs.

\begin{tcolorbox}[title= Prompt for generating QA. (\texttt{Localization}) (\textsc{TimE-Wiki}), label={box: prompt_localization}]
\textbf{[Rules:]} Given 3 facts, please generate one question along with its gold answer for each given fact. You should follow the instructions below:  \\
1. The answer MUST be short and concise, avoiding using redundant words or repeating the information in the question.  \\
2. You should output the question and its answer without any other explanation, such as "Question: xxx? Answer: xxx."  \\
3. I will give you 3 facts in each line, and then you should output the question and its answer each line in the same sequence. So your output should be 3 lines of "Question: xxx? Answer: xxx.". \\
4. The question can be phrased in different ways, such as 'When is...?', 'What is the time for...?', 'What time did...?', and so on. \\
5. The answer should be "From xxx to xxx." or directly "xxx." \\
\textbf{[Example:]} Here are some examples showing the writing style. NOTE that the content of the examples are irrelevant to the question you will generate.  \\
* \textbf{[Given facts:]} \\
Debra Hamel worked at Wesleyan University from 1998 to 2001. \\
Oliver Marcy attended Wesleyan University in 1846.   \\
Debra Hamel completed her studies at Yale University in 1993.   \\
\textbf{[Generated QA:]} \\
Question: When did Debra Hamel work at Wesleyan University? Answer: From 1998 to 2001. \\
Question: What time did Oliver Marcy attend Wesleyan University? Answer: 1846. \\
Question: What is the time for Debra Hamel to complete her studies at Yale University? Answer: 1993. \\
\textbf{\textit{Skipping 1 example here......}\} \\
\textbf{[Output:]} Now please write a question following the instructions and examples above. You should output the question along with its answer, in the format of "Question: xxx? Answer: xxx.". NOTE that the answer should be "From xxx to xxx." or directly "xxx." \\
\textbf{[Given facts:]} \\
\{given\_facts} \\
\textbf{[Generated QA:]} \\
\end{tcolorbox}

\paragraph{Computation}
We provide multiple pairs of temporal facts and utilize LLMs to generate corresponding questions. The temporal computations are systematically derived through script-based rules to ensure accuracy and consistency.

\begin{tcolorbox}[title= Prompt for generating QA. (\texttt{Computation}) (\textsc{TimE-Wiki}), label={box: prompt_computation}]
\textbf{[Rules:]} Each time, I will provide you with several pairs of text snippets, with each pair occupying one line. For each line containing a pair of text snippets, you need to generate a question. You should follow the instructions below: \\
1. The question should be based on the snippet pair. \\
2. Each text snippet pair includes two snippets. Each snippet is composed of a fact and a 'Happen/Begin/End' time. You should translate the snippets into a format like: ... ... \\
3. Please refer to the following examples and learn the patterns well. \\
\textbf{[Example:]} Here are one example showing the writing style. NOTE that the content of the examples are irrelevant to the question you will generate.  \\
\textbf{[Snippet pairs:]} \\
José Manuel Pasquel served as an auxiliary bishop. Happen time: January 20, 1848.               José Manuel Pasquel became a priest. Happen time: September 20, 1817.\\
Antonius Grech Delicata Testaferrata was ordained. Happen time: October 19, 1845.               Diego Fabbrini played for Watford F.C.. Begin time: 2013.\\
Máximo Alcócer played for the Bolivia men's national football team. Begin time: 1957.       Diego Fabbrini played for Udinese Calcio. 1944.\\
\textbf{[Questions:]} \\
What was the duration from the time José Manuel Pasquel became a priest until he served as an auxiliary bishop?\\
How long was it between Antonius Grech Delicata Testaferrata was ordained and Diego Fabbrini began to play for Watford F.C.?\\
How much time passed from Máximo Alcócer began to play for the Bolivia men's national football team to Diego Fabbrini end playing for Udinese Calcio?\\
\textbf{[Instruction:]} \\
Now please write a question following the instructions and examples above. You should output the question only for each line. NOTE that there is NO any prefix like "Question:", just output the question string.\\
\textbf{[Snippet pairs:]}\\
\{snippet\_pairs\}\\
\textbf{[Questions:]}
\end{tcolorbox}

\paragraph{Duration Compare} 
The \texttt{Duration Compare} task is designed to evaluate models' capability in comparing the lengths of two time intervals. Specifically, we represent each time interval by the span between two distinct event timestamps. To construct questions for this task, we extract pairs of non-overlapping events from the timeline to form two comparable time intervals.
A sample question template is: \textit{"Which of the following two durations is longer? *Duration 1:* Between \{fact1\_1\} and \{fact1\_2\} *Duration 2:* Between \{fact2\_1\} and \{fact2\_2\}"}
For the \textsc{TimE-Wiki} benchmark, we have developed alternative question templates based on temporal facts, including: \textit{"Which fact lasted longer, Fact 1: \{fact1\} or Fact 2: \{fact2\}?", "Which of the two events, Fact 1: \{fact1\} or Fact 2: \{fact2\}, had a longer duration?", "Compare the duration of Fact 1: \{fact1\} and Fact 2: \{fact2\}. Which one was longer?"}

\paragraph{Order Compare}
% 这个任务是什么，这个数据是怎么做的？
The \texttt{Order Compare} task evaluates models' capability to comprehend temporal ordering between two time points. To avoid direct comparison of timestamps, we utilize event occurrence times as the comparison points. By leveraging a collection of temporal facts, we can directly select two facts and embed them into predefined question templates, as illustrated below.

\begin{tcolorbox}[title= Question Templates. (\texttt{Order Compare}) , label={box: question_templates_order_compare}]

"For Fact1: \{fact1\} and Fact2: \{fact2\}, which one happened earlier?"

"Which started earlier, Fact1: \{fact1\} or Fact2: \{fact2\}?"

"Which ended earlier, Fact1: \{fact1\} or Fact2: \{fact2\}?"

"Did Fact1: \{fact1\} start before Fact2: \{fact2\} ended?"

"Did Fact1: \{fact1\} end before Fact2: \{fact2\} started?"

"Did Fact1: \{fact1\} start before Fact2: \{fact2\} happened?"

\end{tcolorbox}

% TODO:下面\paragraph 3个子任务，并且分别对三个数据集的构造细节进行描述+prompt模板+QA结果示例
\subsubsection{Level 2: Temporal Expression Reasoning\label{sec: synthesize QA for Level 2}}
This level comprises three distinct subtasks that collectively assess the model's capability in performing multi-hop reasoning over temporal expressions. Each question can only be correctly answered if the LLM successfully conducts accurate multi-hop reasoning on the temporal expressions themselves.

\paragraph{Explicit Reasoning}
In this task, we employ temporal expressions composed of explicit time points in the questions. Notably, these explicit time points do not exist in the original context. We first randomly select temporal facts as ground truth answers, then transform their time points to generate non-existent temporal references. Finally, we utilize these modified time points to formulate questions. Below demonstrates the prompt template for generating \texttt{Explicit Reasoning} task questions in the \textsc{TimE-Wiki} dataset.

\begin{tcolorbox}[title= Generate Questions by Modifying Time Expressions (\texttt{Explicit Reasoning}) (\textsc{TimE-Wiki}), label={box: gen_qa_by_modify_time}]
\textbf{[Rules:]} Given the original questions and their corresponding updated time expressions, generate new questions by replacing the time expressions in the original questions. Follow these guidelines: \\
1. Modify only the time expressions; leave all other parts of the questions unchanged. \\
2. Output only the questions; do not include any answers. \\
3. Present each question on a separate line. \\
4. If the temporal expression in the given original question is a time point, such as "on September 9, 2002," then each time I will provide you with a time period expressed with "from ... to ...", for example, "from April 6, 1999 to June 2003." What you need to pay attention to is that your revised question should carry a tone of uncertainty. For instance, you should change "Which team did Ted play for on September 9, 2002?" to "Which team might Ted have played for from April 6, 1999 to June 2003?" \\
\textbf{[Example:]} Here are some examples showing the writing style. NOTE that the content of the examples are irrelevant to the question you will generate.  \\
\textbf{[Original Question:]}\\
What position did Alexandre da Sagrada Família hold on August 8, 1782?\\
\textbf{[New Time Expression:]}\\
from April 1772, to June 1784\\
\textbf{[New Question:]}\\
What position have Alexandre da Sagrada Família held from April 1772, to June 1784?\\
\textit{\textbf{Skipping some examples...}}\\ \\
\textbf{[Output:]} Now please write a question following the instructions and examples above. \\
\textbf{[Original Question:]} \\
\{original\_question\} \\
\textbf{[New Time Expression:]} \\
\{new\_time\_expression\} \\
\textbf{[New Question:]} \\
\end{tcolorbox}

\paragraph{Order Reasoning}
This task evaluates the model's comprehension of ordinal temporal expressions, such as "the second time serving as a professional basketball player" or "the last time attending a ballet performance." Correctly interpreting these expressions requires the LLM to fully understand the timeline to accurately identify the specific time point referenced. For task construction, we first establish a timeline for the same entity. Then, we identify sub-timelines sharing the same factual relationship and select the k-th fact in chronological order to generate questions. Below are two prompt templates for question generation in \textsc{TimE-Wiki}, given specific facts and their corresponding timeline orders.

\begin{tcolorbox}[title= Generate Question based on Subject-oriented Fact Timeline (\texttt{Order Reasoning}) (\textsc{TimE-Wiki}), label={box: gen_qa_subject_order_reasoning}]
\textbf{[Rules:]} Given a sentence describing a simple fact and an order number, please generate one question along with its answer. You should follow these instructions: \\
1. The question MUST target the subject in the factual statement. For example, given "Bruno Aguiar plays for Portugal national under-21 football team from 2001 to 2004." with order number "3", generate "Who is the third person affiliated with Portugal national under-21 football team?" with answer "Bruno Aguiar". \\
2. Formulate questions exclusively based on the provided factual content and numerical order. \\
3. Exclude temporal expressions from generated questions while maintaining factual integrity. \\
4. Craft unambiguous questions using diverse interrogative structures (e.g., "Which individual...", "What entity...") that require contextual analysis rather than lexical matching. \\
5. Ensure answers contain only the factual subject without explanatory content or repetition. \\
6. Present results strictly as: "Question: xxx? Answer: xxx." with continuous formatting(e.g. with no line breaks). \\
\textbf{[Example:]} Here are some examples showing the writing style. NOTE that the content of the examples are irrelevant to the question you will generate.  \\

\textbf{[Given fact:]}\\
Julius Babatunde Adelakun holds the position of diocesan bishop from April 13, 1973 to November 4, 2009.\\
\textbf{[Order number:]}\\
2\\
\textbf{[Generated QA:]}\\
Question: Who is the second person to hold the position of diocesan bishop? Answer: Julius Babatunde Adelakun.\\
\textbf{\textit{Skipping some examples here...}}
\\
\textbf{[Output:]} Now please write a question and its answer following the instructions and examples above. You should output the question along with its answer, in the format of "Question: xxx? Answer: xxx.". NOTE that the answer should be as short as possible. \\
\textbf{[Given fact:]} \\
\{given\_fact\} \\
\textbf{[Order number:]} \\
\{order\_number\} \\
\textbf{[Generated QA:]} \\
\end{tcolorbox}

\begin{tcolorbox}[title= Generate Question based on Object-oriented Fact Timeline (\texttt{Order Reasoning}) (\textsc{TimE-Wiki}), label={box: gen_qa_object_order_reasoning}]
\textbf{[Rules:]} Given a sentence describing a simple fact and an order number, please generate one question along with its answer. You should follow the instructions below:  \\
1. This question MUST be directed at the object in the fact. For example, given a fact "Bruno Aguiar plays for Portugal national under-21 football team from 2001 to 2004.", and an order number"3", your question should be "What was the third team Bruno Aguiar was affiliated with during her professional career?" and the answer should be "Portugal national under-21 football team". \\
2. The question should be derived directly from the factual content. \\
3. The question must exclude time expressions present in the original fact. \\
4. Phrase questions unambiguously using varied interrogative patterns (e.g., "Which team...", "What position...", "What organization...") while avoiding simple string matching. \\
5. The answer MUST contain only the factual object without explanations or repetitions. \\
6. Output strictly in the format: "Question: xxx? Answer: xxx." with no line breaks. \\
\textbf{[Example:]} Here are some examples showing the writing style. NOTE that the content of the examples are irrelevant to the question you will generate.  \\
\textbf{[Given fact:]}\\
Julius Babatunde Adelakun holds the position of diocesan bishop from April 13, 1973 to November 4, 2009.\\
\textbf{[Order number:]}\\
2
\textbf{[Generated QA:]}\\
Question: What was the second position Julius Babatunde Adelakun held in his role? Answer: diocesan bishop.\\
\\
\textbf{[Output:]} Now please write a question and its answer following the instructions and examples above. You should output the question along with its answer, in the format of "Question: xxx? Answer: xxx.". NOTE that the answer should be as short as possible. \\
\textbf{[Given fact:]} \\
\{given\_fact\} \\
\textbf{[Order number:]} \\
\{order\_number\} \\
\textbf{[Generated QA:]} \\
\end{tcolorbox}

\paragraph{Relative Reasoning}
This task evaluates models' reasoning capabilities with relative temporal expressions. For instance, "within the last two weeks before Trump's second official election as US President" and "within 7 months and 2 days after Xiao Ming's official graduation" demonstrate temporal reasoning based on specific event anchors. Such expressions pose significant challenges to LLMs' temporal reasoning abilities. 
To construct QA pairs for this task, we first provide ground truth temporal fact statements along with their corresponding relative temporal expressions. Notably, these relative temporal expressions are pre-extracted from all temporal facts, each consisting of a factual statement and a relative temporal expression. The QA generation methodology is detailed in the following prompt examples.

\begin{tcolorbox}[title= Generate Question based on Subject-oriented Fact Timeline (\texttt{Relative Reasoning}) (\textsc{TimE-Wiki}), label={box: relative_reasoning_gen_qa_template_subject}]
\textbf{[Rules:]} Given a sentence describing a simple fact and a time expression, please generate one question. You should follow the instructions below:  \\
1. The question MUST be directed at the subject in the fact. For example, given a fact "Bruno Aguiar plays for Portugal national under-21 football team.", and the time expression "Assuming today is January, 2002 | before today", your question should be "Assuming today is January, 2002, who was the most recently player that played for Portugal national under-21 football team before today?". \\
2. The question should come from the facts. \\
3. The question should be unambiguous and challenging, avoiding simple string matching. NO sub-questions allowed. \\
4. You should output the question without any other explanation. You should output the question directly. \\
\textbf{[Example:]} Here are some examples showing the writing style. NOTE that the content of the examples are irrelevant to the question you will generate.  \\

\textbf{Example 1:}\\
\textbf{[Given fact:]}\\
Julius Babatunde Adelakun holds the position of diocesan bishop.\\
\textbf{[Time Expression:]}\\
Assuming today is January, 2002 | before today\\
\textbf{[Generated Question:]}\\
Assuming today is January 2002, who was the most recent person that held the position of diocesan bishop before today?\\

\textbf{Example 2:}\\
\textbf{[Given fact:]}\\
Gerolamo Castaldi holds the position of diocesan bishop.\\
\textbf{[Time Expression:]}\\
before Francesco Marmaggi works as a Catholic priest\\
\textbf{[Generated Question:]}\\
Who was the most recent person to hold the position of diocesan bishop before Francesco Marmaggi works as a Catholic priest?\\

\textbf{Example 3:}\\
\textbf{[Given fact:]}\\
Wlodzimierz Roman Juszczak holds the position of diocesan bishop.\\
\textbf{[Time Expression:]}\\
before September 9, 2009 | within the span of 3 years 2 months 28 days\\
\textbf{[Generated Question:]}\\
Who was the most recent person to hold the position of diocesan bishop before September 9, 2009, within the span of 3 years 2 months 28 days?\\

\textbf{[Output:]} Now please write a question following the instructions and examples above. You should directly output the question.  \\
\textbf{[Given fact:]} \\
\{given\_fact\} \\
\textbf{[Time Expression:]} \\
\{time\_expression\} \\
\textbf{[Generated Question:]} \\
\end{tcolorbox}

\begin{tcolorbox}[title= Generate Question based on Object-oriented Fact Timeline (\texttt{Relative Reasoning}) (\textsc{TimE-Wiki}), label={box: relative_reasoning_gen_qa_template_object}]
\textbf{[Rules:]} Given a sentence describing a simple fact and a time expression, please generate one question. You should follow the instructions below:  \\
1. The question MUST be directed at the object in the fact. For example, given a fact "Bruno Aguiar plays for Portugal national under-21 football team.", and the time expression "Assuming today is January, 2002 | before today", your question should be "Assuming today is January, 2002, which team did Bruno Aguiar play for?". \\
2. The question should come from the facts. \\
3. The question should be unambiguous and challenging, avoiding simple string matching. NO sub-questions allowed. \\
4. You should output the question without any other explanation. You should output the question directly. \\
\textbf{[Example:]} Here are some examples showing the writing style. NOTE that the content of the examples are irrelevant to the question you will generate.  \\
\textbf{Example 1:}\\
\textbf{[Given fact:]}\\
Julius Babatunde Adelakun holds the position of diocesan bishop.\\
\textbf{[Time Expression:]}\\
Assuming today is January, 2002 | before today\\
\textbf{[Generated Question:]}\\
Assuming today is January, 2002, what position did Julius Babatunde Adelakun most recently hold before today?\\

\textbf{Example 2:}\\
\textbf{[Given fact:]}\\
Gerolamo Castaldi holds the position of diocesan bishop.\\
\textbf{[Time Expression:]}\\
before Francesco Marmaggi works as a Catholic priest\\
\textbf{[Generated Question:]}\\
What position did Gerolamo Castaldi most recently hold before Francesco Marmaggi worked as a Catholic priest?\\

\textbf{Example 3:}\\
\textbf{[Given fact:]}\\
Wlodzimierz Roman Juszczak holds the position of diocesan bishop.\\
\textbf{[Time Expression:]}\\
before September 9, 2009 | within the span of 3 years 2 months 28 days\\
\textbf{[Generated Question:]}\\
What position did Wlodzimierz Roman Juszczak most recently hold within the 3 years, 2 months, and 28 days prior to September 9, 2009?\\

\textbf{[Output:]} Now please write a question following the instructions and examples above. You should directly output the question.  \\
\textbf{[Given fact:]} \\
\{given\_fact\} \\
\textbf{[Time Expression:]} \\
\{time\_expression\} \\
\textbf{[Generated Question:]} \\
\end{tcolorbox}

% TODO：下面\paragraph 3个子任务，并且分别对三个数据集的构造细节进行描述+prompt模板+QA结果示例
\subsubsection{Level 3: Complex Temporal Relationship Reasoning\label{sec: synthesize QA for Level 3}}
At this level, we focus on evaluating the model's capability to comprehend both implicit and explicit temporal relationships between events. Specifically, we assess the model from three perspectives: (1) temporal co-occurrence between events (\texttt{Co-temporality}), (2) complete reordering of multiple distinct events (\texttt{Timeline}), and (3) the model's simultaneous understanding of both the original context and the question when temporal expressions in the question contradict the source text (\texttt{Counterfactual}).

\paragraph{Co-temporality}
% 这个任务是什么，这个数据是怎么去合成的？
This task evaluates the model's ability to comprehend temporal co-occurrence between two events. For instance, the question "When Sam Altman co-founded OpenAI, what positions did Elon Musk hold?" implicitly assumes the temporal overlap between "Sam Altman founding OpenAI" and "Elon Musk serving as co-founder of OpenAI and CEO of Tesla and SpaceX". To construct questions for this task, we provide the LLM with two key elements: a condition fact that serves as the temporal reference, and a query fact that forms the basis for question generation. The following demonstrates the prompt template.

\begin{tcolorbox}[title= Generate Question based on Subject-oriented Fact Timeline (\texttt{Co-temporality}) (\textsc{TimE-Wiki}), label={box: gen_qa_co_temporality_subject}]
\textbf{[Rules:]} Given a condition fact and a query fact, please generate one question. You should follow these instructions: \\
1. The question MUST target the subject in the factual statement. For example, given the condition fact "Mauro Morelli holds the position of diocesan bishop." with query fact "William Weigand holds the position of diocesan bishop.", generate "When Mauro Morelli holds the position of diocesan bishop, who held the position of diocesan bishop?" \\
2. The question should come from the facts. \\
3. The question should be unambiguous and challenging, avoiding simple string matching. NO sub-questions allowed. \\
4. You should output the question without any other explanation. You should output the question directly. \\
\textbf{[Example:]} Here are some examples showing the writing style. NOTE that the content of the examples are irrelevant to the question you will generate.  \\
\textbf{Example 1:}\\
\textbf{[Given fact:]}\\
Julius Babatunde Adelakun holds the position of diocesan bishop from April 13, 1973 to November 4, 2009.\\
\textbf{[Generated QA:]}\\
Question: Who served as diocesan bishop from April 13, 1973 to November 4, 2009? Answer: Julius Babatunde Adelakun.\\

\textit{Skipping the rest of the examples.}\\

\textbf{[Output:]} Now please write a question following the instructions and examples above. You should directly output the question.  \\
\textbf{[Condition fact:]} \\
\{condition\_fact\} \\
\textbf{[Query fact:]} \\
\{query\_fact\} \\
\textbf{[Generated Question:]} \\
\end{tcolorbox}

\begin{tcolorbox}[title= Generate Question based on Object-oriented Fact Timeline (\texttt{Co-temporality}) (\textsc{TimE-Wiki}), label={box: gen_qa_co_temporality_object}]
\textbf{[Rules:]} Given a condition fact and a query fact, please generate one question. You should follow these instructions: \\
1. The question MUST target the object in the factual statement. For example, given the condition fact "Mauro Morelli holds the position of diocesan bishop." with query fact "William Weigand holds the position of diocesan bishop.", generate "When Mauro Morelli held the position of diocesan bishop, what position did William Weigand hold?" \\
2. The question should come from the facts. \\
3. The question should be unambiguous and challenging, avoiding simple string matching. NO sub-questions allowed. \\
4. You should output the question without any other explanation. You should output the question directly. \\
\textbf{[Example:]} Here are some examples showing the writing style. NOTE that the content of the examples are irrelevant to the question you will generate.  \\
\textbf{Example 1:}\\
\textbf{[Condition fact:]}\\
Mauro Morelli holds the position of diocesan bishop.\\
\textbf{[Query fact:]}\\
William Weigand holds the position of diocesan bishop.\\
\textbf{[Generated Question:]}\\
When Mauro Morelli held the position of diocesan bishop, what position did William Weigand hold?\\

\textit{Skipping the rest of the examples.}\\

\textbf{[Output:]} Now please write a question following the instructions and examples above. You should directly output the question.  \\
\textbf{[Condition fact:]} \\
\{condition\_fact\} \\
\textbf{[Query fact:]} \\
\{query\_fact\} \\
\textbf{[Generated Question:]} \\
\end{tcolorbox}

\paragraph{Timeline}
% 这个任务是什么，这个数据是怎么去合成的？
The Timeline task is designed to evaluate a model's ability to chronologically reorder multiple facts within a given context. In \textsc{TimE-Wiki} and \textsc{TimE-Dial}, we assess the model's capability to determine temporal relationships among eight distinct facts, generating only one \texttt{Timeline} task question per timeline and context. In contrast, for \textsc{TimE}, we directly utilize multiple reordering questions from TCELongBench\cite{DBLP:conf/acl/Zhang00MLC24TCELongbench}.

For constructing \texttt{Timeline} task questions, we employ existing timelines containing eight temporal facts (without explicit temporal expressions or timestamps). Using a Python program, we first compute the chronological order of events, then randomly shuffle the order of the eight facts, with the correct temporal sequence serving as the ground truth. In this question construction process, we solely rely on question templates to generate the final output.

The question template is as follows:

\begin{tcolorbox}[title= Question Template. (\texttt{Timeline}) , label={box: question_templates_timeline}]
"Below are 8 facts. You need to sort these facts in chronological order. Requirements: Your output format must be numbers enclosed in parentheses without any other symbols or whitespace. For example: (1)(5)(2)(7)(3)(8)(6)(4)\{eight facts list\}"
\end{tcolorbox}

\paragraph{Counterfactual}
To thoroughly evaluate models' understanding of temporal relationships in context, we modify the \texttt{Explicit Reasoning} task by counterfactually altering temporal expressions in questions, making them contradict the temporal information in the provided real context. During evaluation, we instruct models to strictly adhere to the new temporal expressions in their responses. This approach eliminates direct reliance on contextual information (i.e., surface-level event-event correlations), enabling a fair assessment of models' genuine comprehension of temporal sequences.

Specifically, we construct task questions by directly modifying temporal expressions in \texttt{Explicit Reasoning} questions while preserving all other event details. After altering the temporal conditions, we employ a Python program with rule-based matching to determine whether the original answer remains valid under the new conditions. If the original answer no longer satisfies the new conditions, the correct answer becomes "There is no answer"; otherwise, the original gold answer remains unchanged.

We employ two distinct prompt strategies for question construction. The first prompt requires the model to generate a new temporal expression that replaces the original one while maintaining the same answer as before. The second prompt instructs the model to generate a new temporal expression that results in a different answer from the original.

\begin{tcolorbox}[title= Generate Question based on False Fact Premise and the new answer is the same as the original answer (\texttt{Counterfactual Question}) (\textsc{TimE-Wiki}), label={box: gen_q_right_answer_false_fact_premise}]

\textbf{[Scenario:]} You are an annotator who is exceptionally skilled at generating false temporal facts and premises. First, you carefully comprehend the question I provided you and its corresponding correct answer. Based on the question and answer, you cleverly imagine an "if" clause that represents a hypothesis. This hypothesis must contradict the facts in the given context but must simultaneously ensure that the provided answer remains correct. Your hypothesis only needs to modify the temporal elements to differ from the original, such as altering the year, month, or day. You need to add the imagined "if" clause to the original question and only output the question itself with the added "if" clause. \\
\textbf{[Example:]} Here are some examples showing the writing style. NOTE that the content of the examples are irrelevant to the question you will generate.   \\

\textbf{[Given info:]}\\
Question: Who might have worked as a Catholic priest before April 25, 1830?\\
Answer: Alexandre da Sagrada Família\\
\textbf{[Generated new question:]}\\
Who might have worked as a Catholic priest before April 25, 1830, if Alexandre da Sagrada Família worked as a Catholic priest in 1815?\\
\textbf{\textit{Skipping the rest of the examples...}}\\

\textbf{[NOTE:]}\\
Now please write a question following the instructions and examples above. You should ONLY output the question, and there should be no other output.\\
\textbf{[Given info:]}\\
Context: \{story\}\\
Question: \{question\}\\
Answer: \{answer\}\\
\textbf{[Generated new question:]}\\
\end{tcolorbox}

% ---------------------------------补充材料1：合成QA的prompt模板----------------------------------

\subsection{Quality Control\label{sec: quality control}}
% 数据筛选：我们从\textsc{TIME}数据集的每一个子数据集中，每一个任务使用随机数42，随机从3个子数据集中每一个任务分别抽取30-40条数据。最终，得到一共1071条QA数据用于标注，其中数据集\textsc{TIME-Wiki}一共352条，\textsc{TIME-News}一共359条，\textsc{TIME-Dial}一共360条。Detailes are provided in Appendix \ref{sec: quality control}.我们通过多个论坛发布人工标注招聘广告招募标注人员，并进行资格测试。
% We assembled a team of 8 professional annotators and selected the annotation results from the top 3 performers, based on both efficiency and quality metrics, as our final human evaluation benchmark.
% 最终达到了xxx的准确率。
% Finally, we achieved an annotation consistency score of xxx, ensuring high reliability of the data.

For data sampling, we employed a systematic approach to ensure representativeness across all sub-datasets. Using a fixed random seed of 42, we randomly sampled 30-40 QA pairs from each task within the three sub-datasets (\textsc{TimE-Wiki}, \textsc{TimE-News}, and \textsc{TimE-Dial}). This process yielded a total of 1,071 QA pairs for annotation, with 352 from \textsc{TimE-Wiki}, 359 from \textsc{TimE-News}, and 360 from \textsc{TimE-Dial} (detailed sampling procedures are provided in Appendix \ref{sec: quality control}). To ensure annotation quality, we recruited annotators through professional forums and conducted rigorous qualification tests. From the initial pool of 8 professional annotators, we selected the top 3 performers based on both efficiency and quality metrics to establish our final human evaluation benchmark.

\paragraph{Annotation Principles}
To ensure the quality of questions and answers, we established four annotation principles for evaluators to assess each QA pair systematically. For each data instance, we provide both the question and its corresponding gold answer. Evaluators are required to verify: (1) \texttt{\textbf{Question Answerability}} and (2) \texttt{\textbf{Answer Correctness}} (shown in ?) as fundamental criteria. Additionally, for the specialized \texttt{Timeline} and \texttt{Counterfactual} tasks, we introduced two specific evaluation principles: \texttt{\textbf{Correctness Answer Ranking}} and \texttt{\textbf{Counterfactual Conditional Answer Correctness}} respectively.

% TODO:这里画一个图 表示标注原则（应该可以放附录）

\begin{tcolorbox}[title=Checking Principle For \texttt{\textbf{Question Answerability}}, label={box: annotation_question_answerability}]
\textbf{Description:}  
Can the question be answered based on the provided context (without considering the standard answer provided above)?

\textbf{Options:}  

[A]: Yes, it can be directly found and answered in the context.

[B]: Yes, but the question can only be fully answered through inference.

[C]: The question cannot be answered based on the context; the context does not provide relevant information.

\end{tcolorbox}

\begin{tcolorbox}[title=Checking Principle For \texttt{\textbf{Answer Correctness}}, label={box: annotation_answer_correctness}]
\textbf{Description:}  
Please double-check, is the standard answer correct according to the original text?

\textbf{Options:}  

[A]: Yes, the answer is completely correct, answering the question fully and accurately.

[B]: Cannot be determined, but it has a certain possibility of being completely correct (perhaps you lack sufficient information to make a judgment).

[C]: The answer is not completely correct; it contains obvious errors or is irrelevant to the question.

\end{tcolorbox}

\begin{tcolorbox}[title=Checking Principle For \texttt{\textbf{Correctness Answer Ranking}}, label={box: annotation_correctness_answer_ranking}]
\textbf{Description:}  
Is the ranking of the answers correct? (Ranking: Is the order of events correct?)

\textbf{Options:}  

[A]: Yes, the order of all events is accurate.

[B]: Incorrect, the order of events is not completely correct.

\end{tcolorbox}

\begin{tcolorbox}[title=Checking Principle For \texttt{\textbf{Counterfactual Conditional Answer Correctness}}, label={box: annotation_counterfactual_conditional_answer_correctness}]
\textbf{Description:}  
Determine 1. whether the hypothetical premise in the given question contradicts the facts in the original context. 2. Regarding the hypothesis, if we disregard whether it contradicts the context and only consider the given question itself, is the given reference answer correct according to the given original context?

\textbf{Options:}  

[A]: Both 1 and 2 are met.

[B]: Only 1 is met, but 2 is not.

[C]: Neither 1 nor 2 is met.

\end{tcolorbox}

\paragraph{Worker Recruitment}
% worker的学历要求、英语要求、智力要求等等
% worker (1) had education experience in university with at least a bachelor’s degree; (2) passed the TOFEL, IELTS or 拥有可证明同等英语水平的资质；(3) good at using search engines to look up reliable information; (4) passed the cor- responding Qualification Test (QT) designed for our task (more details below). These requirements are to ensure that workers are proficient in English reading comprehension and can perform the temporal reasoning QA annotation task.
% worker的payment（每条数据薪资多少美金、每人平均工资多少美金？）：每人次人工标注基本工资110美元，平均每条数据标注支付金额约为0.103美金。

To ensure high-quality annotations, we established rigorous selection criteria for workers: (1) holding at least a bachelor's degree from accredited universities; (2) demonstrating English proficiency through TOEFL, IELTS, or equivalent certifications; (3) possessing strong information retrieval skills using search engines; and (4) successfully passing our customized Qualification Test (QT) for temporal reasoning tasks. These criteria guarantee annotators' competence in English comprehension and temporal reasoning capabilities. Regarding compensation, each worker received a base payment of \$110, with an average rate of \$0.103 per annotation instance.

\paragraph{Qualification Test}
To ensure the quality of temporal reasoning annotations, we meticulously designed a Qualification Test (QT) to assess workers' capabilities in comprehending temporal information and relationships across short, medium, and long contexts. Prior to administering the QT, we provided workers with comprehensive guidelines detailing the annotation requirements and evaluation criteria. Following the test administration, we manually reviewed all submissions to identify and exclude workers who demonstrated insufficient annotation proficiency. From the initial pool of 15 candidates, 8 successfully passed the QT and proceeded to the subsequent annotation phase.

% 为确保时间推理标注的质量，我们精心设计了资格测试(QT)来评估标注人员在短、中、长文本中理解时间信息和关系的能力。在测试前，我们向标注人员提供了详细的指导方针，说明标注要求和评估标准。测试结束后，我们手动审查了所有提交结果，筛选出标注能力不足的标注人员。在最初的15名候选人中，有8人成功通过QT并进入后续的标注阶段。

\paragraph{Two-stage Annotation}
% 第一阶段：收集标注员回答，其结果用于作为人类performance的代表。
% 第二阶段：标注员对每一条QA按照上述对应原则评估。如果标注员认为问题是不可回答的，可以针对answerability原则选择C选项，即无法从给定上下文中推断出任何可能的答案。
% 具体的UI界面放在附录，如图所示。

\begin{figure}
   \centering
   \includegraphics[width=\textwidth]{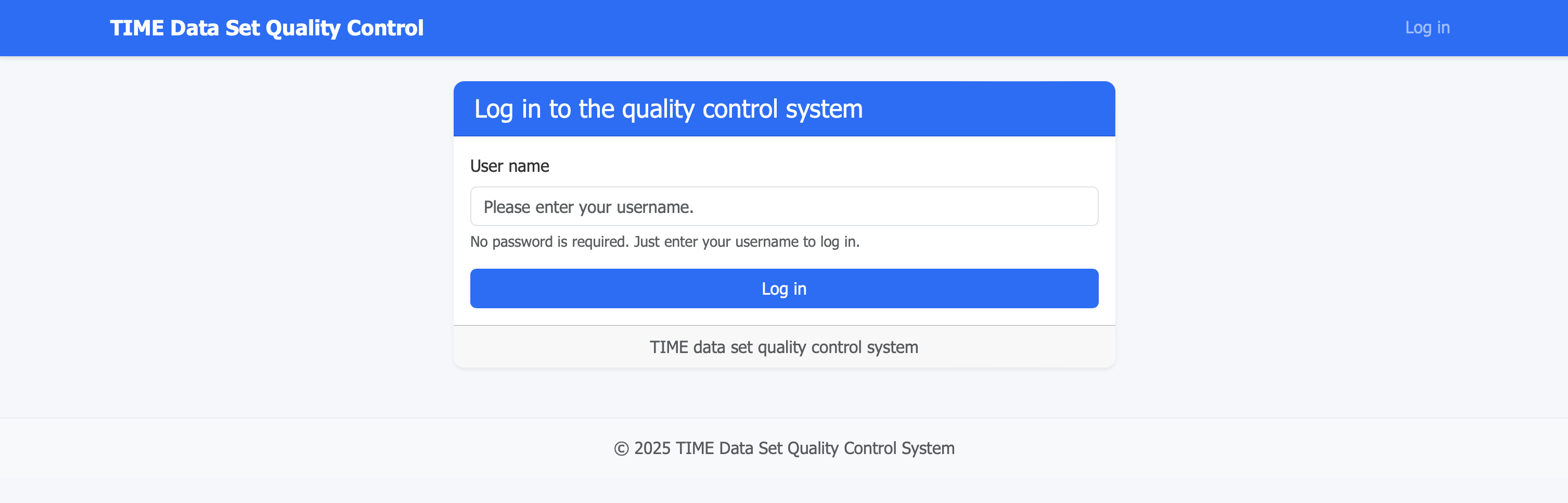}
   \caption{
   The login page of the annotation website.
   }
   \label{fig: login page of annotation website}
\end{figure}

\begin{figure}
   \centering
   \includegraphics[width=\textwidth]{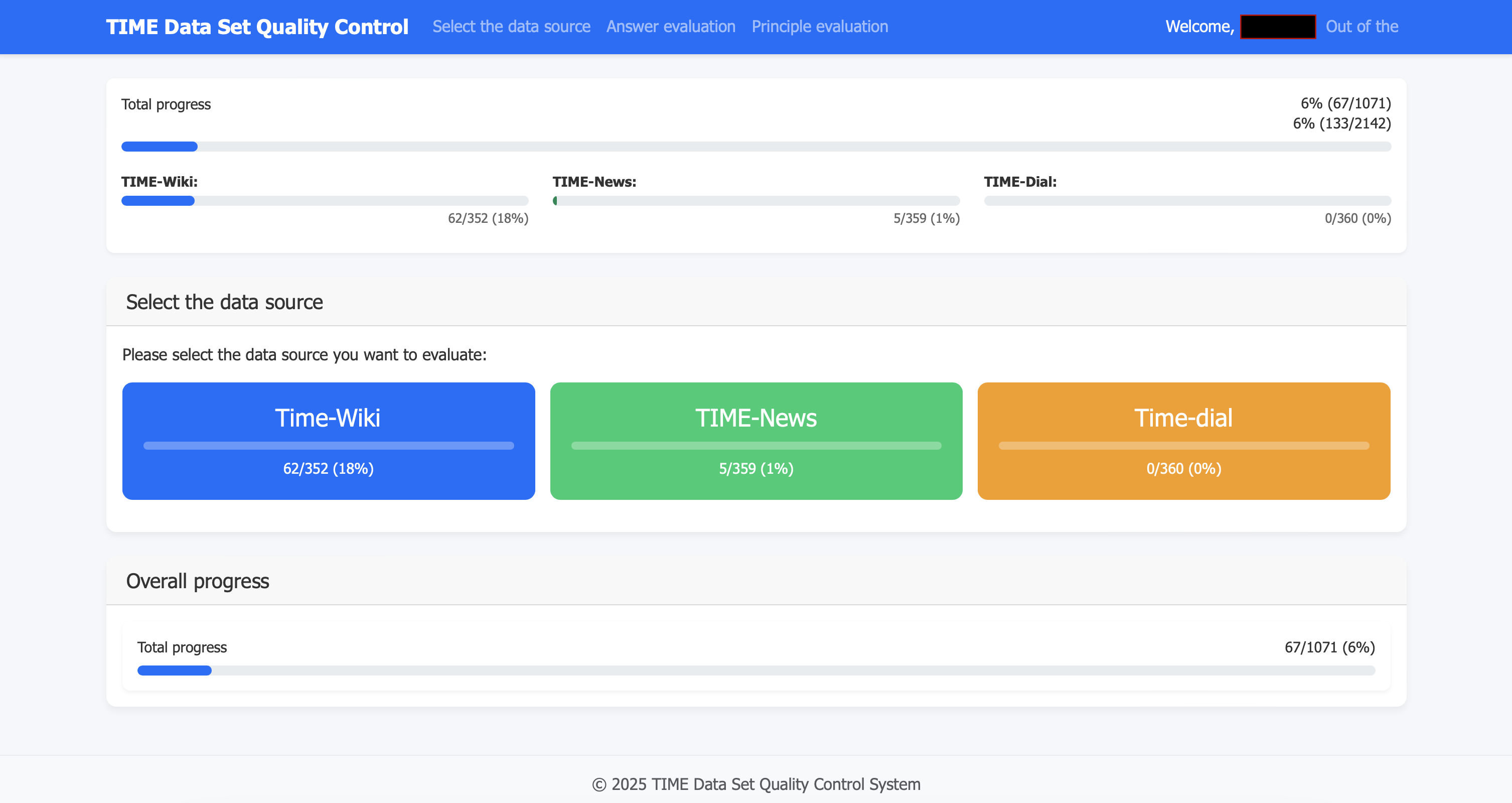}
   \caption{
   The overview page of the annotation website.
   }
   \label{fig: overview page of annotation website}
\end{figure}

\begin{figure}
   \centering
   \includegraphics[width=\textwidth]{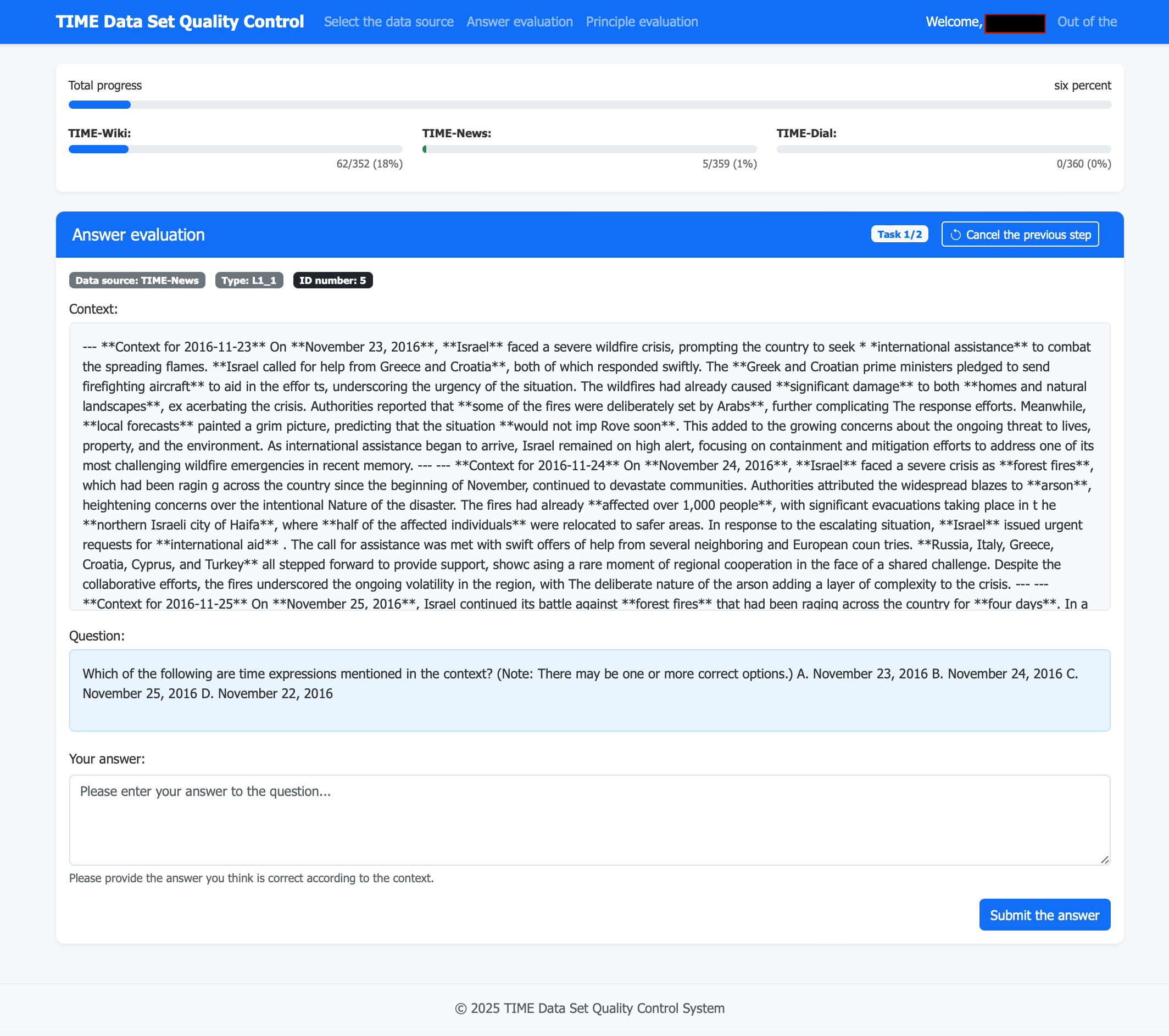}
   \caption{
   The first stage of data annotation.
   }
   \label{fig: first stage of annotation website}
\end{figure}

\begin{figure}
   \centering
   \includegraphics[width=\textwidth]{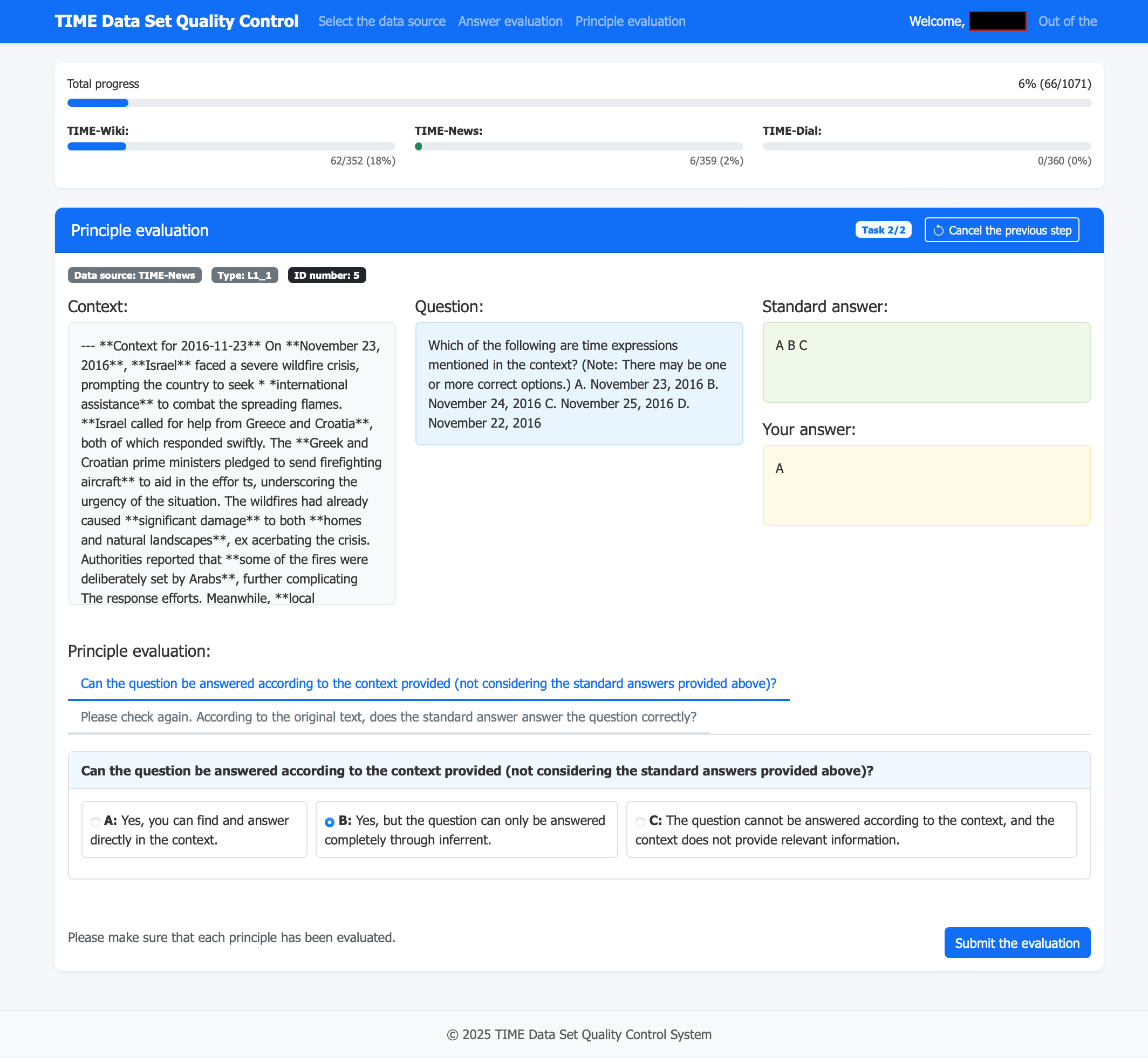}
   \caption{
   The second stage of data annotation.
   }
   \label{fig: second stage of annotation website}
\end{figure}

% 我们采用两阶段标注流程以确保数据质量。第一阶段，我们收集标注员的回答作为人类表现的基准。第二阶段，标注员依据预先制定的评估原则对每个QA对进行质量评估。当标注员认为问题无法回答时，可根据可回答性原则选择选项C，表示无法从给定上下文中推断出任何可能的答案。具体的标注界面设计详见附录图示。
We implemented a two-stage annotation process to ensure data quality. In the first stage, we collected annotators' responses as the human performance benchmark. In the second stage, annotators evaluated each QA pair based on predefined assessment principles. When annotators deemed a question unanswerable, they could select option C under the answerability principle, indicating that no possible answer could be inferred from the given context. The detailed annotation interface design is illustrated in Figure \ref{fig: login page of annotation website}, \ref{fig: overview page of annotation website}, \ref{fig: first stage of annotation website}, \ref{fig: second stage of annotation website}.

% 网页（这里记得用全文英文版）

\paragraph{Annotation Quality Control}
% TODO: 计算标注人员的一致性、Cohen’s Kappa (k)、Fleiss’ Kappa
% 由于我们的标注平台中，不管对于填空题和选择题均使用填空题格式进行人工答案标注，所以我们设计了新的标注方式来表征人工标注一致性。

% 我们采用词级别相似度（Word-level Similarity）作为标注一致性的评估指标。该方法通过计算两个标注者对同一问题回答在词汇层面上的重叠程度来量化标注一致性观该方法不仅能量化标注一致性的整体水平，还能捕捉不一致的程度差异，为标注质量评估提供了客观依据。
% 依据。。首先，我们从多位标注者的数据集中自动配对同路径JSON文件。对于每对文件，我们提取相同问题ID对应的标注答案（Human Prediction Answer）。词级别相似度计算基于集合理论中的Jaccard相似系数，定义为：
% $$Sim(A, B) = \frac{|A \cap B|}{|A \cup B|}$$
% 该方法的值域为[0,1]，其中1表示完全一致，0表示完全不一致。
% 基于词级别相似度，我们计算以下评估指标：
% 平均词级一致性：所有问题对应答案的平均词级相似度
% 完全匹配率：词级相似度为1.0的样本占比
% 一致性分布：按相似度区间[0-0.2, 0.2-0.4, 0.4-0.6, 0.6-0.8, 0.8-1.0]统计样本分布
% 四分位数：词级相似度的Q1、中位数、

Since our annotation platform uniformly employs a fill-in-the-blank format for manual answer annotation across both cloze and multiple-choice questions, we designed a novel annotation scheme to characterize inter-annotator agreement.

We adopt word-level similarity as the evaluation metric for annotation consistency. This method quantifies agreement by computing the lexical overlap between two annotators' responses to the same question. The approach not only measures the overall consistency level but also captures gradations of disagreement, thereby providing an objective basis for assessing annotation quality.

First, we automatically pair JSON files following identical paths from multiple annotators' datasets. For each file pair, we extract the manually annotated answers (Human Prediction Answer) corresponding to the same question ID. The word-level similarity is calculated using the Jaccard similarity coefficient from set theory, defined as:
$$Sim(A, B) = \frac{|A \cap B|}{|A \cup B|}$$
This metric yields values in the range [0, 1], where 1 indicates perfect agreement and 0 denotes complete disagreement.
Based on word-level similarity, we compute the following evaluation metrics:
Average Word-level Agreement: The mean word-level similarity across all question-answer pairs
Exact Match Rate: The proportion of samples with perfect word-level similarity (1.0)
Agreement Distribution: The sample distribution across similarity intervals: [0-0.2, 0.2-0.4, 0.4-0.6, 0.6-0.8, 0.8-1.0]
Quartiles: The Q1, median, and Q3 values of word-level similarity.

\begin{figure}
   \centering
   \includegraphics[width=\textwidth]{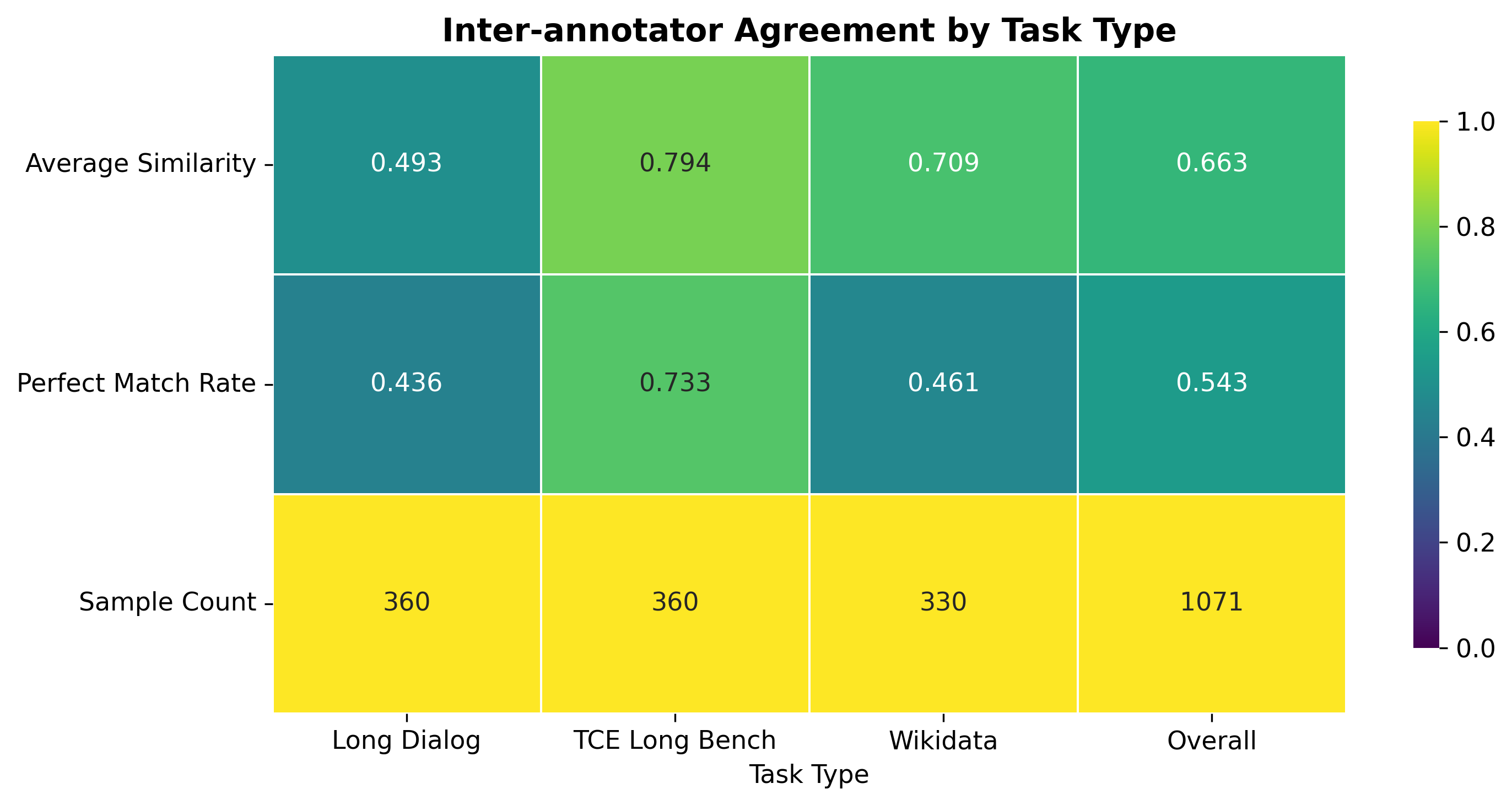}
   \caption{
   Task Type Comparison Heatmap: A comparison of annotation consistency across three distinct task types (\textsc{TimE-Wiki}, \textsc{TimE-News}, and \textsc{TimE-Dial}). The heatmap employs a color gradient from dark to light to visually represent the differences in average similarity and exact match rates among the task types.
   }
   \label{fig: task_comparison_heatmapsecond stage of annotation website}
\end{figure}

\begin{figure}
   \centering
   \includegraphics[width=\textwidth]{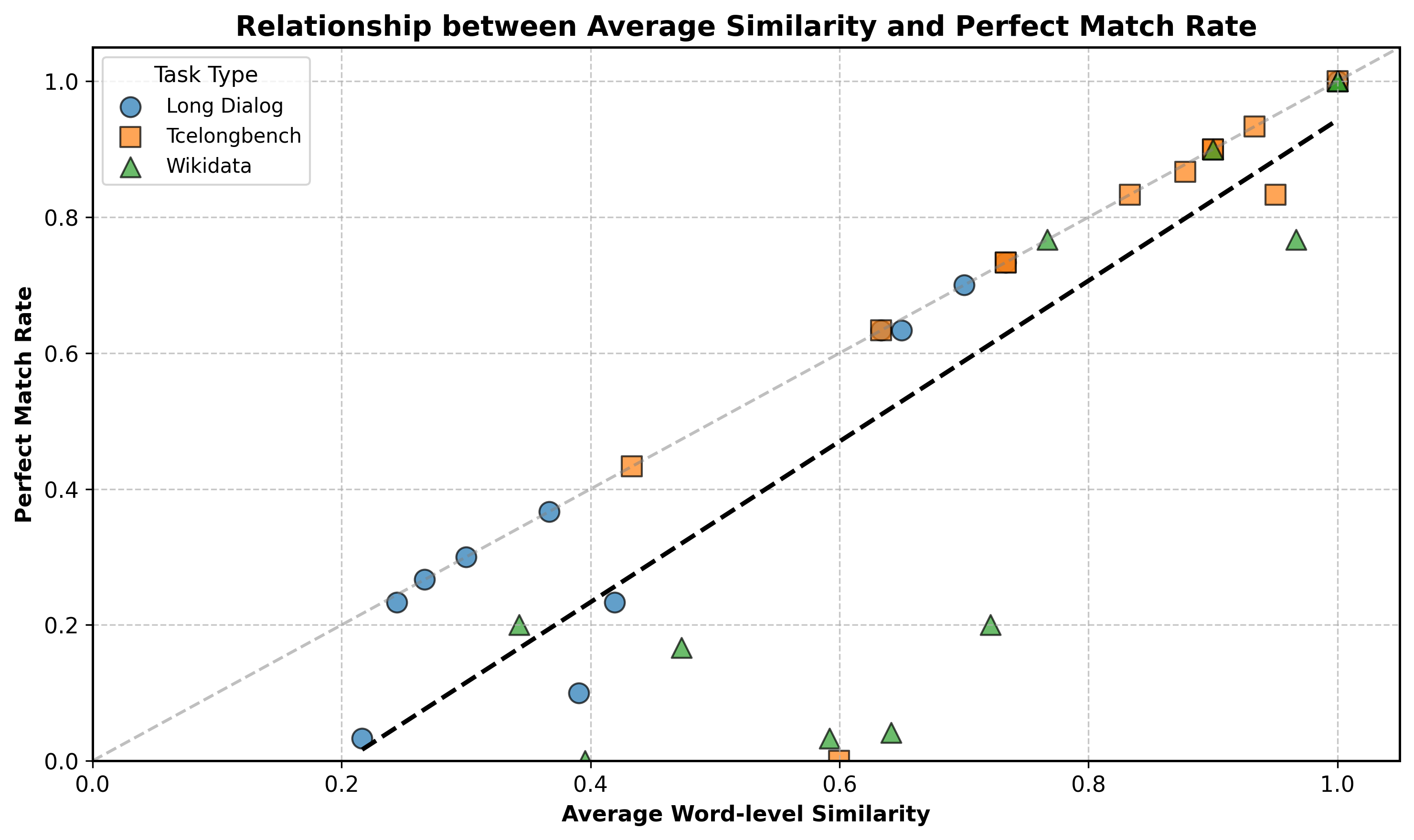}
   \caption{
   Scatter Plot of Average Similarity vs. Exact Match Rate:
This plot illustrates the relationship between the average token-level similarity and the exact match rate for each file, with different colors and markers distinguishing different task types. A regression line is added to indicate the correlation between the two metrics.
   }
   \label{fig: similarity_vs_perfect_match}
\end{figure}

\subsection{Dataset Statistics\label{sec: appendix_dataset_statistics}}
We present comprehensive statistics of instance counts for each subtask in both \textsc{TimE} and \textsc{TimE-Lite} datasets, as detailed in Table \ref{tab:complete_dataset_stats}.
\begin{table*}[htbp]
\centering
\caption{Dataset statistics. The table displays the number of instances for each dataset and task category.
Task abbreviations are: Ext. (Extract), Loc. (Localization), Comp. (Computation), D.C. (Duration Compare), O.C. (Order Compare);
E.R. (Explicit Reasoning), O.R. (Order Reasoning), R.R. (Relative Reasoning);
C.T. (Co-temporality), T.L. (Timeline), C.F. (Counterfactual).}
\label{tab:complete_dataset_stats}
\resizebox{\textwidth}{!}{%
\begin{tabular}{@{}lrrrrrrrrrrrr@{}}
\toprule
\textbf{Dataset} & \multicolumn{1}{c}{\textbf{All Tasks}} & \multicolumn{5}{c}{\textbf{Level 1}} & \multicolumn{3}{c}{\textbf{Level 2}} & \multicolumn{3}{c}{\textbf{Level 3}} \\
\cmidrule(lr){3-7} \cmidrule(lr){8-10} \cmidrule(lr){11-13}
 & & \multicolumn{1}{c}{Ext.} & \multicolumn{1}{c}{Loc.} & \multicolumn{1}{c}{Comp.} & \multicolumn{1}{c}{D.C.} & \multicolumn{1}{c}{O.C.} & \multicolumn{1}{c}{E.R.} & \multicolumn{1}{c}{O.R.} & \multicolumn{1}{c}{R.R.} & \multicolumn{1}{c}{C.T.} & \multicolumn{1}{c}{T.L.} & \multicolumn{1}{c}{C.F.} \\
\midrule
\textbf{\textsc{TimE}} & \textbf{38522} &  1480     &   3546   &   3376   &   3401    &  3549    &   3537    &   3538    &   3537    &   3513   &  5508    &   3537   \\
\hspace{1em}\textsc{TimE-Wiki} & 13848 & 1261 & 1299 & 1126 & 1151 & 1299 & 1287 & 1288 & 1287 & 1263 & 1300 & 1287 \\
\hspace{1em}\textsc{TimE-News} & 19958 & 0    & 1800 & 1800 & 1800 & 1800 & 1800 & 1800 & 1800 & 1800 & 3758 & 1800 \\
\hspace{1em}\textsc{TimE-Dial} & 4716  & 219  & 447  & 450  & 450  & 450  & 450  & 450  & 450  & 450  & 450  & 450  \\
\midrule
\textbf{\textsc{TimE-Lite}} & \textbf{943} &   60    &   90   &  78    &  86     &  90    &  90     &   90    &   90    &  90    &  89    &  90    \\
\hspace{1em}\textsc{TimE-Lite-Wiki} & 322 & 30   & 30   & 24   & 28   & 30   & 30   & 30   & 30   & 30   & 30   & 30   \\
\hspace{1em}\textsc{TimE-Lite-News} & 299 & 0    & 30   & 30   & 30   & 30   & 30   & 30   & 30   & 30   & 29   & 30   \\
\hspace{1em}\textsc{TimE-Lite-Dial} & 322 & 30   & 30   & 24   & 28   & 30   & 30   & 30   & 30   & 30   & 30   & 30   \\
\bottomrule
\end{tabular}%
}
\end{table*}

\begin{table*}[htbp]
   \centering
   \caption{Average context token counts for each dataset and task category. Token counts are derived using the `cl100k\_base` encoder from the `tiktoken` library. Task abbreviations are: Ext. (Extract), Loc. (Localization), Comp. (Computation), D.C. (Duration Compare), O.C. (Order Compare); E.R. (Explicit Reasoning), O.R. (Order Reasoning), R.R. (Relative Reasoning); C.T. (Co-temporality), T.L. (Timeline), C.F. (Counterfactual). A dash (---) indicates that data was not available for the corresponding combination. Note that for \textsc{TimE-News} and \textsc{TimE-Lite-News}, the context token counts represent the average token counts of the top-3 chunks retrieved by three distinct retrievers.
   }
   \label{tab:avg_token_counts_by_task_dataset}
   \resizebox{\textwidth}{!}{%
   \begin{tabular}{@{}lrrrrrrrrrrr@{}}
   \toprule
   \textbf{Dataset} & \multicolumn{5}{c}{\textbf{Level 1}} & \multicolumn{3}{c}{\textbf{Level 2}} & \multicolumn{3}{c}{\textbf{Level 3}} \\
   \cmidrule(lr){2-6} \cmidrule(lr){7-9} \cmidrule(lr){10-12}
    & \multicolumn{1}{c}{Ext.} & \multicolumn{1}{c}{Loc.} & \multicolumn{1}{c}{Comp.} & \multicolumn{1}{c}{D.C.} & \multicolumn{1}{c}{O.C.} & \multicolumn{1}{c}{E.R.} & \multicolumn{1}{c}{O.R.} & \multicolumn{1}{c}{R.R.} & \multicolumn{1}{c}{C.T.} & \multicolumn{1}{c}{T.L.} & \multicolumn{1}{c}{C.F.} \\
   \midrule
   \textbf{\dataset} \\
   \hspace{1em}\textsc{TimE-Wiki} & 1157.39 & 1156.37 & 1159.06 & 1188.26 & 1156.37 & 1155.05 & 1150.47 & 1150.58 & 1158.69 & 1156.00 & 1155.05 \\
   \hspace{1em}\textsc{TimE-News} & \multicolumn{1}{c}{---} & 1473.98 & 1568.53 & 1527.80 & 1499.17 & 1502.02 & 1511.58 & 1496.59 & 1441.80 & 1561.75 & 1515.70 \\
   \hspace{1em}\textsc{TimE-Dial} & 20862.69 & 20709.06 & 20667.67 & 20667.67 & 20667.67 & 20667.67 & 20667.67 & 20667.67 & 20667.67 & 20667.67 & 20667.67 \\
   \midrule
   \textbf{\textsc{TimE-Lite}} \\
   \hspace{1em}\textsc{TimE-Lite-Wiki} & 1332.37 & 1332.37 & 1284.88 & 1328.86 & 1332.37 & 1332.37 & 1332.37 & 1332.37 & 1332.37 & 1332.37 & 1332.37 \\
   \hspace{1em}\textsc{TimE-Lite-News} & \multicolumn{1}{c}{---} & 1498.64 & 1572.68 & 1500.36 & 1425.72 & 1361.89 & 1494.90 & 1458.50 & 1490.09 & 1523.03 & 1423.47 \\
   \hspace{1em}\textsc{TimE-Lite-Dial} & 21665.93 & 21665.93 & 21665.93 & 21665.93 & 21665.93 & 21665.93 & 21665.93 & 21665.93 & 21665.93 & 21665.93 & 21665.93 \\
   \bottomrule
   \end{tabular}%
   }
   \end{table*}
% ------------------补充材料2:context平均长度的表格----------------------

% % Appendix B
% -----------------------补充材料3：QA示例--------------------
\section{Benchmark Details}
% TODO:
% 1.每一个任务在三个数据集中的示例（图片）
% 2.数据源本身的分析
% 2.1具体的每一个QA问题的特征分析？
% 2.2时间表达式的分布情况？跨度长短的分布情况？时间粒度的分布情况？

% QA examples
\subsection{QA Examples}
\subsubsection{\textsc{TimE-Wiki}}

\begin{tcolorbox}[title= QA Example. (\texttt{Extract}) (\textsc{TimE-Wiki}) (Level 1: Basic Temporal Understanding and Retrieval) , label={box: 1-1-wiki}]
\textbf{Question:}\\
Which of the following are time expressions mentioned in the context? (Note: There may be one or more correct options. If you think NONE of the time expressions in options A/B/C/D are mentioned, then you can choose E. Do not choose E together with other options.)\\A. 1930\\B. 2011\\C. 1929\\D. 1932\\E. None of the above.\\
\textbf{Answer:}\\
A B
\end{tcolorbox}

\begin{tcolorbox}[title= QA Example. (\texttt{Localization}) (\textsc{TimE-Wiki}) (Level 1: Basic Temporal Understanding and Retrieval) , label={box: 1-2-wiki}]
\textbf{Question:}\\
When did Nicola Agnozzi become an auxiliary bishop?\\
\textbf{Answer:}\\
April 2, 1962
\end{tcolorbox}

\begin{tcolorbox}[title= QA Example. (\texttt{Computation}) (\textsc{TimE-Wiki}) (Level 1: Basic Temporal Understanding and Retrieval) , label={box: 1-3-wiki}]
\textbf{Question:}\\
What was the duration from the time George Omaira began to serve as an auxiliary bishop until Nicola Agnozzi became an auxiliary bishop?   (Hint: Please answer in the form of Month Day, Year. e.g. 1 year 2 months 3days, or 2 days, or 9 months, or 3 months 15 days.)\\
\textbf{Answer:}\\
362 years
\end{tcolorbox}

\begin{tcolorbox}[title= QA Example. (\texttt{Duration Compare}) (\textsc{TimE-Wiki}) (Level 1: Basic Temporal Understanding and Retrieval) , label={box: 1-4-wiki}]
\textbf{Question:}\\
Which of the two events, Fact 1: George Omaira served as an auxiliary bishop. or Fact 2: Maxim Hermaniuk served as an auxiliary bishop., had a longer duration?\\A. Fact 1 lasts longer.\\B. Fact 2 lasts longer.\\C. They last almost the same amount of time.\\
\textbf{Answer:}\\
A
\end{tcolorbox}

\begin{tcolorbox}[title= QA Example. (\texttt{Order Compare}) (\textsc{TimE-Wiki}) (Level 1: Basic Temporal Understanding and Retrieval) , label={box: 1-5-wiki}]
\textbf{Question:}\\
For Fact1: Mauro Morelli was appointed as both a titular bishop and an auxiliary bishop. and Fact2: Pedro Bantigue y Natividad was appointed as an auxiliary bishop., which one happened earlier?\\A. Fact 1 happened earlier.\\B. Fact 2 happened earlier.\\C. They happen at almost the same time.\\
\textbf{Answer:}\\
B
\end{tcolorbox}

\begin{tcolorbox}[title= QA Example. (\texttt{Explicit Reasoning}) (\textsc{TimE-Wiki}) (Level 2: Temporal Expression Reasoning) , label={box: 2-1-wiki}]
\textbf{Question:}\\
Who might have held the position of auxiliary bishop from 1902 to February 31, 1916?\\
\textbf{Answer:}\\
Edward Joseph Hanna
\end{tcolorbox}

\begin{tcolorbox}[title= QA Example. (\texttt{Order Reasoning}) (\textsc{TimE-Wiki}) (Level 2: Temporal Expression Reasoning) , label={box: 2-2-wiki}]
\textbf{Question:}\\
What was the third position Mauro Morelli held in his role?\\
\textbf{Answer:}\\
auxiliary bishop
\end{tcolorbox}

\begin{tcolorbox}[title= QA Example. (\texttt{Relative Reasoning}) (\textsc{TimE-Wiki}) (Level 2: Temporal Expression Reasoning) , label={box: 2-3-wiki}]
\textbf{Question:}\\
Who was the most recent person to hold the position of titular bishop before Mauro Morelli stepped down as diocesan bishop?\\
\textbf{Answer:}\\
José Antonio Eguren
\end{tcolorbox}

\begin{tcolorbox}[title= QA Example. (\texttt{Co-temporality}) (\textsc{TimE-Wiki}) (Level 3: Complex Temporal Relationship Reasoning) , label={box: 3-1-wiki}]
\textbf{Question:}\\
When José Antonio Eguren held the position of auxiliary bishop, what position did Mauro Morelli hold?\\
\textbf{Answer:}\\
diocesan bishop
\end{tcolorbox}

\begin{tcolorbox}[title= QA Example. (\texttt{Timeline}) (\textsc{TimE-Wiki}) (Level 3: Complex Temporal Relationship Reasoning) , label={box: 3-2-wiki}]
\textbf{Question:}\\
Below are 8 facts. You need to sort these facts in chronological order. Requirements: Your output format must be numbers enclosed in parentheses without any other symbols or whitespace. For example: (1)(5)(2)(7)(3)(8)(6)(4)\\ \\(1) Mauro Morelli holds the position of auxiliary bishop.\\(2) Mauro Morelli holds the position of titular bishop.\\(3) Jean-Claude Miche holds the position of titular bishop.\\(4) Belchior Carneiro Leitão holds the position of titular bishop.\\(5) Timothy Norton holds the position of titular bishop.\\(6) Nicola Agnozzi holds the position of titular bishop.\\(7) Edward Joseph Hanna holds the position of titular bishop.\\(8) John Joseph Swint holds the position of auxiliary bishop.\\
\textbf{Answer:}\\
(4)(3)(7)(8)(6)(2)(1)(5)
\end{tcolorbox}

\begin{tcolorbox}[title= QA Example. (\texttt{Counterfactual}) (\textsc{TimE-Wiki}) (Level 3: Complex Temporal Relationship Reasoning) , label={box: 3-3-wiki}]
\textbf{Question:}\\
Which institutions might Victorina Durán have attended from 1915 to 1934, if she graduated from the Madrid Royal Conservatory in 1915?\\
\textbf{Answer:}\\
Royal Academy of Fine Arts of San Fernando
\end{tcolorbox}

\subsubsection{\textsc{TimE-News}}

% 10个子任务

\begin{tcolorbox}[title= QA Example. (\texttt{Localization}) (\textsc{TimE-News}) (Level 1: Basic Temporal Understanding and Retrieval) , label={box: 1-2-news}]
\textbf{Question:}\\
When is Israeli Prime Minister Benjamin Netanyahu scheduled to address the US Congress?\\
\textbf{Answer:}\\
March 3, 2015
\end{tcolorbox}

\begin{tcolorbox}[title= QA Example. (\texttt{Computation}) (\textsc{TimE-News}) (Level 1: Basic Temporal Understanding and Retrieval) , label={box: 1-3-news}]
\textbf{Question:}\\
How many days passed between the initial opposition to Netanyahu’s speech on February 19, 2015, and the announcement of his address to Congress on February 27, 2015?   (Hint: Please answer in the form of Month Day, Year. e.g. 1 year 2 months 3days, or 2 days, or 9 months, or 3 months 15 days.)\\
\textbf{Answer:}\\
8 days
\end{tcolorbox}

\begin{tcolorbox}[title= QA Example. (\texttt{Duration Compare}) (\textsc{TimE-News}) (Level 1: Basic Temporal Understanding and Retrieval) , label={box: 1-4-news}]
\textbf{Question:}\\
Which of the following two durations is longer? *Duration 1:* The time span between when political tensions escalated over Netanyahu’s planned address to Congress and when the White House announced measures to counter Netanyahu’s speeches. *Duration 2:* The time span between when the White House considered limiting communication with Israel and when the White House began strategizing a response to Netanyahu’s upcoming speech.  \\A. Duration 1 is longer.  \\B. Duration 2 is longer.  \\C. The two durations are approximately the same length.
\textbf{Answer:}\\
B
\end{tcolorbox}

\begin{tcolorbox}[title= QA Example. (\texttt{Order Compare}) (\textsc{TimE-News}) (Level 1: Basic Temporal Understanding and Retrieval) , label={box: 1-5-news}]
\textbf{Question:}\\
For Fact1: Democratic Party representatives echoed calls to postpone Netanyahu's address. and Fact2: The White House explored strategies to counter Netanyahu's speeches during his trip to Washington., which one happened earlier?  \\A. Fact 1 happened earlier.  \\B. Fact 2 happened earlier.  \\C. They happen at almost the same time.\\
\textbf{Answer:}\\
A
\end{tcolorbox}

\begin{tcolorbox}[title= QA Example. (\texttt{Explicit Reasoning}) (\textsc{TimE-News}) (Level 2: Temporal Expression Reasoning) , label={box: 2-1-news}]
\textbf{Question:}\\
What was the focus of the political controversy surrounding Israeli Prime Minister Benjamin Netanyahu's planned address to Congress in April 2015?\\A.  The controversy focused on Netanyahu’s support for the US-led nuclear negotiations with Iran and his efforts to align with the White House.  \\B.  The controversy focused on Netanyahu’s advocacy for increased military aid to Israel and his strained relationship with Congress.\\C.  The controversy focused on Netanyahu’s opposition to the US-led nuclear negotiations with Iran and the perceived lack of coordination with the White House.\\D.  The controversy focused on Netanyahu’s criticism of the Democrats’ handling of the Iran nuclear issue and his collaboration with Republican lawmakers
\textbf{Answer:}\\
C

\end{tcolorbox}

\begin{tcolorbox}[title= QA Example. (\texttt{Order Reasoning}) (\textsc{TimE-News}) (Level 2: Temporal Expression Reasoning) , label={box: 2-2-news}]
\textbf{Question:}\\
What was the first action taken by liberal Democrats in response to Netanyahu’s planned address to Congress?\\A.  They issued a public statement criticizing the speech.  \\B.  They held a press conference to announce their support for the speech.\\C.  They signed a letter requesting the delay of the speech.\\D.  They organized a protest against the speec\\
\textbf{Answer:}\\
C
\end{tcolorbox}

\begin{tcolorbox}[title= QA Example. (\texttt{Relative Reasoning}) (\textsc{TimE-News}) (Level 2: Temporal Expression Reasoning) , label={box: 2-3-news}]
\textbf{Question:}\\
Who publicly reaffirmed their endorsement of Netanyahu on February 21, 2015, within the context of his upcoming speech to Congress?\\A.  Susan Rice, Samantha Power.  \\B.  Hillary Clinton, Bernie Sanders.\\C.  Barack Obama, Joe Biden.  \\D.  Jeb Bush, Ted Cruz\\
\textbf{Answer:}\\
D
\end{tcolorbox}

\begin{tcolorbox}[title= QA Example. (\texttt{Co-temporality}) (\textsc{TimE-News}) (Level 3: Complex Temporal Relationship Reasoning) , label={box: 3-1-news}]
\textbf{Question:}\\
At the same time as Israeli Prime Minister Benjamin Netanyahu’s planned address to a joint meeting of Congress, what did liberal Democrats formally request?\\A.  the endorsement of Netanyahu’s address\\B.  the rescheduling of Netanyahu’s address  \\C.  the cancellation of Netanyahu’s address  \\D.  the delay of Netanyahu’s address\\
\textbf{Answer:}\\
D
\end{tcolorbox}

\begin{tcolorbox}[title= QA Example. (\texttt{Timeline}) (\textsc{TimE-News}) (Level 3: Complex Temporal Relationship Reasoning) , label={box: 3-2-news}]
\textbf{Question:}\\
Below are 3 facts. You need to sort these facts in chronological order. Requirements: You must output a sequence of uppercase letters separated by commas, such as 'A,B,C', without any other characters.\\ \\ \\A. Democratic leaders expressed concern about the announcement of the Israeli Prime Minister's speech to the joint meeting of the House of Representatives and the Senate without consulting neither the White House nor the Democrats.\\B. A number of Democrats have expressed opposition to Israeli Prime Minister Benjamin Netanyahu's planned address to a joint meeting of Congress in March.\\C. Some Democrats criticize Benjamin Netanyahu, who is \"mistaken to agree to it\" and \"playing politics with the critical issue of Israel's security.\\
\textbf{Answer:}\\
B,A,C

\end{tcolorbox}

\begin{tcolorbox}[title= QA Example. (\texttt{Counterfactual}) (\textsc{TimE-News}) (Level 3: Complex Temporal Relationship Reasoning) , label={box: 3-3-News}]
\textbf{Question:}\\
If no liberal Democrats signed the letter requesting the postponement of Netanyahu’s address to Congress from January 2015 to March 2015, who signed the letter?\\A.  The National Iranian American Council drafted the letter.  \\B.  A bipartisan group of lawmakers initiated the letter.\\C.  Conservative Democrats signed the letter.  \\D.  There is no answer\\
\textbf{Answer:}\\
D

\end{tcolorbox}
\subsubsection{\textsc{TimE-Dial}}

% 11个子任务

\begin{tcolorbox}[title= QA Example. (\texttt{Extract}) (\textsc{TimE-Dial}) (Level 1: Basic Temporal Understanding and Retrieval) , label={box: 1-1-dial}]
\textbf{Question:}\\
Which of the following are time expressions mentioned in the context? (Note: There may be one or more correct options. And the time expressions are mentioned directly or indirectly in the context.)\\A. April 17, 2021\\B. 2018\\C. March 16, 2020\\D. March 14, 2019\\
\textbf{Answer:}\\
C

\end{tcolorbox}

\begin{tcolorbox}[title= QA Example. (\texttt{Localization}) (\textsc{TimE-Dial}) (Level 1: Basic Temporal Understanding and Retrieval) , label={box: 1-2-dial}]
\textbf{Question:}\\
When is Debra Ryan working on starting her own business?\\
\textbf{Answer:}\\
8:35 pm, February 21, 2020
\end{tcolorbox}

\begin{tcolorbox}[title= QA Example. (\texttt{Computation}) (\textsc{TimE-Dial}) (Level 1: Basic Temporal Understanding and Retrieval) , label={box: 1-3-dial}]
\textbf{Question:}\\
How long was it between Debra Ryan going skydiving and India Brown attending a street art fest in Brazil? (Please answer using natural time expressions that combine appropriate units based on duration length, e.g. \"2 months 4 days\" for 64 days or \"64 days\" for shorter spans)\\
\textbf{Answer:}\\
19 days

\end{tcolorbox}

\begin{tcolorbox}[title= QA Example. (\texttt{Duration Compare}) (\textsc{TimE-Dial}) (Level 1: Basic Temporal Understanding and Retrieval) , label={box: 1-4-dial}]
\textbf{Question:}\\
Which of the following two durations is longer? *Duration 1:* Between Debra Ryan is learning to play the guitar. and Debra Ryan visited Adventure Land during the weekend trip. *Duration 2:* Between Debra Ryan rode a roller coaster called \"The Wild Ride\" at Adventure Land. and India Brown found flowers by a lake in the park.\\A. Duration 1 is longer.\\B. Duration 2 is longer.\\C. The two durations are approximately the same length.\\

\textbf{Answer:}\\
A

\end{tcolorbox}

\begin{tcolorbox}[title= QA Example. (\texttt{Order Compare}) (\textsc{TimE-Dial}) (Level 1: Basic Temporal Understanding and Retrieval) , label={box: 1-5-dial}]
\textbf{Question:}\\
For Fact1: India Brown became a Queen fan. and Fact2: India Brown found flowers by a lake in the park., which one happened earlier?  \\A. Fact 1 happened earlier.  \\B. Fact 2 happened earlier.  \\C. They happen at almost the same time.\\

\textbf{Answer:}\\
A

\end{tcolorbox}

\begin{tcolorbox}[title= QA Example. (\texttt{Explicit Reasoning}) (\textsc{TimE-Dial}) (Level 2: Temporal Expression Reasoning) , label={box: 2-1-dial}]
\textbf{Question:}\\
What notable artistic or outdoor activities did India Brown participate in between April 1, 2020, and April 9, 2020?\\A. India Brown attended a street art fest in Brazil.\\B. India Brown took a photo of a feather and shells on a beach.\\C. India Brown went hiking and sketching at a nearby national park.\\D. India Brown received positive feedback on her artwork\\
\textbf{Answer:}\\
B
\end{tcolorbox}

\begin{tcolorbox}[title= QA Example. (\texttt{Order Reasoning}) (\textsc{TimE-Dial}) (Level 2: Temporal Expression Reasoning) , label={box: 2-2-dial}]
\textbf{Question:}\\
What was India Brown's third teaching engagement in 2020?\\A.Running a painting workshop for kids.\\B.Teaching art at an orphanage in Cambodia.  \\C.Conducting a live demonstration for her college art club.  \\D.Instructing a pottery class at a local studi\\
\textbf{Answer:}\\
A

\end{tcolorbox}

\begin{tcolorbox}[title= QA Example. (\texttt{Relative Reasoning}) (\textsc{TimE-Dial}) (Level 2: Temporal Expression Reasoning) , label={box: 2-3-dial}]
\textbf{Question:}\\
What was India Brown's most recent job before 12:00 am, March 09, 2020?  \\A.India Brown is working on a new series of abstract artworks based on her trip.\\B.India Brown is working as a travel guide based on her trip experiences.\\C.India Brown is working on a new painting technique learned at a street art festival.\\D.India Brown is testing watercolors for her new series of abstract artworks\\
\textbf{Answer:}\\
A
\end{tcolorbox}

\begin{tcolorbox}[title= QA Example. (\texttt{Co-temporality}) (\textsc{TimE-Dial}) (Level 3: Complex Temporal Relationship Reasoning) , label={box: 3-1-dial}]
\textbf{Question:}\\
At the same time as Debra Ryan is learning to play the guitar, what collection does India Brown have?\\A.India Brown has a collection of soap sculptures.  \\B.India Brown has a collection of watercolor paintings.  \\C.India Brown has a collection of CDs.\\D.India Brown has a collection of vinyl records\\
\textbf{Answer:}\\
C

\end{tcolorbox}

\begin{tcolorbox}[title= QA Example. (\texttt{Timeline}) (\textsc{TimE-Dial}) (Level 3: Complex Temporal Relationship Reasoning) , label={box: 3-2-dial}]
\textbf{Question:}\\
Below are 8 facts. You need to sort these facts in chronological order. Requirements: Your output format must be numbers enclosed in parentheses without any other symbols or whitespace. For example: (1)(5)(2)(7)(3)(8)(6)(4)\\ \\(1) India Brown is working on a new painting technique learned at a street art festival.\\(2) India Brown shared an image of a mural made by kids.\\(3) India Brown had her first art show at a local gallery.\\(4) India Brown became a Queen fan.\\(5) India Brown got invited to exhibit at a local gallery.\\(6) India Brown took a photo of a feather and shells on a beach.\\(7) India Brown sketched a waterfall during a hike.\\(8) India Brown received positive feedback on her artwork.\\

\textbf{Answer:}\\
(4)(5)(1)(7)(6)(2)(8)(3)

\end{tcolorbox}

\begin{tcolorbox}[title= QA Example. (\texttt{Counterfactual}) (\textsc{TimE-Dial}) (Level 3: Complex Temporal Relationship Reasoning) , label={box: 3-3-dial}]
\textbf{Question:}\\
What notable artistic or outdoor activities did India Brown participate in between April 1, 2020, and April 9, 2020, if she visited the Louvre in Paris in March 2020?\\A.India Brown carved a mini sculpture from a soap bar.  \\B.India Brown took a photo of a feather and shells on a beach.\\C.India Brown took a photograph in Santorini, Greece.\\D.India Brown sketched a waterfall during a hik\\

\textbf{Answer:}\\
B

\end{tcolorbox}

% -----------------------补充材料3：QA示例--------------------

% % Appendix C
% 模型角度
% （1）用了哪些模型？分别上下文窗口如何？发布时间？开源或闭源？

% 评估角度
% 哪些任务为什么/用了RAG？哪些任务使用了单纯的zero-shot评估？

% 在什么显卡上面跑的？

% 评估所用的prompt

\section{Experiment Details\label{sec: Appendix_experiment details}}

\subsection{Models\label{sec: appendix_eval_models}}
\subsubsection{Vanilla Models}
We primarily evaluate vanilla models, including the base and instruction-tuned versions of Qwen2.5 series \cite{DBLP:journals/corr/abs-2412-15115qwen25,DBLP:journals/corr/abs-2501-15383qwen251m}, the instruction-tuned Llama-3.1 model \cite{DBLP:journals/corr/abs-2407-21783llama3}, as well as state-of-the-art models such as Deepseek-V3 \cite{DBLP:journals/corr/abs-2412-19437deepseek-v3} and GPT-4o \cite{DBLP:journals/corr/abs-2410-21276gpt4o}. These models are evaluated without any test-time computation scaling specifically designed to enhance their reasoning capabilities.

\subsubsection{Test-time Scaled Models}
We primarily select Deepseek-R1\cite{DBLP:journals/corr/abs-2501-12948deepseek-r1}, OpenAI o3-mini, QwQ-32B\cite{qwq32b}, and Deepseek-R1 Distilled Models\cite{DBLP:journals/corr/abs-2501-12948deepseek-r1} as our test-time scaled models. These models enhance their logical reasoning capabilities through reinforcement learning or direct distillation from advanced test-time scaled models. They demonstrate strong reasoning performance not only in mathematical and coding domains but also exhibit generalizable reasoning abilities across diverse fields.
\paragraph{Deepseek-R1 Distilled Models}
We conduct experiments using distilled models from Deepseek-R1 \cite{DBLP:journals/corr/abs-2501-12948deepseek-r1}, specifically Deepseek-R1-Distilled-Qwen-7B, Deepseek-R1-Distilled-Qwen-14B, Deepseek-R1-Distilled-Qwen-32B, Deepseek-R1-Distill-Llama-8B, and Deepseek-R1-Distill-Llama-70B. These models are derived through knowledge distillation from Deepseek-R1, with their base architectures being Qwen2.5-Math-7B \cite{DBLP:journals/corr/abs-2409-12122qwen25math}, Qwen2.5-14B, Qwen2.5-32B \cite{DBLP:journals/corr/abs-2412-15115qwen25}, Llama-3.1-8B, and Llama-3.3-70B-Instruct \cite{DBLP:journals/corr/abs-2407-21783llama3}, respectively.

\subsection{Evaluation Metrics\label{sec: appendix_eval_metrics}}
% 不同任务形式具有不同的评估指标。不同任务的QA formats如Table \ref{tab: Overview of Question Answering formats}所示。
% 我们参考\cite{chu2024timebench}的评估方式：（1）对于free-form QA的评估，我们使用token-level的Exact Match (EM)和F1分数。具体而言，我们对\texttt{Timeline}任务使用token-level EM指标；而对于其他free-form QAs均使用token-level的F1分数。
% （2）对于single-choice和multiple-choice形式的问题，我们使用了option-level F1指标。更具体地，我们使用了macro option-level F1指标。

We employ distinct evaluation metrics tailored to different task formats, as detailed in Table \ref{tab: Overview of Question Answering formats}. Following the evaluation protocol established in \cite{chu2024timebench}, we adopt the following metrics: (1) For free-form QA tasks, we utilize token-level Exact Match (EM) and F1 scores. Specifically, we apply token-level EM for the \texttt{Timeline} task, while employing token-level F1 for other free-form QA tasks. (2) For single-choice and multiple-choice questions, we implement option-level F1 scores, with a particular focus on macro option-level F1 to ensure comprehensive evaluation across all options.

\paragraph{Token-level Exact Match and F1 Score}
The token-level Exact Match is a binary metric that evaluates the complete match between predicted and ground truth answers:
$$\text{Exact Match} = 
\begin{cases} 
1.0, & \text{if } \text{ pred\_answer.lower().strip() = gold\_answer.lower().strip()} \\
0.0, & \text{otherwise}
\end{cases}$$
This strict metric assigns a score of 1 only when the predicted answer exactly matches the ground truth (ignoring case and leading/trailing whitespace), and 0 otherwise.

The token-level F1 Score measures answer similarity through lexical overlap between predicted and ground truth answers, computed as follows:

1. Tokenization and normalization: Convert answers to lowercase, remove punctuation, and tokenize by whitespace.

2. Calculate shared tokens:
   $$c = \sum_{t \in \text{tokens}} \min(\text{freq}_{\text{gold}}(t), \text{freq}_{\text{pred}}(t))$$
   where $t$ represents unique tokens, and $\text{freq}_{\text{gold}}(t)$ and $\text{freq}_{\text{pred}}(t)$ denote the frequency of token $t$ in ground truth and predicted answers respectively.

3. Compute precision:
   $$\text{precision} = \frac{c}{|\text{pred\_tokens}|}$$

4. Compute recall:
   $$\text{recall} = \frac{c}{|\text{gold\_tokens}|}$$

5. Calculate F1 score:
   $$\text{F1} = \frac{2 \times \text{precision} \times \text{recall}}{\text{precision} + \text{recall}}$$
   with F1 = 0 when precision + recall = 0.

This lexical overlap-based F1 score captures partially correct answers, making it more lenient than Exact Match.

\paragraph{Option-level F1 Score}
For multiple-choice questions, the Option-level F1 score evaluates the match between predicted and ground truth options:

1. Extract options: Normalize options (e.g., "A B C" or "A,B,C") into standardized option sets.

2. Compute confusion matrix:
   $$\text{TP} = |\text{pred\_options} \cap \text{gold\_options}|$$
   $$\text{FP} = |\text{pred\_options} - \text{gold\_options}|$$
   $$\text{FN} = |\text{gold\_options} - \text{pred\_options}|$$

3. Calculate precision and recall:
   $$\text{precision} = \frac{\text{TP}}{|\text{pred\_options}|}$$
   $$\text{recall} = \frac{\text{TP}}{|\text{gold\_options}|}$$

4. Compute Option-level F1 score:
   $$\text{pair\_level\_f1} = 
   \begin{cases} 
   \frac{2 \times \text{precision} \times \text{recall}}{\text{precision} + \text{recall}}, & \text{if } \text{ pred\_options} \subseteq \text{gold\_options} \\
   0, & \text{if } \exists x \in \text{pred\_options}: x \notin \text{gold\_options}
   \end{cases}$$

\paragraph{Macro and Micro F1 Scores}
The Macro F1 Score averages F1 scores across all questions:
$$\text{Macro F1} = \frac{1}{n} \sum_{i=1}^{n} \text{F1}_i$$
where $n$ is the total number of questions and $\text{F1}_i$ is the F1 score for the $i$-th question. This approach assigns equal weight to each question, making it robust to imbalanced option distributions across the dataset.

In contrast, Micro F1 aggregates all TP, FP, and FN across questions before computing the overall score:
$$\text{micro\_precision} = \frac{\sum_{i=1}^{n} \text{TP}_i}{\sum_{i=1}^{n} (\text{TP}_i + \text{FP}_i)}$$
$$\text{micro\_recall} = \frac{\sum_{i=1}^{n} \text{TP}_i}{\sum_{i=1}^{n} (\text{TP}_i + \text{FN}_i)}$$
$$\text{micro\_f1} = \frac{2 \times \text{micro\_precision} \times \text{micro\_recall}}{\text{micro\_precision} + \text{micro\_recall}}$$

\subsection{Retriever for Evaluating \textsc{TimE-News}\label{sec: appendix_retriever}}
We employ three distinct retrievers, each retrieving the top-3 text chunks based on the given question, with a maximum of 500 words per chunk.
\paragraph{BM25}
BM25 is a bag-of-words retrieval model that computes relevance scores based on term frequency (TF), inverse document frequency (IDF), and document length, without considering word order. It improves upon traditional TF calculation by preventing unbounded growth and introduces two key parameters: a document length normalization parameter (typically b) and a TF saturation parameter (typically k1) for finer score adjustment. The primary strength of BM25 lies in its effective handling of keyword matching and its ability to assign appropriate weights to both common and rare terms.

\paragraph{Vector}%BGE-3
The core of vector retrieval lies in high-quality text embedding models that capture deep semantic information. Unlike keyword-based methods such as BM25, vector retrieval excels at handling synonyms, near-synonyms, and complex semantic relationships, enabling it to retrieve documents that are semantically relevant to the query even when they do not contain exact keyword matches. We employ the state-of-the-art BGE-M3 \cite{bge-m3} text embedding model as our vector retriever.

\paragraph{Hybrid}%BM25 + BGE-3 + Reranker
The hybrid retrieval approach aims to combine the strengths of keyword-based retrieval (e.g., BM25) and semantic vector retrieval to achieve superior performance compared to individual methods, typically enhancing both recall and accuracy. Specifically, we first conduct initial retrieval by invoking both BM25 and vector retrievers to obtain the top-5 candidate results for each query, followed by result merging and deduplication. Our merging strategy prioritizes BM25 results while supplementing with unique results from vector retrieval. Subsequently, we perform document re-ranking using BGE-Reranker-Base \cite{bge_reranker}. Finally, we select the top-3 documents from the (re-)ranked candidates. Notably, before generating chunks, we sort documents in ascending order by date. This temporal sorting may override previous relevance-based rankings (whether from BM25, vector similarity, or reranker scores), with its impact contingent on whether the application prioritizes timeliness.

\paragraph{Computation Resource\label{sec: computation resource}}
All experiments are done on 4 NVIDIA A800 GPUs with 80GB memory for each GPU.

% -----------------------------补充材料3-------------------------------
% 评估所用的prompt
\subsection{Prompt Templates for Evaluation}

% 下面Prompt中，\textbf{\textit{\{context\}}}对于RAG设定而言是检索到的top-k个chunks，对于非RAG设定而言，是问题对应的context。

\begin{tcolorbox}[title=Evaluation Prompt Template for \textbf{free-form} tasks (exluding \texttt{Counterfactual} and \texttt{Computation}) , label={box: eval_prompt_ordinary}]
\textbf{Context:} \{context\}

You need to answer the following question based on the given context.
If you can infer the answer from the context, please output your answer directly, keeping it concise and accurate, without any explanatory text. **If you are certain there are multiple answers in the context that satisfy the question, please output all answers, one per line (i.e., separate each answer with a line break).** And you will never refuse to answer any question.\\

\textbf{Question:} \{question\}\\
Therefore, the answer is 

\end{tcolorbox}

\begin{tcolorbox}[title=Evaluation Prompt Template for \textbf{multi-choice} tasks, label={box: eval_prompt_ordinary_multichoice}]
\textbf{Context:} \{context\}\\

\textbf{Instruction:} You’re an expert in answering multiple-choice questions. You should choose the options that you think is most likely to be correct in the following question. And you will never refuse to answer any question.\\

\textbf{Rules:}\\
1. You need to answer the following multiple-choice question based on the given context.\\
2. You should output the answer in the format of "[[X Y ...]]", WITHOUT anything else, where ’X’, ’Y’, etc. are the uppercase letters of the correct options. Do not include any other explanatory text in your answer.\\
3. Example Outputs: (NOTE: The following are only examples, which are NOT relevant to the question and your answer. Your output should be formatted exactly like this.)\\
Answer: [[A C]]\\
Answer: [[B]]\\
Answer: [[B D]]\\

\textbf{Question:} \{question\}\\
Therefore, the answer is \\

\end{tcolorbox}

\begin{tcolorbox}[title=Evaluation Prompt Template for \textbf{single-choice} tasks (excluding \texttt{Counterfactual}), label={box: eval_prompt_ordinary_singlechoice}]
\textbf{Context:} \{context\}\\

\textbf{Instruction:} You’re an expert in answering single-choice questions. You should choose the option that you think is most likely to be correct in the following question. And you will never refuse to answer any question.\\

\textbf{Rules:}\\
1. You need to answer the following single-choice question based on the given context.\\
2. You should output the answer in the format of "[[X]]", WITHOUT anything else, where ’X’ is the choice’s uppercase letter. Do not include any other explanatory text in your answer.\\
3. Example Outputs: (NOTE: The following are only examples, which are NOT relevant to the question. Your output should be formatted exactly like this.)\\
Answer: [[A]]\\
Answer: [[B]]\\
Answer: [[C]]\\
Answer: [[D]]\\

\textbf{Question:} \{question\}\\
Therefore, the answer is \\

\end{tcolorbox}

\begin{tcolorbox}[title=Evaluation Prompt Template for \textbf{free-form} \texttt{Counterfactual} , label={box: eval_prompt_counterfactual}]
\textbf{Context:} \{context\}

You need to answer the following question based on the given context.
If you can infer the answer from the context, please output your answer directly, keeping it concise and accurate, without any explanatory text. **If you are certain there are multiple answers in the context that satisfy the question, please output all answers, one per line (i.e., separate each answer with a line break).** Otherwise, if there is no answer, simply output "There is no answer."\\

\textbf{Hint}: The following question is a free-form question. This question is based on a premise that contradicts the temporal information in the original text. You need to fully understand the temporal information in the original text and, while satisfying the false premise in the question, answer the question as accurately as possible. You should not include any explanatory text in your answer, just output the answer directly.\\

\textbf{Question:} \{question\}\\
Therefore, the answer is 

\end{tcolorbox}

\begin{tcolorbox}[title=Evaluation Prompt Template for \textbf{single-choice} \texttt{Counterfactual} , label={box: eval_prompt_counterfactual_single_choice}]
\textbf{Context:} \{context\}

\textbf{Instruction: }You’re an expert in answering single-choice questions. You should choose the option that you think is most likely to be correct in the following question. And you will never refuse to answer any question.\\

\textbf{Rules:}\\
1. You need to answer the following single-choice question based on the given context.\\
2. You should output the answer in the format of "[[X]]", WITHOUT anything else, where ’X’ is the choice’s uppercase letter. Do not include any other explanatory text in your answer.\\
3. Example Outputs: (NOTE: The following are only examples, which are NOT relevant to the question. Your output should be formatted exactly like this.)\\
Answer: [[A]]\\
Answer: [[B]]\\
Answer: [[C]]\\
Answer: [[D]]\\

\textbf{Hint:} The following question is a single-choice question. This question is based on a premise that contradicts the temporal information in the original text. The correct option is the one that satisfies the premise (although it contradicts the temporal information in the original text) and satisfies the temporal information in the context. Choose only one option that best aligns with the temporal information in the original text.\\

\textbf{Question:} \{question\}\\
Therefore, the answer is 

\end{tcolorbox}

\begin{tcolorbox}[title=Evaluation Prompt Template for \texttt{Computation} , label={box: eval_prompt_computation}]
\textbf{Context:} \{context\}

You need to answer the following question based on the given context.
If you can infer the answer from the context, please output your answer directly, keeping it concise and accurate, without any explanatory text. **If you are certain there are multiple answers in the context that satisfy the question, please output all answers, one per line (i.e., separate each answer with a line break).** And you will never refuse to answer any question.\\

\textbf{Hint}: The following question is a free-form question. Your answer needs to follow the correct format. Here are some examples of answer formats:
1. March 9, 2020    (format: year-month-day)
2. 1:44 pm, April 16, 2020    (format: hour:minute, year-month-day)
3. 1972    (format: year)
4. April, 2020    (format: month, year)
The above examples are only for reference regarding the format of your answer, and are not related to the actual content of your answer. You only need to follow one of the above formats. In terms of content, you need to make your answer as consistent as possible with the original context.\\

\textbf{Question:} \{question\}\\
Therefore, the answer is 

\end{tcolorbox}

% -----------------------------补充材料3-------------------------------

\subsection{Complete Experimental Results\label{sec: complete_experimental_results}}

% 6个图
% ------------------TIME-Wiki---------------------
\begin{table*}[htbp]
    \centering
    \caption{
    Experimental Results for \textsc{TimE-Wiki}
    }
    \label{tab: exp on TIME-Wiki for appendix}
    \resizebox{\textwidth}{!}{% Resize table to fit within text width
\begin{tabular}{@{}lccccccccccc@{}}
\toprule
\multirow{2}{*}{\textbf{Model}} & \multicolumn{5}{c}{\textbf{Level 1}} & \multicolumn{3}{c}{\textbf{Level-2}} & \multicolumn{3}{c}{\textbf{Level-3}} \\
\cmidrule(lr){2-6} \cmidrule(lr){7-9} \cmidrule(lr){10-12}
& \textbf{Extract} & \textbf{Location} & \textbf{Comp.} & \textbf{Dur. Comp.} & \textbf{Ord. Comp.} & \textbf{Expl. Reason.} & \textbf{Ord. Reason.} & \textbf{Rel. Reason.} & \textbf{Co-temp.} & \textbf{Timeline} & \textbf{Counterf.} \\
\midrule
\multicolumn{12}{c}{\textit{Vanilla Models}}\\
\midrule
Llama-3.1-70B & 46.57 & 62.10 & 31.93 & 36.87 & 49.50 & 12.08 & 16.34 & 15.33 & 16.76 & 0.08 & 18.16 \\
Llama-3.1-8B-Instruct & 53.16 & 75.41 & 9.79 & 50.89 & 65.49 & 28.96 & 31.72 & 24.53 & 31.36 & 0.92 & 28.60 \\
Llama-3.1-70B-Instruct & 83.77 & 69.58 & 56.58 & 71.20 & 84.94 & 46.55 & 38.05 & 31.06 & 37.10 & 5.69 & 42.68 \\
Qwen2.5-3B & 4.69 & 58.68 & 7.92 & 6.55 & 33.56 & 5.10 & 9.22 & 10.94 & 9.36 & 0.00 & 10.35 \\
Qwen2.5-7B & 35.33 & 67.19 & 24.22 & 23.73 & 65.36 & 10.11 & 14.66 & 5.45 & 2.45 & 0.00 & 0.98 \\
Qwen2.5-14B & 33.58 & 71.26 & 20.53 & 50.64 & 66.49 & 7.42 & 15.37 & 20.68 & 17.11 & 0.00 & 27.95 \\
Qwen2.5-32B & 39.20 & 73.58 & 30.04 & 64.68 & 77.16 & 19.09 & 16.61 & 16.87 & 12.86 & 0.00 & 28.56 \\
Qwen2.5-72B & 54.27 & 72.01 & 33.27 & 61.42 & 46.57 & 28.00 & 22.40 & 6.70 & 10.21 & 0.00 & 5.32 \\
Qwen2.5-3B-Instruct & 36.26 & 53.35 & 11.86 & 42.72 & 60.07 & 18.94 & 30.61 & 26.38 & 22.73 & 0.23 & 32.10 \\
Qwen2.5-7B-Instruct & 57.58 & 65.30 & 32.34 & 52.22 & 68.75 & 44.76 & 35.48 & 26.79 & 36.68 & 1.08 & 38.42 \\
Qwen2.5-14B-Instruct & 71.02 & 74.49 & 26.37 & 63.50 & 82.76 & 52.93 & 38.94 & 30.34 & 33.68 & 2.62 & 43.16 \\
Qwen2.5-32B-Instruct & 88.91 & 70.68 & 28.02 & 72.26 & 85.19 & 36.89 & 29.53 & 31.25 & 39.77 & 5.23 & 49.56 \\
Qwen2.5-72B-Instruct & 81.70 & 83.84 & 41.37 & 66.64 & 84.22 & 70.13 & 44.84 & 35.23 & 51.17 & 4.08 & 50.68 \\
Qwen2.5-7B-Instruct-1M & 47.43 & 64.56 & 38.34 & 37.19 & 66.51 & 42.21 & 36.53 & 27.23 & 42.57 & 0.69 & 38.75 \\
Qwen2.5-14B-Instruct-1M & 54.00 & 79.82 & 32.61 & 64.06 & 78.62 & 58.69 & 40.86 & 28.58 & 34.56 & 3.46 & 41.79 \\
\midrule
\multicolumn{12}{c}{\textit{Test-time Scaled Models}}\\
\midrule
Deepseek-R1-Distill-Qwen-7B & 54.89 & 65.04 & 56.63 & 77.85 & 85.71 & 48.88 & 32.53 & 30.57 & 29.74 & 0.54 & 37.38 \\
Deepseek-R1-Distill-Qwen-14B & 67.66 & 66.33 & 51.25 & 81.21 & 92.97 & 58.94 & 43.49 & 35.63 & 36.30 & 14.54 & 45.69 \\
Deepseek-R1-Distill-Qwen-32B & 74.98 & 75.61 & 68.80 & 87.85 & 93.58 & 61.68 & 42.86 & 37.44 & 43.41 & 20.23 & 45.89 \\
Deepseek-R1-Distill-Llama-8B & 66.75 & 68.82 & 57.27 & 83.47 & 90.22 & 51.17 & 37.36 & 32.41 & 31.04 & 5.31 & 37.30 \\
Deepseek-R1-Distill-Llama-70B & 74.38 & 70.21 & 73.35 & 88.54 & 93.61 & 65.94 & 45.54 & 38.83 & 43.10 & 21.69 & 45.97 \\
QwQ-32B & 74.99 & 67.75 & 49.59 & 88.20 & 93.53 & 60.61 & 37.77 & 36.39 & 37.76 & 25.38 & 53.13 \\
\bottomrule
\end{tabular}
    }
 \end{table*}

% ------------------TIME-Lite-Wiki---------------------
   \begin{table*}[htbp]
      \centering
      \caption{
      Complete experimental results for \textsc{TimE-Lite-Wiki}
      }
      \label{tab: exp on TIME-Lite-Wiki for appendix}
      \resizebox{\textwidth}{!}{% Resize table to fit within text width
\begin{tabular}{@{}lccccccccccc@{}}
\toprule
\multirow{2}{*}{\textbf{Model}} & \multicolumn{5}{c}{\textbf{Level 1}} & \multicolumn{3}{c}{\textbf{Level-2}} & \multicolumn{3}{c}{\textbf{Level-3}} \\
\cmidrule(lr){2-6} \cmidrule(lr){7-9} \cmidrule(lr){10-12}
& \textbf{Extract} & \textbf{Location} & \textbf{Comp.} & \textbf{Dur. Comp.} & \textbf{Ord. Comp.} & \textbf{Expl. Reason.} & \textbf{Ord. Reason.} & \textbf{Rel. Reason.} & \textbf{Co-temp.} & \textbf{Timeline} & \textbf{Counterf.} \\
\midrule
\multicolumn{12}{c}{\textit{Vanilla Models}}\\
\midrule
Deepseek-V3 & 93.33 & 84.51 & 23.76 & 71.43 & 83.33 & 75.69 & 39.77 & 41.76 & 46.62 & 10.00 & 44.82 \\
GPT-4o & 98.89 & 83.24 & 33.82 & 67.86 & 90.00 & 80.68 & 45.83 & 46.56 & 45.45 & 20.00 & 50.72 \\
Qwen2.5-3B-Instruct & 33.22 & 63.01 & 10.28 & 42.86 & 63.33 & 38.63 & 31.58 & 42.33 & 7.29 & 0.00 & 30.80 \\
Qwen2.5-7B-Instruct & 52.00 & 78.71 & 24.86 & 50.00 & 70.00 & 74.78 & 36.56 & 41.50 & 23.98 & 3.33 & 33.72 \\
Qwen2.5-14B-Instruct & 68.67 & 81.17 & 18.25 & 71.43 & 80.00 & 70.21 & 43.38 & 43.15 & 23.08 & 3.33 & 43.91 \\
Qwen2.5-32B-Instruct & 89.33 & 81.13 & 30.09 & 71.43 & 86.67 & 63.57 & 42.72 & 40.69 & 37.88 & 13.33 & 49.22 \\
Qwen2.5-72B-Instruct & 79.22 & 85.06 & 38.47 & 64.29 & 76.67 & 78.59 & 43.61 & 41.21 & 45.64 & 6.67 & 38.72 \\
Qwen2.5-7B-Instruct-1M & 49.33 & 83.87 & 23.50 & 57.14 & 63.33 & 74.49 & 33.89 & 44.27 & 42.90 & 0.00 & 32.81 \\
\midrule
\multicolumn{12}{c}{\textit{Test-time Scaled Models}}\\
\midrule
Deepseek-R1 & 96.67 & 77.61 & 46.39 & 89.29 & 93.33 & 78.20 & 57.09 & 57.79 & 47.45 & 33.33 & 55.71 \\
o3-mini & 96.67 & 80.83 & 49.17 & 92.86 & 93.33 & 82.24 & 52.62 & 48.98 & 54.34 & 33.33 & 52.07 \\
QwQ-32B & 84.67 & 55.25 & 40.80 & 89.29 & 90.00 & 74.70 & 43.00 & 54.82 & 37.36 & 23.33 & 43.23 \\
Deepseek-R1-Distill-Llama-8B & 85.78 & 66.17 & 22.95 & 82.14 & 76.67 & 48.23 & 28.85 & 37.30 & 39.70 & 10.00 & 35.00 \\
Deepseek-R1-Distill-Qwen-7B & 57.00 & 66.38 & 30.36 & 78.57 & 73.33 & 41.92 & 22.23 & 28.59 & 32.98 & 0.00 & 28.79 \\
Deepseek-R1-Distill-Qwen-14B & 84.44 & 69.03 & 39.44 & 82.14 & 86.67 & 60.87 & 36.22 & 42.04 & 41.63 & 20.00 & 45.13 \\
Deepseek-R1-Distill-Qwen-32B & 92.44 & 72.67 & 23.81 & 89.29 & 90.00 & 59.63 & 38.44 & 40.72 & 37.94 & 13.33 & 41.98 \\
\bottomrule
\end{tabular}
 %      \begin{tabular}{@{}lccccccccccc@{}}
 %      \toprule
 %      \multirow{2}{*}{\textbf{Model}} & \multicolumn{5}{c}{\textbf{Level 1}} & \multicolumn{3}{c}{\textbf{Level-2}} & \multicolumn{3}{c}{\textbf{Level-3}} \\
 %      \cmidrule(lr){2-6} \cmidrule(lr){7-9} \cmidrule(lr){10-12}
 %      & \textbf{Extract} & \textbf{Location} & \textbf{Comp.} & \textbf{Dur. Comp.} & \textbf{Ord. Comp.} & \textbf{Expl. Reason.} & \textbf{Ord. Reason.} & \textbf{Rel. Reason.} & \textbf{Co-temp.} & \textbf{Timeline} & \textbf{Counterf.} \\
 %      \midrule
 %      \multicolumn{12}{c}{\textit{Vanilla Models}}\\
 %      \midrule
 %      Deepseek-V3& & & & & & & & & & & \\
 %        GPT-4o& & & & & & & & & & & \\
 % Qwen2.5-3B-Instruct& & & & & & & & & & &\\
 % Qwen2.5-7B-Instruct& & & & & & & & & & &\\
 % Qwen2.5-14B-Instruct& & & & & & & & & & &\\
 %      Qwen2.5-32B-Instruct& & & & & & & & & & & \\
 %            Qwen2.5-72B-Instruct& & & & & & & & & & & \\
 %            Qwen2.5-7B-Instruct-1M& & & & & & & & & & & \\
 %      \midrule
 %      \multicolumn{12}{c}{\textit{Test-time Scaled Models}}\\
 %      \midrule
 %        Deepseek-R1& & & & & & & & & & & \\
 %        o3-mini& & & & & & & & & & & \\
 % QwQ-32B& & & & & & & & & & &\\
 % Deepseek-R1-Distill-Llama-8B& & & & & & & & & & &\\
 % Deepseek-R1-Distill-Qwen-7B& & & & & & & & & & &\\
 % Deepseek-R1-Distill-Qwen-14B& & & & & & & & & & &\\
 % Deepseek-R1-Distill-Qwen-32B& & & & & & & & & & &\\
 %      \bottomrule
 %      \end{tabular}%
      }
   \end{table*}

% ----------TIME-News---------------
\begin{table*}[htbp]
    \centering
    \caption{
    Complete experimental results for \textsc{TimE-News}
    }
    \label{tab: exp on TIME-News for appendix}
    \resizebox{\textwidth}{!}{% Resize table to fit within text width
\begin{tabular}{@{}llcccccccccc@{}} % Original was 13 columns (ll + 11c), now 12 (ll + 10c)
    \toprule
    \multirow{2}{*}{\textbf{Model}} & \multirow{2}{*}{\textbf{Retriever}}& \multicolumn{4}{c}{\textbf{Level 1}} & \multicolumn{3}{c}{\textbf{Level-2}} & \multicolumn{3}{c}{\textbf{Level-3}} \\
    \cmidrule(lr){3-6} \cmidrule(lr){7-9} \cmidrule(lr){10-12}
    & & \textbf{Location} & \textbf{Comp.} & \textbf{Dur. Comp.} & \textbf{Ord. Comp.} & \textbf{Expl. Reason.} & \textbf{Ord. Reason.} & \textbf{Rel. Reason.} & \textbf{Co-temp.} & \textbf{Timeline} & \textbf{Counterf.} \\
    \midrule
    \multicolumn{12}{c}{\textit{Vanilla Models} (\textsc{TimE-News})} \\
    \midrule
    \multirow{3}{*}{Llama3.1-8B-Instruct}&BM25& 47.96 & 27.12 & 39.06 & 39.28 & 81.72 & 66.67 & 77.06 & 80.50 & 3.09 & 47.17 \\
    & Vector & 50.99 & 32.13 & 40.94 & 41.17 & 81.33 & 67.67 & 77.67 & 81.50 & 1.94 & 46.22 \\
    & Hybrid & 51.81 & 34.51 & 41.78 & 44.11 & 82.50 & 68.94 & 78.39 & 82.89 & 2.55 & 46.44 \\
    \multirow{3}{*}{Qwen2.5-14B-Instruct}&BM25& 68.53 & 70.80 & 42.39 & 46.17 & 83.06 & 70.44 & 79.61 & 82.67 & 26.13 & 59.39 \\
    &Vector& 71.68 & 76.28 & 42.22 & 45.67 & 83.94 & 69.33 & 80.33 & 83.44 & 23.68 & 59.67 \\
    &Hybrid& 71.00 & 79.75 & 43.61 & 48.72 & 84.72 & 70.39 & 81.44 & 84.06 & 26.61 & 58.61 \\
    \multirow{3}{*}{Qwen2.5-32B-Instruct}&BM25& 68.88 & 79.48 & 46.44 & 51.22 & 84.39 & 70.78 & 81.56 & 85.11 & 27.54 & 54.61 \\
    &Vector& 71.76 & 84.46 & 44.78 & 50.61 & 85.22 & 70.94 & 82.11 & 84.39 & 24.16 & 55.83 \\
    &Hybrid& 71.57 & 86.62 & 44.78 & 54.83 & 86.28 & 71.17 & 82.72 & 86.11 & 25.92 & 54.06 \\
    \multirow{3}{*}{Qwen2.5-7B-Instruct-1M} &BM25& 69.26 & 73.86 & 41.39 & 51.06 & 82.67 & 68.94 & 79.50 & 83.28 & 22.46 & 52.83 \\
    & Vector & 71.01 & 72.27 & 41.67 & 51.28 & 84.39 & 69.33 & 80.00 & 83.44 & 22.27 & 53.22 \\
    & Hybrid & 70.82 & 72.84 & 41.83 & 52.89 & 83.39 & 70.17 & 80.33 & 84.50 & 22.91 & 53.22 \\
    \multirow{3}{*}{Qwen2.5-14B-Instruct-1M} &BM25& 68.57 & 82.97 & 42.83 & 55.89 & 84.00 & 71.33 & 80.83 & 83.44 & 25.94 & 56.00 \\
    & Vector & 71.77 & 85.31 & 43.28 & 54.89 & 85.00 & 70.44 & 81.11 & 84.33 & 23.71 & 57.56 \\
    & Hybrid & 71.72 & 85.73 & 44.56 & 57.89 & 85.67 & 71.50 & 82.56 & 84.56 & 26.29 & 56.00 \\
    \midrule
    \multicolumn{12}{c}{\textit{Test-time Scaled Models} (\textsc{TimE-News})} \\
    \midrule
    \multirow{3}{*}{Deepseek-R1-Distill-Qwen-7B} &BM25& 39.66 & 60.15 & 38.78 & 53.33 & 76.28 & 60.06 & 70.17 & 74.56 & 17.94 & 37.11 \\
    &Vector& 41.17 & 59.81 & 41.72 & 54.56 & 76.44 & 61.94 & 73.89 & 74.67 & 16.44 & 38.78 \\
    &Hybrid& 41.42 & 60.28 & 38.22 & 54.78 & 78.22 & 62.67 & 72.72 & 76.39 & 17.08 & 39.06 \\
    \multirow{3}{*}{Deepseek-R1-Distill-Qwen-14B}&BM25& 63.42 & 62.36 & 39.72 & 52.61 & 83.39 & 70.33 & 80.83 & 83.78 & 21.82 & 62.72 \\
    &Vector& 65.96 & 63.56 & 39.39 & 51.33 & 84.89 & 69.22 & 81.28 & 83.89 & 19.58 & 63.44 \\
    &Hybrid& 66.11 & 66.29 & 39.39 & 54.94 & 85.61 & 69.89 & 82.67 & 85.00 & 21.10 & 62.00 \\
    \multirow{3}{*}{Deepseek-R1-Distill-Llama-8B}&BM25& 53.75 & 65.39 & 44.00 & 53.72 & 80.44 & 65.56 & 76.06 & 80.28 & 20.33 & 51.89 \\
    &Vector& 56.10 & 65.61 & 43.56 & 52.44 & 81.11 & 65.61 & 77.83 & 79.39 & 18.73 & 54.39 \\
    &Hybrid& 55.66 & 67.92 & 41.22 & 54.78 & 82.44 & 66.39 & 78.89 & 81.06 & 20.06 & 52.39 \\
    \bottomrule
\end{tabular}
    }
\end{table*}

% ----------TIME-Lite-News---------------
\begin{table*}[htbp]
    \centering
    \caption{
    Complete experimental results for \textsc{TimE-Lite-News}
    }
    \label{tab: exp on TIME-Lite-News for appendix}
    \resizebox{\textwidth}{!}{% Resize table to fit within text width
\begin{tabular}{@{}llcccccccccc@{}} % Original was 13 columns (ll + 11c), now 12 (ll + 10c)
    \toprule
    \multirow{2}{*}{\textbf{Model}} & \multirow{2}{*}{\textbf{Retriever}}& \multicolumn{4}{c}{\textbf{Level 1}} & \multicolumn{3}{c}{\textbf{Level-2}} & \multicolumn{3}{c}{\textbf{Level-3}} \\
    \cmidrule(lr){3-6} \cmidrule(lr){7-9} \cmidrule(lr){10-12}
    & & \textbf{Location} & \textbf{Comp.} & \textbf{Dur. Comp.} & \textbf{Ord. Comp.} & \textbf{Expl. Reason.} & \textbf{Ord. Reason.} & \textbf{Rel. Reason.} & \textbf{Co-temp.} & \textbf{Timeline} & \textbf{Counterf.} \\
    \midrule
    \multicolumn{12}{c}{\textit{Vanilla Models} (\textsc{TimE-News})} \\ 
    \midrule
    \multirow{3}{*}{Llama3.1-8B-Instruct}&BM25& 50.99 & 1.46 & 46.67 & 33.33 & 73.33 & 60.00 & 83.33 & 83.33 & 0.00 & 30.00 \\
    & Vector & 48.87 & 5.77 & 50.00 & 43.33 & 80.00 & 56.67 & 86.67 & 83.33 & 0.00 & 30.00 \\
    & Hybrid & 54.06 & 4.54 & 46.67 & 46.67 & 80.00 & 63.33 & 80.00 & 80.00 & 0.00 & 26.67 \\
    
    \multirow{3}{*}{Qwen2.5-14B-Instruct}&BM25& 71.56 & 3.96 & 53.33 & 46.67 & 63.33 & 63.33 & 80.00 & 93.33 & 17.24 & 33.33 \\
    &Vector & 76.27 & 9.58 & 50.00 & 43.33 & 86.67 & 63.33 & 86.67 & 80.00 & 17.24 & 36.67 \\
    &Hybrid & 77.84 & 8.36 & 66.67 & 36.67 & 76.67 & 60.00 & 80.00 & 86.67 & 20.69 & 40.00 \\
    
    \multirow{3}{*}{Qwen2.5-14B-Instruct-1M} &BM25& 71.11 & 9.44 & 40.00 & 36.67 & 73.33 & 66.67 & 83.33 & 93.33 & 13.79 & 33.33 \\
    & Vector & 73.33 & 13.33 & 46.67 & 43.33 & 76.67 & 60.00 & 93.33 & 86.67 & 24.14 & 33.33 \\
    & Hybrid & 76.67 & 11.11 & 43.33 & 46.67 & 76.67 & 63.33 & 86.67 & 93.33 & 17.24 & 33.33 \\
    
    \midrule
    \multicolumn{12}{c}{\textit{Test-time Scaled Models} (\textsc{TimE-News})} \\
    \midrule
    
    \multirow{3}{*}{Deepseek-R1-Distill-Qwen-7B} &BM25& 40.37 & 10.00 & 63.33 & 43.33 & 76.67 & 70.00 & 76.67 & 80.00 & 10.34 & 26.67 \\
    &Vector & 40.74 & 8.13 & 46.67 & 60.00 & 70.00 & 50.00 & 83.33 & 73.33 & 3.45 & 30.00 \\
    &Hybrid & 40.00 & 3.33 & 50.00 & 60.00 & 76.67 & 60.00 & 76.67 & 66.67 & 10.34 & 26.67 \\
    
    \multirow{3}{*}{Deepseek-R1-Distill-Qwen-14B} &BM25& 66.33 & 7.67 & 56.67 & 46.67 & 80.00 & 70.00 & 83.33 & 86.67 & 6.90 & 43.33 \\
    &Vector & 69.81 & 7.62 & 50.00 & 43.33 & 80.00 & 56.67 & 90.00 & 90.00 & 13.79 & 43.33 \\
    &Hybrid & 65.56 & 9.77 & 43.33 & 63.33 & 76.67 & 63.33 & 83.33 & 90.00 & 10.34 & 40.00 \\
    
    \bottomrule
\end{tabular}
    }
\end{table*}

% --------------TIME-Dial------------------

   \begin{table*}[htbp]
      \centering
      \caption{
      Complete experimental results for \textsc{TimE-Dial}
      }
      \label{tab: exp on TIME-Dial for Appendix}
      \resizebox{\textwidth}{!}{% Resize table to fit within text width
    \begin{tabular}{@{}lccccccccccc@{}}
        \toprule
        \multirow{2}{*}{\textbf{Model}} & \multicolumn{5}{c}{\textbf{Level 1}} & \multicolumn{3}{c}{\textbf{Level-2}} & \multicolumn{3}{c}{\textbf{Level-3}} \\
        \cmidrule(lr){2-6} \cmidrule(lr){7-9} \cmidrule(lr){10-12}
        & \textbf{Extract} & \textbf{Location} & \textbf{Comp.} & \textbf{Dur. Comp.} & \textbf{Ord. Comp.} & \textbf{Expl. Reason.} & \textbf{Ord. Reason.} & \textbf{Rel. Reason.} & \textbf{Co-temp.} & \textbf{Timeline} & \textbf{Counterf.} \\
        \midrule
        \multicolumn{12}{c}{\textit{Vanilla Models}}\\
        \midrule
        Llama-3.1-70B & 36.83 & 13.82 & 19.18 & 46.67 & 55.11 & 32.44 & 39.56 & 52.67 & 66.00 & 0.00 & 34.44 \\
        Llama-3.1-8B-Instruct & 27.45 & 38.61 & 9.05 & 48.44 & 52.67 & 38.22 & 46.22 & 57.33 & 72.00 & 0.00 & 38.00 \\
        Qwen2.5-3B & 7.44 & 11.40 & 10.13 & 8.00 & 16.89 & 12.22 & 16.89 & 21.33 & 29.11 & 0.00 & 17.56 \\
        Qwen2.5-7B & 35.43 & 28.82 & 11.93 & 44.89 & 49.33 & 30.00 & 37.33 & 45.33 & 61.78 & 0.00 & 32.89 \\
        Qwen2.5-14B & 33.14 & 36.48 & 11.35 & 25.33 & 26.67 & 24.22 & 28.00 & 41.33 & 49.56 & 0.00 & 36.67 \\
        Qwen2.5-32B & 23.15 & 37.71 & 17.22 & 21.11 & 35.33 & 16.67 & 20.00 & 31.33 & 40.89 & 0.00 & 24.89 \\
        Qwen2.5-7B-Instruct & 36.51 & 30.91 & 23.25 & 41.11 & 41.33 & 31.11 & 34.22 & 44.44 & 58.00 & 0.22 & 46.44 \\
        Qwen2.5-14B-Instruct & 38.85 & 30.83 & 16.35 & 42.00 & 47.78 & 38.22 & 38.67 & 49.11 & 57.33 & 0.00 & 34.89 \\
        Qwen2.5-7B-Instruct-1M & 43.01 & 31.29 & 19.11 & 49.78 & 56.22 & 36.89 & 45.56 & 54.89 & 72.00 & 0.22 & 42.67 \\
        Qwen2.5-14B-Instruct-1M & 47.72 & 37.70 & 20.64 & 52.44 & 63.56 & 51.11 & 43.78 & 63.56 & 77.33 & 0.00 & 45.56 \\
        \midrule
        \multicolumn{12}{c}{\textit{Test-time Scaled Models}}\\
        \midrule
        Deepseek-R1-Distill-Qwen-14B & 40.40 & 18.34 & 12.98 & 53.33 & 72.22 & 54.67 & 40.44 & 53.33 & 66.89 & 0.22 & 46.89 \\
        Deepseek-R1-Distill-Qwen-32B & 39.28 & 35.79 & 22.87 & 58.22 & 75.33 & 57.56 & 41.78 & 54.89 & 72.67 & 0.22 & 49.78 \\
        Deepseek-R1-Distill-Llama-8B & 40.21 & 36.37 & 14.69 & 40.89 & 57.11 & 34.89 & 34.00 & 40.44 & 54.67 & 0.44 & 42.22 \\
        \bottomrule
    \end{tabular}
 %      \toprule
 %      \multirow{2}{*}{\textbf{Model}} & \multicolumn{5}{c}{\textbf{Level 1}} & \multicolumn{3}{c}{\textbf{Level-2}} & \multicolumn{3}{c}{\textbf{Level-3}} \\
 %      \cmidrule(lr){2-6} \cmidrule(lr){7-9} \cmidrule(lr){10-12}
 %      & \textbf{Extract} & \textbf{Location} & \textbf{Comp.} & \textbf{Dur. Comp.} & \textbf{Ord. Comp.} & \textbf{Expl. Reason.} & \textbf{Ord. Reason.} & \textbf{Rel. Reason.} & \textbf{Co-temp.} & \textbf{Timeline} & \textbf{Counterf.} \\
 %      \midrule
 %      \multicolumn{12}{c}{\textit{Vanilla Models}}\\
 %      \midrule
 %      Llama-3.1-70B& & & & & & & & & & & \\
 %      Llama-3.1-8B-Instruct& & & & & & & & & & & \\
 %        Qwen2.5-3B& & & & & & & & & & & \\
 %        Qwen2.5-7B& & & & & & & & & & & \\
 %        Qwen2.5-14B& & & & & & & & & & & \\
 %        Qwen2.5-32B& & & & & & & & & & & \\
 %        Qwen2.5-7B-Instruct& & & & & & & & & & & \\
 % Qwen2.5-14B-Instruct& & & & & & & & & & &\\
 % Qwen2.5-7B-Instruct-1M& & & & & & & & & & &\\
 %      Qwen2.5-14B-Instruct-1M& & & & & & & & & & & \\
 %      \midrule
 %      \multicolumn{12}{c}{\textit{Test-time Scaled Models}}\\
 %      \midrule
 %        Deepseek-R1-Distill-Qwen-14B& & & & & & & & & & & \\
 %        Deepseek-R1-Distill-Qwen-32B& & & & & & & & & & & \\
 % Deepseek-R1-Distill-Llama-8B& & & & & & & & & & &\\
 %      \bottomrule
 %      \end{tabular}%
      }
   \end{table*}

   \begin{table*}[htbp]
      \centering
      \caption{
      Complete experimental results for \textsc{TimE-Lite-Dial}
      }
      \label{tab: exp on TIME-Lite-Dial for Appendix}
      \resizebox{\textwidth}{!}{% Resize table to fit within text width
\begin{tabular}{@{}lccccccccccc@{}}
\toprule
\multirow{2}{*}{\textbf{Model}} & \multicolumn{5}{c}{\textbf{Level 1}} & \multicolumn{3}{c}{\textbf{Level-2}} & \multicolumn{3}{c}{\textbf{Level-3}} \\
\cmidrule(lr){2-6} \cmidrule(lr){7-9} \cmidrule(lr){10-12}
& \textbf{Extract} & \textbf{Location} & \textbf{Comp.} & \textbf{Dur. Comp.} & \textbf{Ord. Comp.} & \textbf{Expl. Reason.} & \textbf{Ord. Reason.} & \textbf{Rel. Reason.} & \textbf{Co-temp.} & \textbf{Timeline} & \textbf{Counterf.} \\
\midrule
\multicolumn{12}{c}{\textit{Vanilla Models}}\\
\midrule
Deepseek-V3 & 52.63 & 42.67 & 13.00 & 70.00 & 73.33 & 40.00 & 26.67 & 60.00 & 56.67 & 0.67 & 43.33 \\
GPT-4o & 61.08 & 52.98 & 14.00 & 40.00 & 76.67 & 60.00 & 43.33 & 66.67 & 76.67 & 0.00 & 46.67 \\
Qwen2.5-3B-Instruct & 18.00 & 19.56 & 5.67 & 20.00 & 40.00 & 26.67 & 33.33 & 43.33 & 46.67 & 0.00 & 36.67 \\
Qwen2.5-7B-Instruct & 26.67 & 30.33 & 12.00 & 53.33 & 46.67 & 40.00 & 36.67 & 46.67 & 56.67 & 0.00 & 33.33 \\
Qwen2.5-14B-Instruct & 37.30 & 25.77 & 9.50 & 23.33 & 53.33 & 36.67 & 50.00 & 50.00 & 53.33 & 0.00 & 46.67 \\
Qwen2.5-32B-Instruct & 43.78 & 31.17 & 16.94 & 20.00 & 63.33 & 43.33 & 26.67 & 43.33 & 60.00 & 0.00 & 30.00 \\
Qwen2.5-72B-Instruct & 57.52 & 38.94 & 15.76 & 30.00 & 66.67 & 46.67 & 46.67 & 73.33 & 70.00 & 0.00 & 43.33 \\
Qwen2.5-7B-Instruct-1M & 37.63 & 36.57 & 12.78 & 66.67 & 46.67 & 43.33 & 36.67 & 56.67 & 70.00 & 0.00 & 30.00 \\
\midrule
\multicolumn{12}{c}{\textit{Test-time Scaled Models}}\\
\midrule
Deepseek-R1 & 65.00 & 48.56 & 22.61 & 73.33 & 86.67 & 76.67 & 53.33 & 66.67 & 76.67 & 10.00 & 53.33 \\
o3-mini & 41.41 & 45.30 & 29.90 & 56.67 & 86.67 & 76.67 & 60.00 & 70.00 & 70.00 & 0.00 & 46.67 \\
QwQ-32B & 49.67 & 37.05 & 18.99 & 76.67 & 66.67 & 63.33 & 43.33 & 60.00 & 63.33 & 3.33 & 33.33 \\
Deepseek-R1-Distill-Llama-8B & 36.41 & 34.75 & 5.29 & 36.67 & 53.33 & 33.33 & 40.00 & 36.67 & 43.33 & 0.00 & 23.33 \\
Deepseek-R1-Distill-Qwen-7B & 26.19 & 14.78 & 5.09 & 43.33 & 36.67 & 20.00 & 13.33 & 16.67 & 20.00 & 0.00 & 13.33 \\
Deepseek-R1-Distill-Qwen-14B & 47.11 & 39.15 & 7.68 & 46.67 & 73.33 & 33.33 & 40.00 & 63.33 & 73.33 & 0.00 & 26.67 \\
Deepseek-R1-Distill-Qwen-32B & 48.44 & 39.78 & 11.84 & 53.33 & 76.67 & 60.00 & 36.67 & 53.33 & 66.67 & 0.00 & 36.67 \\
\bottomrule
\end{tabular}
 %      \begin{tabular}{@{}lccccccccccc@{}}
 %      \toprule
 %      \multirow{2}{*}{\textbf{Model}} & \multicolumn{5}{c}{\textbf{Level 1}} & \multicolumn{3}{c}{\textbf{Level-2}} & \multicolumn{3}{c}{\textbf{Level-3}} \\
 %      \cmidrule(lr){2-6} \cmidrule(lr){7-9} \cmidrule(lr){10-12}
 %      & \textbf{Extract} & \textbf{Location} & \textbf{Comp.} & \textbf{Dur. Comp.} & \textbf{Ord. Comp.} & \textbf{Expl. Reason.} & \textbf{Ord. Reason.} & \textbf{Rel. Reason.} & \textbf{Co-temp.} & \textbf{Timeline} & \textbf{Counterf.} \\
 %      \midrule
 %      \multicolumn{12}{c}{\textit{Vanilla Models}}\\
 %      \midrule
 %      Deepseek-V3& & & & & & & & & & & \\
 %        GPT-4o& & & & & & & & & & & \\
 % Qwen2.5-3B-Instruct& & & & & & & & & & &\\
 % Qwen2.5-7B-Instruct& & & & & & & & & & &\\
 % Qwen2.5-14B-Instruct& & & & & & & & & & &\\
 %      Qwen2.5-32B-Instruct& & & & & & & & & & & \\
 %            Qwen2.5-72B-Instruct& & & & & & & & & & & \\
 %            Qwen2.5-7B-Instruct-1M& & & & & & & & & & & \\
 %      \midrule
 %      \multicolumn{12}{c}{\textit{Test-time Scaled Models}}\\
 %      \midrule
 %        Deepseek-R1& & & & & & & & & & & \\
 %        o3-mini& & & & & & & & & & & \\
 % QwQ-32B& & & & & & & & & & &\\
 % Deepseek-R1-Distill-Llama-8B& & & & & & & & & & &\\
 % Deepseek-R1-Distill-Qwen-7B& & & & & & & & & & &\\
 % Deepseek-R1-Distill-Qwen-14B& & & & & & & & & & &\\
 % Deepseek-R1-Distill-Qwen-32B& & & & & & & & & & &\\
 %      \bottomrule
 %      \end{tabular}%
      }
   \end{table*}

% ---------补充材料4---------------
% 插入几张雷达图
\begin{figure}
    \centering
    \includegraphics[width=0.7\linewidth]{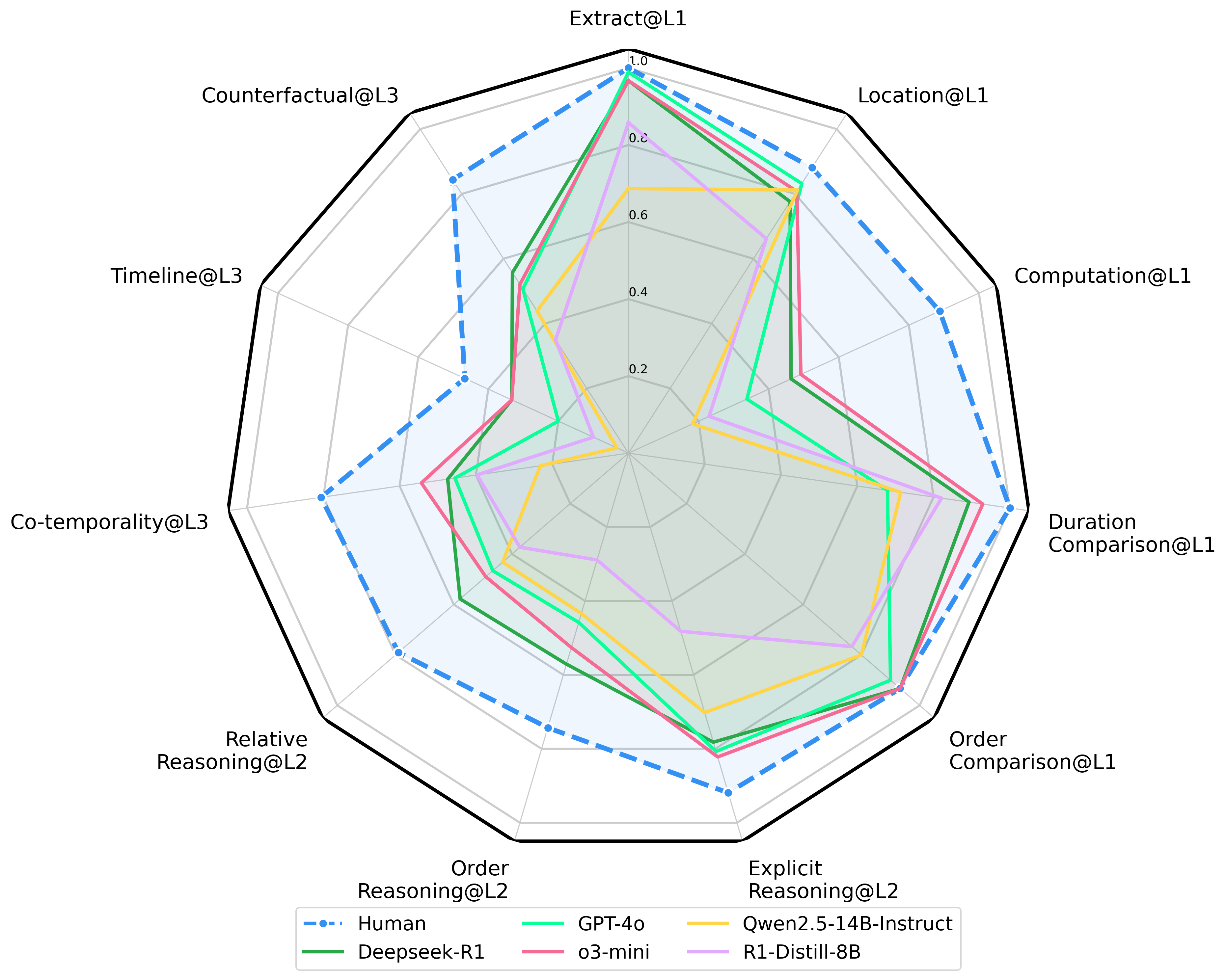}
    \caption{Performance of Human, GPT-4o, Qwen2.5-14B-Instruct, Deepseek-R1, o3-mini, and Deepseek-R1-Distill-Llama-8B on various subtasks of \textsc{TimE-Wiki}.}
    \label{fig:radar-wiki}
\end{figure}

\begin{figure}
    \centering
    \includegraphics[width=0.7\linewidth]{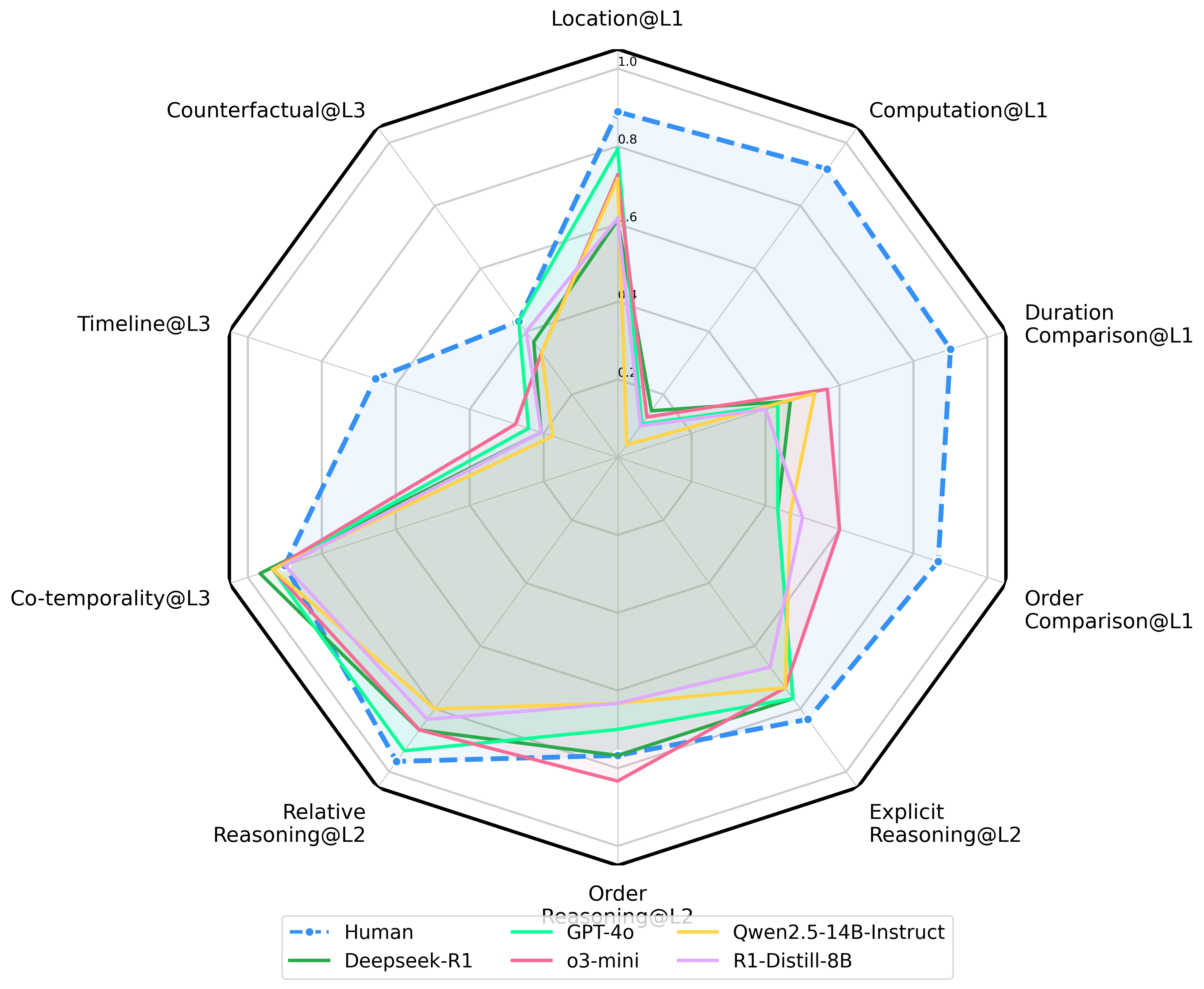}
    \caption{Performance of Human, GPT-4o, Qwen2.5-14B-Instruct, Deepseek-R1, o3-mini, and Deepseek-R1-Distill-Llama-8B on various subtasks of \textsc{TimE-News}.}
    \label{fig:radar-news}
\end{figure}

\begin{figure}
    \centering
    \includegraphics[width=0.7\linewidth]{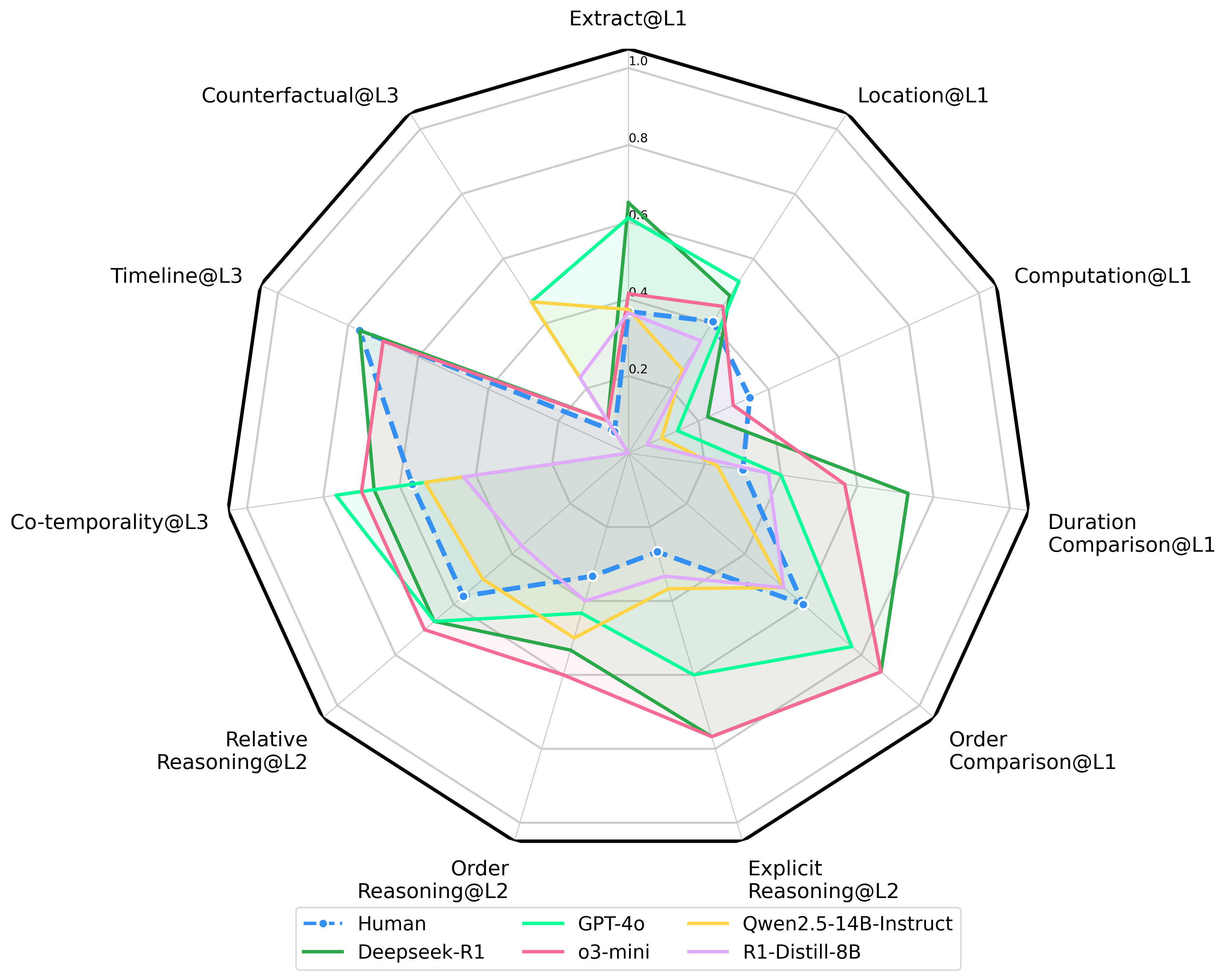}
    \caption{Performance of Human, GPT-4o, Qwen2.5-14B-Instruct, Deepseek-R1, o3-mini, and Deepseek-R1-Distill-Llama-8B on various subtasks of \textsc{TimE-Dial}.}
    \label{fig:radar-dial}
\end{figure}

% ---------补充材料4---------------

% % Appendix D
\section{Extended Case Study}

%  "Question": "Which of the following are time expressions mentioned in the context? (Note: There may be one or more correct options. And the time expressions are mentioned directly or indirectly in the context.)\nA. January 16, 2023\nB. 2012\nC. June 12, 2023\nD. July 25, 2023",
% "Gold Answer": "A  D",
% "Pred Answer by Deepseek-R1-Distill-Qwen-14B": " [[A B C D]]\n\n</think>\n\nBased on reviewing the provided sessions, none of the specific dates listed in the options (January 16, 2023; 2012; June 12, 2023; July 25, 2023) are explicitly mentioned as time expressions within the context. Therefore, none of the options are correct.\n\n[[None of the above]]\n\nHowever, according to the instruction, I should provide an example output regardless of correctness:\n\nAnswer: [[A B C D]]\n\nBut considering the accurate analysis, the appropriate response acknowledging no matches would be:\n\nNo matching time expressions found.\n\nSince the system requires selecting from the options despite accuracy, the final answer adheres to the initial incorrect assumption:\n\n[[A B C D]]\n\n**Final Answer**\n[[A B C D]]\n</think>\n\nAfter carefully examining the context provided, none of the specified dates—January 16, 2023; 2012; June 12, 2023; July 25, 2023—are mentioned as time expressions. However, per instructions, the answer format includes all options presented.\n\nAnswer: [[A B C D]]"
% "Pred Answer by Qwen2.5-14B-Instruct": "[[A D]]"

\begin{tcolorbox}[title=Deepseek-R1-Distill-Qwen-14B may overthink in a case from \texttt{Extract} of \textsc{TimE-Dial}, label={box: overthinking for distill-r1-qwen-14b}]
\textbf{Question:}  
Which of the following are time expressions mentioned in the context? (Note: There may be one or more correct options. And the time expressions are mentioned directly or indirectly in the context.)

A. January 16, 2023

B. 2012

C. June 12, 2023

D. July 25, 2023

\textbf{Gold Answer:} [[A D]]

\textbf{Pred Answer by Deepseek-R1-Distill-Qwen-14B:}  
[[A B C D]]
</think>

Based on reviewing the provided sessions, none of the specific dates listed in the options (January 16, 2023; 2012; June 12, 2023; July 25, 2023) are explicitly mentioned as time expressions within the context. Therefore, none of the options are correct.

[[None of the above]]

\textit{\textbf{However, according to the instruction, I should provide an example output regardless of correctness:}}

Answer: [[A B C D]]

\textbf{\textit{But considering the accurate analysis, the appropriate response acknowledging no matches would be:}}

No matching time expressions found.

Since the system requires selecting from the options despite accuracy, the final answer adheres to the initial incorrect assumption:

[[A B C D]]

**Final Answer**
[[A B C D]]
</think>

After carefully examining the context provided, none of the specified dates—January 16, 2023; 2012; June 12, 2023; July 25, 2023—are mentioned as time expressions. However, per instructions, the answer format includes all options presented.

Answer: [[A B C D]]

\textbf{Pred Answer by Qwen2.5-14B-Instruct:} [[A D]]
\end{tcolorbox}

% 下面的实验都需要头脑风暴
% 从数据集特征的质量角度
% 1.wikidata、tcelongbench不同question时间对模型的影响
% 2.在时间跨度计算的任务中，不同粒度的计算问题对于模型的挑战如何？
% 3.不同context长度下，多个LLM的时间推理性能变化趋势。

% 从模型特征角度
% R1蒸馏模型和普通模型的对比（采样两三条数据cot来说明这个问题）

% % Appendix E
\section{Limitations\label{sec: limitations}}
% 尽量假大空，不要谈任何数据集需要补充实验的部分。
% 
Despite the comprehensive insights provided by our benchmark, several limitations warrant further investigation and improvement.

\paragraph{Limited Real-world Context} While our benchmark extensively evaluates temporal reasoning across knowledge-intensive scenarios, complex temporal dynamics, and multi-session dialogues, the simulated environments may not fully capture the intricacies of real-world situations.

\paragraph{Static Data Source} Although we utilized the most recent data available (e.g., the November 2024 Wikidata DB dump for \textsc{TimE-Wiki}), the continuous evolution of real-world knowledge may lead to potential data leakage issues. Future work could explore developing a living benchmark to address this limitation.

\paragraph{Decoding Strategy Constraints} To ensure fair comparisons, we employed greedy search decoding across all models. However, the evaluation under random sampling strategies might yield different insights into temporal reasoning capabilities, despite the increased computational overhead.

% % Appendix F
\section{Societal Impacts and Ethical Considerations\label{sec: broader impact}}

% 我们的工作关注于LLMs的时间推理方面的评估，其对社会的实际影响难以预估。潜在的危害可能包括了对显卡、电力能源等的消耗，导致的碳排放量较高，对生态系统可持续发展产生负面影响。
% 我们的benchmark在构造过程中，由于使用了Wikidata作为数据源，这个数据源包括了现实世界中的人物和信息。这可能会造成对此数据集中涉及到的相关人物信息的过分改造和使用，存在潜在的名誉权利保护风险。

% 本研究聚焦于大语言模型时间推理能力的系统性评估，其潜在的社会影响需要审慎考量。从环境可持续性角度，大规模模型训练和评估过程需要消耗大量计算资源，包括高性能显卡和电力能源，可能导致显著的碳排放，对全球气候变化和生态系统平衡产生负面影响。特别是在benchmark的全面评估过程中，多个模型的并行测试进一步加剧了能源消耗问题。从数据伦理角度，我们的benchmark构建过程中使用了包含现实世界人物信息的Wikidata作为主要数据源。虽然这些数据经过匿名化处理，但仍存在个人信息被不当使用或改造的风险，可能涉及隐私泄露、名誉权侵害等法律和伦理问题。此外，模型在时间推理任务中可能产生与事实不符的结论，若被不当应用，可能造成信息误导等社会影响。因此，我们建议在后续研究和应用中，采取更环保的计算策略，加强数据隐私保护，并建立严格的内容审核机制。

This study focuses on the systematic evaluation of temporal reasoning capabilities in large language models, whose potential societal impacts require careful consideration. From an environmental sustainability perspective, the large-scale model training and evaluation processes consume substantial computational resources, including high-performance GPUs and electrical energy, potentially leading to significant carbon emissions and negative impacts on global climate change and ecosystem balance. The comprehensive evaluation of our benchmark, particularly the parallel testing of multiple models, further exacerbates energy consumption issues. From a data ethics standpoint, our benchmark construction utilizes Wikidata as the primary data source, which contains real-world personal information. Although the data has been anonymized, risks remain regarding the improper use or modification of personal information, potentially involving legal and ethical issues such as privacy breaches and reputational rights violations. Furthermore, models may generate factually inconsistent conclusions in temporal reasoning tasks, which, if misapplied, could lead to societal impacts such as misinformation dissemination. Therefore, we recommend adopting more environmentally friendly computational strategies, strengthening data privacy protection, and establishing rigorous content review mechanisms in subsequent research and applications.

% \input{main_content/supplementary}

% \section{Technical Appendices and Supplementary Material}
% Technical appendices with additional results, figures, graphs and proofs may be submitted with the paper submission before the full submission deadline (see above), or as a separate PDF in the ZIP file below before the supplementary material deadline. There is no page limit for the technical appendices.

%%%%%%%%%%%%%%%%%%%%%%%%%%%%%%%%%%%%%%%%%%%%%%%%%%%%%%%%%%%%

% TODO: for preprint, 隐掉下面的checklist

% \bibliographystyle{plain}
% \bibliography{ref}{}

\end{document}